\documentclass[10pt,final,journal,twocolumn,twoside]{IEEEtran}

\usepackage{cite} %
\PassOptionsToPackage{subfigure}{tocloft}

\usepackage[pdftex]{graphicx}
\graphicspath{{figures/}}
\DeclareGraphicsExtensions{.pdf}%
\usepackage[caption=false,font=footnotesize]{subfig}

\usepackage{amsfonts, amssymb, bm}
\usepackage[cmex10]{amsmath}
\usepackage{algorithm, algorithmicx, algpseudocode}

\makeatletter
\newcommand{\removelatexerror}{\let\@latex@error\@gobble}
\makeatother
\makeatletter
\algnewcommand{\NextLineComment}[1]{\Statex \hfill \(\triangleright\) #1}
\makeatother
\makeatletter
\algnewcommand{\LineComment}[1]{ \hskip\ALG@thistlm \(\triangleright\) #1}
\makeatother
\makeatletter
\newcommand{\algmargin}{\the\ALG@thistlm}
\makeatother
\newlength{\ifwidth}
\algdef{SE}[parIF]{parIf}{EndparIf}[1]
{\parbox[t]{\dimexpr\linewidth-\algmargin}{%
		\hangindent\ifwidth\strut\algorithmicif\ #1\ \algorithmicthen\strut}}{\algorithmicend\ \algorithmicif}%
\algnewcommand{\parState}[2]{\State%
	\parbox[t]{\dimexpr\linewidth-\algmargin}{\strut #1\strut#2}}
\algnewcommand{\myAnd}{\textbf{and}}
\algnewcommand{\myOr}{\textbf{or}}

\usepackage{booktabs}

\usepackage[usenames,dvipsnames]{xcolor}

\usepackage{fixmetodonotes}

\usepackage{url}
\usepackage{hyperref}

\usepackage[nolist]{acronym}

\usepackage[normalem]{ulem} %
\usepackage{color} %

\newcommand{\sizecorr}[1]{\makebox[0cm]{\phantom{$\displaystyle #1$}}}

\newcommand{\x}{\mathbf{x}} %
\newcommand{\X}{\mathbf{X}} %
\newcommand{\z}{\mathbf{z}} %

\newcommand{\A}{\mathbf{A}} %
\newcommand{\pinvA}{\mathbf{A}^-} %

\newcommand{\W}{\mathbf{W}} %
\newcommand{\w}{\mathbf{w}} %

\newcommand{\y}{\mathbf{y}} %
\newcommand{\Y}{\mathbf{Y}} %
\newcommand{\e}{\mathbf{e}} %

\newcommand{\PP}{\mathbf{P}} %
\newcommand{\R}{\mathbf{R}} %

\newcommand{\eye}{\mathbf{I}} %
\newcommand{\I}{\mathbf{I}}
\newcommand{\Sigmabf}{\bm{\Sigma}} %
\newcommand{\Lambdabf}{\bm{\Lambda}} %
\newcommand{\thetabf}{\bm{\theta}} %
\newcommand{\D}{\mathbf{D}} %

\newcommand{\norm}[1]{\left\lVert#1\right\rVert}
\newcommand{\frob}[1]{\norm{#1}_{\mathrm{F}}}
\newcommand{\abs}[1]{\left|#1\right|}
\DeclareMathOperator{\condi}{cond}
\newcommand{\cond}[1]{\condi\left(#1\right)}

\DeclareMathOperator{\argmin}{argmin}

\DeclareMathOperator{\mdim}{dim}
\DeclareMathOperator{\vect}{vec}

\newcommand{\cost}{\mathsf{cost}}
\newcommand{\find}{\mathsf{find}}
\newcommand{\match}{\mathsf{match}}
\newcommand{\rer}{\mathsf{remove\_empty\_rows}}
\newcommand{\osp}{\mathsf{subspace\_perm}}
\newcommand{\kurrent}{\mathsf{kurrent}}
\newcommand{\vals}{\mathsf{vals}}
\newcommand{\eps}{\mathsf{eps}}
\newcommand{\scc}{\mathsf{scale\_control}}
\newcommand{\false}{\mathsf{False}}
\newcommand{\true}{\mathsf{True}}
\newcommand{\ix}{\mathsf{ix}}

\newcommand{\MISA}{\textsf{MISA}}
\newcommand{\GP}{\textsf{GP}}
\newcommand{\MISAGP}{\textsf{MISA-GP}}
\newcommand{\MISAGPSDM}{\textsf{MISA-GP}_\textsf{SDM}}

\newenvironment{stretchpars}
 {\par\setlength{\parfillskip}{0pt}}
 {\par}

\begin{document}

\title{Multidataset Independent Subspace Analysis \\with Application to Multimodal Fusion}

\author{Rogers~F.~Silva$^\star$,~\IEEEmembership{Member,~IEEE,}
        Sergey~M.~Plis,
        T\"{u}lay~Adal{\i},~\IEEEmembership{Fellow,~IEEE,}
        Marios~S.~Pattichis,~\IEEEmembership{Senior~Member,~IEEE,}
        and~Vince~D.~Calhoun,~\IEEEmembership{Fellow,~IEEE}%
\thanks{Manuscript received November 7, 2019.}%
\thanks{This work was primarily funded by NIH grant R01EB005846 and NIH NIGMS Center of Biomedical Research Excellent (COBRE) grant 5P20RR021938/P20GM103472 to Vince Calhoun (PI), NSF 1539067, NSF 1631838, and NSF-CCF 1618551.}
\thanks{$^\star$Corresponding author.}%
\thanks{R.F. Silva, S.M. Plis, and V.D. Calhoun are with the Tri-Institutional Center for Translational Research in Neuroimaging and Data Science (TReNDS), Georgia State University, Georgia Institute of Technology, and Emory University, Atlanta, GA and The Mind Research Network, Albuquerque, NM USA (e-mail: rsilva@gsu.edu; splis@gsu.edu; vcalhoun@gsu.edu).}%
\thanks{M.S. Pattichis, and V.D. Calhoun are with the Dept. of ECE at The University of New Mexico, NM USA (e-mail: pattichi@unm.edu).}%
\thanks{T. Adal{\i} is with the Dept. of CSEE, University of Maryland Baltimore County, Baltimore, Maryland USA (e-mail: adali@umbc.edu).}%
\thanks{Digital Object Identifier }%
}

\markboth{Under Review}%
{Rogers F. Silva \MakeLowercase{\textit{et al.}}: Multidataset Independent Subspace Analysis with Application to Multimodal Fusion}

\IEEEpubid{\begin{minipage}{\textwidth}
\centering
\copyright~2019 Rogers Ferreira da Silva \MakeLowercase{\textit{et al.}}. Personal use of this material is permitted.\\
However, permission to use this material for any other purposes must be obtained from the authors.
\end{minipage}}

\maketitle

\begin{abstract}

  In the last two decades, unsupervised latent variable models---blind source separation (BSS) especially---have enjoyed a strong reputation for the interpretable features they produce.
  Seldom do  these models combine
  the rich diversity of information available in multiple datasets.
  Multidatasets, on the other hand, yield joint solutions otherwise unavailable in isolation, with a potential for pivotal insights into complex systems.

  To take advantage of the complex multidimensional subspace structures that capture underlying modes of shared and unique variability across and within datasets, %
  we present a direct, principled approach to multidataset combination.
  We design a new method called multidataset independent subspace analysis (MISA) that leverages joint information from multiple heterogeneous datasets in a flexible and synergistic fashion.

Methodological innovations exploiting the Kotz distribution for subspace modeling
in conjunction with a novel combinatorial optimization for evasion of local minima enable MISA to produce a robust generalization of independent component analysis (ICA), independent vector analysis (IVA), and independent subspace analysis (ISA) in a single unified model.

We highlight the utility of MISA for multimodal information fusion, including sample-poor regimes and low signal-to-noise ratio scenarios, promoting novel applications in both unimodal and multimodal brain imaging data.

\end{abstract}

\begin{IEEEkeywords}
BSS, MISA, multidataset, fusion, ICA, ISA, IVA, subspace, unimodal, multimodality, multiset data analysis, unify%
\end{IEEEkeywords}

\IEEEpeerreviewmaketitle

\begin{acronym}[L-BFGS-B]
	\setlength{\parskip}{-1ex}
	\input{myacronyms.acr}
\end{acronym}

\section{Introduction}

\acused{BSS} \acused{SOS} \acused{HOS} \IEEEPARstart{B}{lind} source separation (\ac{BSS})~\cite{Silva2016_UnifyBSSReview,ComonJutten2010Handbook} is widely adopted across multiple research areas in signal, image, and video processing, including chemometrics~\cite{YiL2016_ChemoMethodsReview}, speech~\cite{SaitoS2015_BSSAJDSpeech}, multispectral imaging~\cite{NielsenA2002_MCCA_MultispectralTemporal,AmmanouilR2014_BSSSparseHyperspectral}, medical imaging~\cite{Calhoun2009ReviewICAJoint,2016CalhounVD_FusionReview}, and video processing~\cite{BhingeS_IVA_MultiCameraObject,NicolaouM2014_DynProbCCA_VideoAnnotation}.
The ``blind'' property of \ac{BSS} models is highly advantageous, especially in applications lacking a precise model of the measured system(s) and with data confounded by noise of unknown or variable characteristics.
Its popularity has led to many formulations, methodologies, and algorithmic variations, at times making it difficult to build clear intuition about their connections and, thus, impeding further development toward flexible, more general models.

\begin{figure}[!t]
	\centering
	\includegraphics[width=\linewidth]{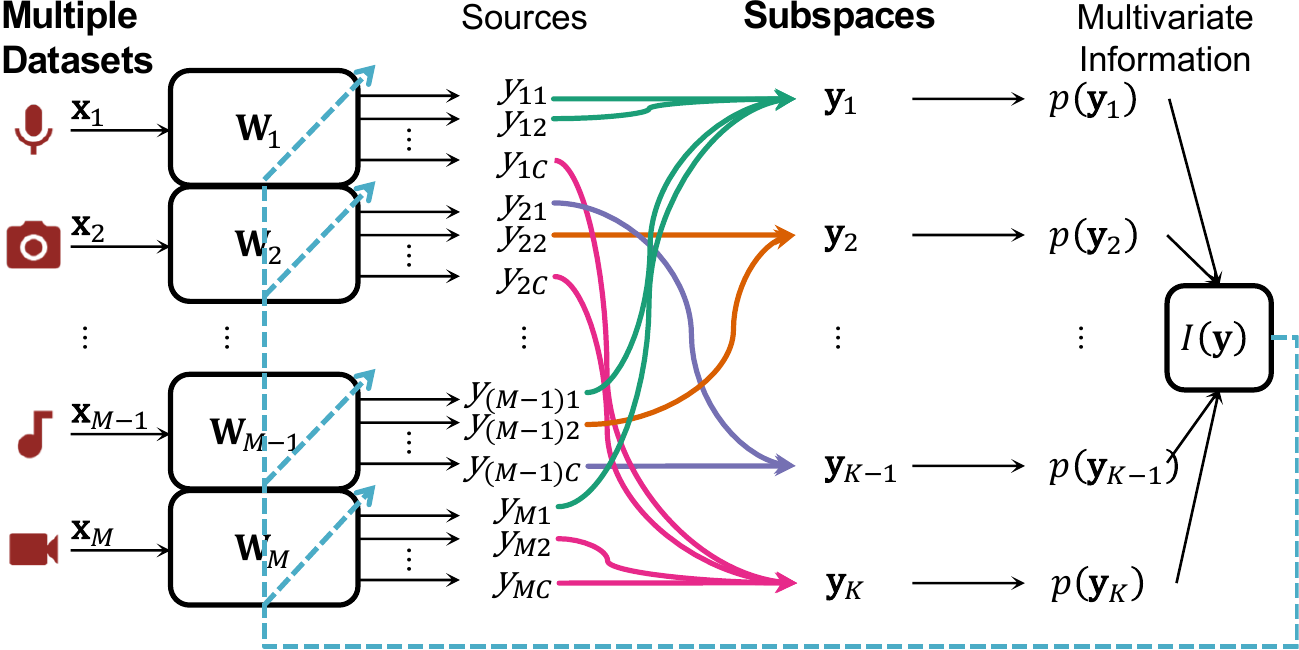}
	\caption{\textbf{Subspace identification from multidatasets with MISA}. We consider the general case of $M$ datasets ($\x_m$) \emph{jointly} decomposed, without loss of generality, into $C$ sources $y_{mi}$ each,  via linear transformations $\W_m$. Here, each $\x_m$ would be either audio or video streams, indicating fusion of different modalities via the joint analysis of all datasets. Sources are combined into $d_k$-dimensional \emph{subspaces} $\y_k$ and \emph{all-order} statistics is utilized to gauge their associations and pursue subspace independence. Only a single correspondence ``axis'' is required to make these assessments, such as temporal alignment in the case of audio/video (a/v) data fusion, although the method is not limited to a/v, fusion, nor temporal alignment specifically.  Subspaces establish links among groups of sources across different datasets/modalities. Therefore, \acf{MISA} blindly recovers hidden linked features of flexible dimensionality from multiple datasets and modalities.
	}
	\label{fig:diagram}
\end{figure}
\IEEEpubidadjcol In our recent review~\cite{Silva2016_UnifyBSSReview}, we addressed this issue by introducing a novel \ac{MMM} \ac{UMF} for subspace modeling. This helped us organize the \ac{BSS} field while casting new light on the relationships among \ac{BSS} models, eventually reconciling research conducted on both \ac{SDM} and \ac{MDU} problems.
The proposed framework revealed several development directions emphasizing new ways to identify general subspace correspondence across datasets, coined therein as \ac{MDM} problems.

Owing to their inherent flexibility, models designed for \ac{MDM} problems are highly advantageous.
With a single \emph{joint} model, it is possible to not only encode higher complexity through features of flexible dimensionality (the subspaces $\y_k$) but also accommodate arbitrary links among these features over multiple datasets/modalities ($\x_m$).
To illustrate, (Fig.~\ref{fig:diagram}) we consider a multivariate information functional $I(\y)$ that operates simultaneously on the joint \acp{pdf} of all subspaces $p(\y_k)$, capturing the association modes underlying multidatasets while adaptively learning multiple linear transformations $\W_m$ (dashed lines). %
This \emph{directly} leverages multidataset joint information and lets it guide the decomposition naturally.

\ac{MDM} problems constitute a largely undeveloped area of research with great potential to impact many fields. %
In classification, short of instances affected by the curse of dimensionality, multidimensional features are prone to yield better results, often owing to the increased feature space size.
The association/dependence inherent to multimodal features $\y_k$ means that good separability in one dataset will promote features with similar property in other datasets, and vice-versa.

In multimodal fusion of heterogeneous data ($\x_m$)~\cite{Lahat2015_MultimodalDataFusionReview, MillerKL2016_Multimodal_UKBiobank5K}, robust identification of joint features that originate from \emph{all} data modalities offers a one-of-a-kind view into the underlying properties of the system at hand.
It is a highly promising direction in mental health research, owing to its potential to identify biological markers of disease for early diagnosis, as well as to convey new strategies for disease severity assessment and translation into personalized treatments~\cite{Silva2016_UnifyBSSReview}.
Insights about the organization and function of complex systems, including the human brain, are indeed highly desirable.

The elevated flexibility of \ac{MDM} models would also be highly beneficial to various multiset analyses.
In the case of multisubject unimodal data ($\x_m$)~\cite{Calhoun2012ReviewNeurodicoveryICA,SeghouaneA2017_SequentialDicLearn,MohammadiNAR2017_ssCCAfusion,LeeIVAfMRI_NIMG08,BhingeS_ContrainIVA,PakravanM2018_JIMDM}, it would better preserve subject specificity.
In analyses that combine multi-site datasets ($\x_m$) from different scanners/devices, it could naturally mitigate harmonization
issues~\cite{YuM2018_HarmonizationMultisiteFMRI,MirzaalianH2016_HarmonizationMultisiteDiffusion} since site/device-variability would seldom explain multidataset associations.
In sensor fusion~\cite{Lahat2015_MultimodalDataFusionReview,AlamF2017_IoTFusion,ElmadanyN2018_GlobalLocalCCAFusion,NielsenA2002_MCCA_MultispectralTemporal}, where noise characteristics can be similar if multiple sensors ($\x_m$) share the same environment, it would allow better detection (and potential removal) of noise.
For hyperspectral imaging~\cite{UzairM2015_PLSRHyperspectralFace,YaoJ2019_NMFHyperspectral,VillaA2011_ICAHyperspectralJutten}, hyperspectral features ($\y_k$) of higher complexity could be identified in time-lapse studies.
For domain-adaptive image recognition~\cite{XuH2018_DomainAdaptiveDictLearn,FanJ2017_VisRecog_Groups,LuH_EmbarassDomainAdapt_ClassMean,LongM_JointMatchDomainAdapt,PatelVM_DomainAdaptOverview}, enhanced common and unique representations ($\y_k$) could be identified across image domains ($\x_m$).
For multi-view image and video processing~\cite{BhingeS_IVA_MultiCameraObject,CaiZ2014_CCA_Multi-viewVideo,TangL2018_CCAMultiViewObjRecog}, objects with complex temporal patterns could be better characterized using (unimodal) higher-dimensional $\y_k$, not to mention potential fusion with audio features~\cite{SarginM2007_CCA_AudioVisualFusion,GaoL2019_LabeledMCCA_AudioVisual,PierreN_IVAAudioVisual,NestaF2017_IVAAudioVisual} via multimodal $\y_k$.

Therefore, aiming to tackle the more general case of \ac{MDM} models that pursue statistical independence among \emph{subspaces} $\y_k$ to achieve joint \ac{BSS}, we initially introduced a solution called \ac{MISA} in~\cite{Silva2014MISA_OHBM,SilvaRF_MISA_ICIP,Silva2015MISASOS_OHBM}, demonstrating its feasibility, potential, and generalization. %
While novel, this preliminary version, to which we now refer as {$\alpha$-\ac{MISA}}, lacked robustness and performance across subspace configurations $\PP$ (Fig.~\ref{fig:MDMmodel}) due to premature convergence to local minima, rigid hard-coded subspace joint \ac{pdf} parameters%
, and a restricted approach for regularization of $\W_m$.

Here, we propose to address these issues by utilizing combinatorial optimization to search over $\PP$, as well as all-order statistics (i.e., both second- and higher-order statistics, SOS and HOS, respectively) to model $p(\y_k)$ and solve the \ac{MDM} problem while identifying statistically independent subspaces $\y_k$, referring to this improved approach simply as \ac{MISA} (Fig.~\ref{fig:diagram}).
Following Figs.~\ref{fig:diagram}~and~\ref{fig:MDMmodel}, let $p\left( \y \right)$ represent the joint \ac{pdf} of all sources, and $p(\y_k)$ the \ac{pdf} of the $k$-th subspace.

Let $I(\y)$ be the \ac{KL} divergence, an information functional useful for comparing two \acp{pdf} $p(\y)$ and $q(\y)$, where, here, $q\left( \y \right) = \prod_{k=1}^{K} p(\y_k)$ is the desired \emph{factor} \ac{pdf} of $p(\y)$.
Then let $h(\cdot)$ be the joint differential entropy, $h(\mathbf{z}) = -\mathbb{E} \left[ \ln p(\mathbf{z}) \right]$, for a random vector $\mathbf{z}$ with \ac{pdf} $p(\z)$, $\mathbb{E}\left[ \cdot \right]$ being the expected value operator, and let $\PP_k$ be an incomplete \emph{permutation matrix} that assigns specific sources into subspace $k$.
Consequently, it follows that:
\begin{IEEEeqnarray}{CCllCl}
	&&& \>\>\>\>\>\> I(\y) &=& -h(\y) + \sum_{k=1}^{K} h(\y_k) \label{eq:MISAprob}\\
	&&&&=& -h(\W\x) + \sum_{k=1}^{K} h(\PP_k\W\x). \nonumber
\end{IEEEeqnarray}
We propose to estimate a collection of linear transformations $\y = \W\x$ simultaneously from all datasets by solving:
\begin{IEEEeqnarray}{CCllCl}
	\min_{\W, \> \PP_k \forall k} &\>\>&  I(\y), & &&\label{eq:minimize}%
\end{IEEEeqnarray}
for any $\W$, subspace assignments $\PP_k$, and data streams $\x$.
This convenient formulation, which gives \ac{MI} when the random vector $\y$ is two-dimensional, only attains its lower bound of $I(\y) = 0$ when $p(\y) = q(\y)$, implying that the identified subspaces are indeed statistically independent.

With MISA, direct study of the interactions and associations among multiple datasets becomes feasible, in a truly synergistic way.
Consequently, joint sources $\y_k$ emerge naturally as a direct result of the shared variability estimated from all-order statistical dependences among datasets, while breaking from the limited, rigid paradigms of unidimensional models in order to allow subspace associations and even absent features in specific datasets.
Furthermore, as a unifying toolkit for independence-based \ac{MDM} models, \ac{MISA} not only takes the form of a single algorithm capable of executing any of the classical tasks as special cases, including \ac{ICA}~\cite{Comon1994ICANewConcept}, \ac{ISA}~\cite{CardosoMICA98}, and \ac{IVA}~\cite{KimT_IVA2006}, in addition to many others, but also outperforms several algorithms in each of these tasks, successfully achieving generalized subspace identification from multidatasets.
Having this uniform implementation makes model connections very intuitive and accessible, thanks to the umbrella formulation and methodologies introduced in this work.

Multiple experiments obeying the principles outlined in~\cite{SilvaRF_MMSimNIMG2014} demonstrate that MISA outperforms algorithms such as Infomax~\cite{BellSejnowski1995Infomax,Amari1998_NaturalGrad}, IVA Laplace (IVA-L)~\cite{KimT_IVA2006}, and IVA Gaussian Laplace (IVA-GL)~\cite{Anderson2012_JBSSGaussianAlgPerf} in various challenging realistic scenarios, with remarkable performance and stability in certain extremely noisy cases (SNR of 0.0043dB), highlighting the benefit of careful multidataset subspace dependence modeling with all-order statistics.
Likewise, MISA with greedy permutations (MISA-GP) clearly outperforms joint blind diagonalization with \ac{SOS} (JBD-SOS)~\cite{LahatD2012_MICA_SOS} and EST\_ISA~\cite{LeQ2011_ISA} even at low \ac{SNR} levels (SNR of 3dB), highlighting the benefit of the proposed combinatorial optimization approach to escape local minima in subspace analyses.
Hybrid data results further support the high estimation quality and flexibility benefits of MISA for novel applications in high-temporal-resolution \ac{fMRI} analysis, and multimodal fusion of heterogenous neurobiological signals.

\section{Background}\label{sec:Background}
\subsection{General \ac{MDM} Problem Statement}\label{sec:probstate}
\begin{figure}[!t]
	\centering
	\includegraphics[scale=.75]{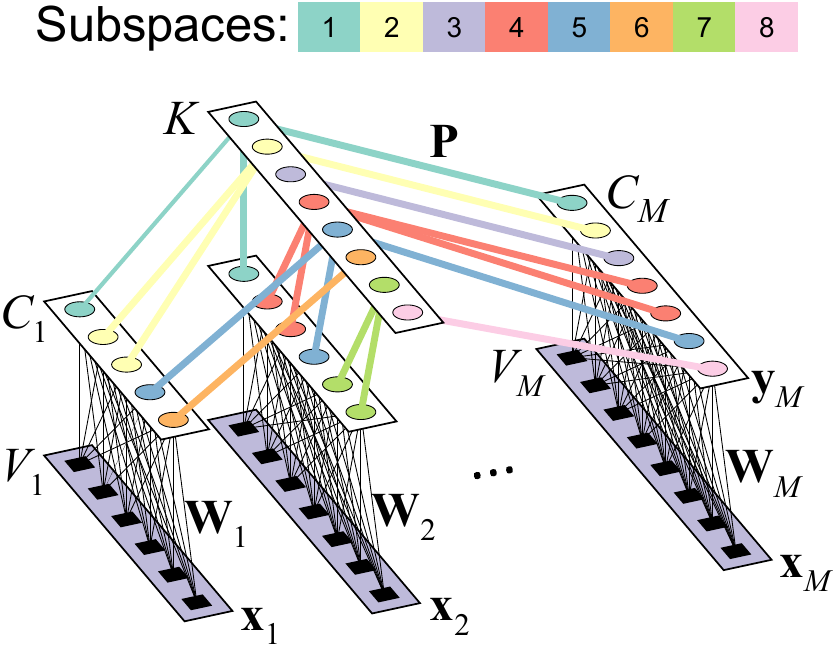}
	\caption{\textbf{General decompositional representation of the linear \ac{MDM} problem}. The lower layer corresponds to one $V_m \times 1$ observation of each input data stream $\x_m$. The middle layer represents the $C_m$ sources. The top layer establishes the $K$ subspaces $\y_k$, which are collections of statistically \emph{dependent} sources (indicated by same-colored connections). They follow an assignment matrix $\PP$, illustrating the different compositions permitted.
	}
	\label{fig:MDMmodel}
\end{figure}

\ac{BSS} admits a hierarchical organization in accordance with the number of datasets comprising $\x$
and the occurrence of multidimensional sources (grouped subsets of  $\y$) within any single dataset~\cite[Ch.~8]{Silva2016_UnifyBSSReview,SilvaRF2019_BookPPP_Ch8}.
For example, \ac{MDU} and \ac{SDM} problems contain the simpler \ac{SDU} case, and the most general \ac{MDM} problem (see Fig.~\ref{fig:MDMmodel}) contains all others as special cases.

Formally, the \ac{MDM} problem can be stated as follows.
Given $N$ observations of $M \ge 1$ datasets, identify an unobservable  latent source random vector (r.v.)  $\y = \left[\y_1^\top \cdots \y_M^\top\right]^\top$, $\y_m = [y_{m1} \cdots y_{mC_m}]^\top$, from an observed r.v. $\x = \left[\x_1^\top \cdots \x_M^\top\right]^\top$, $\x_m= [x_1 \cdots x_{V_m}]^\top$  via   a    mixture \emph{vector}   function  ${\bf f}\left(\y,\thetabf\right)$ with unknown parameters $\thetabf$.
The $m$-th $V_m \times N$ data matrix containing $N$ observations of $\x_m$ along its columns is denoted $\X_m$, and the $\bar{V} \times N$ matrix concatenating all $\X_m$ is denoted simply as $\X$ (likewise for $\Y$ and $\Y_m$).
Both $\y$ and the transformation ${\bf f}\left(\y,\thetabf\right)$ have to be learned \emph{blindly}, i.e., without explicit  knowledge of  either of them.
For tractability, assume:
\begin{IEEEenumerate}
	\item the number of latent sources $C_m$, which may differ in each dataset, is known to the experimenter;
	\item ${\bf f}\left(\y,\thetabf\right) = \A \y$ is a \emph{linear} transformation, with $\thetabf = \A$;
	\item $\A$ is a $\bar{V} \times \bar{C}$ block diagonal matrix with $M$ blocks, representing a \emph{separable layout} structure~\cite{Silva2016_UnifyBSSReview} defined as $\x_m = \A_m \y_m$, $m = 1 \ldots M$, where $\bar{C} = \sum_{m=1}^{M}C_m$, $\bar{V} = \sum_{m=1}^{M}V_m$, each block $\A_m$ is $V_m \times C_m$, and $V_m$ is the intrinsic dimensionality of each dataset;
	\item some latent sources $y_{mi} \in \y$ are statistically related to each other, and this \emph{dependence} is undirected (non-causal), occurring within and/or across datasets;
	\item related sources establish $d_k$-dimensional subspaces%
	\footnote{The subspace terminology stems from~\cite{CardosoMICA98} in which the columns of $\A$ corresponding to $\y_k$ form a linear (sub)space.}
	(or source \emph{groups}) $\y_k$, $k = 1 \ldots K$, with $K$ and the specific subspace compositions known by the experimenter and listed in a sparse assignment matrix ${\PP \in \{0,1\}^{K \times \bar{C}}}$ containing a single non-zero entry in each column;
	\item subspaces do not relate to each other, i.e., either $p\left( \y \right) = \prod_{k=1}^{K} p(\y_k)$ or the cross-correlations $\rho_{k,k'} = 0$, $k \ne k'$.~\label{item:unrelate}
\end{IEEEenumerate}
Under these assumptions, recovering sources $\y$ amounts to finding a linear transformation $\W$ %
for the unmixing vector function $\y = \W \x$.
This is accurate when $\W = \pinvA$, the pseudo-inverse of $\A$, which implies $\W$ is also block diagonal, thus satisfying $\y_m = \W_m \x_m$.
As stated in~\cite{Silva2016_UnifyBSSReview}, the experimenter's priors on the subspace structure within/between one or more datasets, plus the type of statistics describing within/between subspace relation, determines whether and how the model simplifies to any of the special cases.
In the following, we focus on the more general case of a \ac{MDM} model driven by \emph{statistical independence} among subspaces and \emph{dependence} within subspaces, namely \ac{MISA}, in the case of an overdetermined system with $V_m \ge C_m$, without implying $\W$ is square with the typical \ac{PCA}.

\subsection{Current Challenges and Our Contributions}\label{sec:BG:statscale}
Premature convergence to local minima due to the mis-assignment of sources to subspaces is a challenge for \ac{MDM} model fitting.
It has roots in \ac{ICA}'s \emph{permutation ambiguity} property that no particular ordering of the true independent sources alters their independence, in effect, describing a class of equivalent globally optimal solutions.
While this property extends trivially to independent subspaces, arbitrary source permutations are no longer equivalent due to likely incorrect assignments of those sources into the subspaces.
Hence, this is a key point that requires special attention.

To illustrate, assume $\PP_k$ is a user-specified prior.
Using abbreviated notation, suppose $\PP_1 = \left[ 1~1~1~0~0 \right]$ and $\PP_2 = \left[ 0~0~0~1~1 \right]$ defines a partitioning of five sources into two subspaces: $p(\y) = p(\y_{k=1})p(\y_{k=2}) = p(y_1,y_2,y_3)p(y_4,y_5)$, where $p(\cdot)$ is the joint pdf of a subspace.
It would be equally acceptable if the data supported either $p(\y) = p(y_4,y_5)p(y_1,y_2,y_3)$ (entire subspace permutation) or $p(\y) = p(y_1,y_3,y_2)p(y_5,y_4)$ (within-subspace permutation) or even some combination of these two cases.
However, if the data supported $p(\y) = p(y_1,y_4)p(y_2,y_3,y_5)$, then that would not be equivalently acceptable.
The latter characterizes a local minimum, while the former two constitute global minima.

In practice, these are hard-to-escape local minima~\cite{HyvarinenNIS09}.
In \ac{MISA}, we have observed that a drastically more intricate joint interplay exists between within- and cross-dataset subspace-to-subspace interactions, which yields a combinatorial increase in number of these non-optimal local minima.
Specifically, when subspaces span multiple datasets, very high $\prod_{k=1}^{K} {\bar{C} - \sum_{l=0}^{k-1} d_{l} \choose d_k}$ number of possible local minima (upwards of $6\cdot10^{19}$ in Section~\ref{subsubsec:ISA}) cripples the numerical optimization performance (here, $d_0 \triangleq 0$).
Thus, we propose combinatorial optimization algorithms to enable evasion of local minima in the numerical optimization of~\eqref{eq:MISAprob}.

While combinatorial issues are common in other research areas~\cite{BouwmansT2018_RPCA_OtazoR,PontTusetJ2017_CombolGroupingSegment,ElZehiryN2016_ComboElasticaSegmentation}, they have been largely neglected in the \ac{BSS} literature because of how simple (and often irrelevant) they are for \ac{ICA}.
To our knowledge, this is the first attempt at disentangling these permutation ambiguities in the general \ac{MISA} case.
Inspired by a strategy proposed in~\cite{Szabo2012ISAbyICAPerm} for \ac{ISA}, separate combinatorial optimization procedures to address general \ac{SDM} and \ac{MDM} problems are presented in Sections~\ref{sec:isaperm} and~\ref{sec:misaperm}, respectively.
In contrast to~\cite{Szabo2012ISAbyICAPerm}, however, our approach does not rely on additional accessory objective functions to determine residual source dependences.
Moreover, our approach leverages the structural subspace priors contained in $\PP_k$ to guide the combinatorial procedures.
Ultimately, the key difference is that our approach serves only to move a particular solution out of a local minima so that the numerical optimization may resume.

Another challenge stems from the distribution properties of $p(\y_k)$.
The \emph{a priori} selection/hard-coding of distribution $p(\cdot)$ is common practice in \ac{BSS}, often based on some domain knowledge about source properties, e.g., sub- or super-Gaussian attributes~\cite{BellSejnowski1995Infomax,LeeTW_ExtendedICA}.
Likewise, the subspace covariance structure $\Sigmabf_k^{\y}$ is often hard-coded to identity (i.e., uncorrelated)~\cite{HyvarinenESANN06,LeeIVAfMRI_NIMG08} or just iteratively substituted with data estimates~\cite{AndersonM_2012IVAKOTZ, Anderson2012_JBSSGaussianAlgPerf}  in \ac{MDU} and \ac{SDM} problems.
While these practices facilitate tractability (also the case in our previous work) they curb generalization beyond the context of model conception and can altogether fail to capture the subspace structure supported by the data.
Even objective function gradients can become inconsistent in the case of simple iterative substitution, despite its intent to encourage data-supported dependences.

In order to help counter the rigidity and immutability of said choices, we pursue the use of a generalized \ac{pdf}~\cite{Kotz1975_MVDistCrossRoad,KimT_IVA2006,EltoftT2006_MultivariateK} that includes additional parameters to better cater for sources that the data supports, as suggested in~\cite[Sec. 3.3.2]{AndersonM2013_IVAthesis} and adopted in~\cite{LiX_ICAEBM} for \ac{ICA}.
Following~\cite{AndersonM_2012IVAKOTZ}, we introduce the Kotz distribution~\cite{Kotz1975_MVDistCrossRoad} for \ac{MISA} due to its generality, as it includes the multivariate power exponential family as a special case.
Contrary to approaches based on the multivariate Gaussian~\cite{LahatD2012_MICA_SOS,Anderson2010IVAG,Lahat2016_JISASOS_IEEETSP}, which incorporates only \ac{SOS}, and the multivariate Laplace (with uncorrelated subspace covariance $\Sigmabf_k^{\y}=\I$, incorporating only \ac{HOS}) \cite{HyvarinenESANN06,LeeIVAfMRI_NIMG08,Silva2015MISASOS_OHBM}, the Kotz distribution can account for both and enable all-order statistics.
Moreover, we extend~\cite{AndersonM_2012IVAKOTZ} in two ways: 1) we address the \ac{MDM} case rather than just the \ac{MDU} case, and 2) we improve performance considerably by casting $\Sigmabf_k^{\y}$ as a direct function of the linear transformation parameter being optimized, namely $\W$, into $p(\y_k)$, rather than treating it as constant with respect to $\W$  in spite of its iterative updates.

Finally, there is a challenge with source scale estimation.
The property of independence between sources is inherently invariant to arbitrarily scaling each or any source, which is why \ac{ICA} sources have \emph{scale ambiguity}.
This has an important implication on the geometry of the resulting objective function we seek to optimize.
First, visualize the elements of $\W$ into a $\bar{D}$-dimensional vector ($\bar{D} = \bar{V}\bar{C}$) $\w = \vect{(\W)}$ as would be done in a typical numerical optimization setting.
Due to scale invariance, evaluation of the objective function on either $\w$ or $a\w$, where $a$ is a non-zero scalar, yields the same value.

\begin{figure}[!t]
	\centering
	\captionsetup[subfloat]{justification=centering}
	\subfloat[Concentric hyperspheres]{\includegraphics[scale=.58]{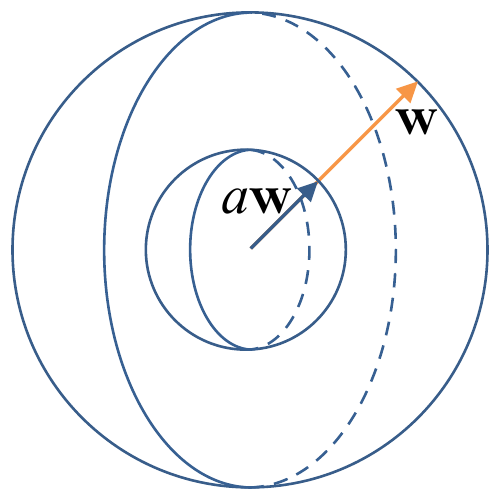}%
		\label{fig:figcircle1}}
	\hfil
	\subfloat[Contour lines]{\includegraphics[scale=.58]{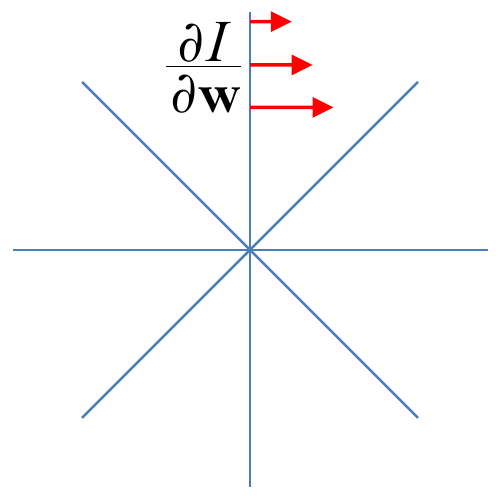}%
		\label{fig:figstar}}
	\hfil
	\subfloat[Tangent plane]{\includegraphics[scale=.58]{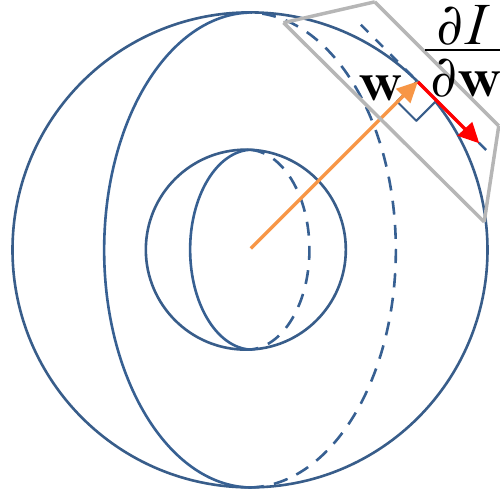}%
		\label{fig:figcircle2}}
	\caption{\textbf{Geometry of the independence-driven objective function in SDU problems}. (a) The solution space of independence-driven SDU problems lies on a hypersphere. (b) Scale invariance induces a ``star'' shape on the contour lines. (c) Consequently, the gradient of a scale invariant function must lie on the tangent hyperplane of the hypersphere associated with a given $\w$.}
	\label{fig:hyper}
\end{figure}

Since the objective function evaluates to the same values along the line%
\footnote{Strictly speaking, this line is only a portion of the entire hyper surface (polyhedron) of ambiguity.}
spanned by $\w$, only certain changes in the \emph{direction} of $\w$ incur changes in the objective function.
Consequently, it suffices to look for a solution on the surface of the \emph{hypersphere} associated with a given $a$, since the landscape of objective function values would be identical across concentric (hyper) shells (Fig.~\ref{fig:hyper}~(a)).
Moreover, scale invariance induces a ``star'' shape to the contour lines of the objective function in this scenario (Fig.~\ref{fig:hyper}~(b)).
Since gradients are orthogonal to contour lines, they also ought to be orthogonal to $\w$ and lie on the \emph{tangent} hyperplane of any given hypersphere (Fig.~\ref{fig:hyper}~(c)).

The main implication is that stepping in the (negative) direction of the gradient towards a local minimum will likely \emph{inflate} $\w$ and lead the search direction in an outward spiral with respect to $\w$.
This can be a problem if the norm of $\w$ grows indefinitely and eventually becomes numerically unstable.
More importantly, as the norm of $\w$ increases toward %
outer shells, the landscape of the objective function starts to stretch (because its values are kept the same while the surface area of the hypersphere grows).
Consequently, the \emph{gradient grows shorter regardless of its proximity to any local minimum}.
The smaller gradient will then lead to shorter step lengths, likely yielding very little improvement at latter stages of the numerical optimization and deterring convergence.

This issue is often disregarded in the \ac{SDU} literature (the Infomax algorithm~\cite{BellSejnowski1995Infomax} is an exception) %
and should be addressed prior to evaluation of the efficient relative gradient~\cite[Ch. 4]{ComonJutten2010Handbook}.
One simple approach to address it is to constrain the norm of $\w$.
While direct, implementing this approach can be quite inefficient.
Rather, since any scale is equally acceptable (at least in theory), we propose to control the estimated source scales by fixing them in the \emph{model}.
Specifically, this is accomplished by assigning the estimated subspace correlation matrix $\R^{\y}_k$ as the model dispersion matrix $\D_k$ in the Kotz distribution, effectively making the objective function \emph{scale selective} rather than scale invariant (Section~\ref{sec:METH:objfun}).
Therefore, whenever the source estimates from the data do not support the model variances associated with this choice of $\D_k = \R^{\y}_k$, the mismatch induces changes in $\W$ that lead their variances towards the prescribed ones.
In summary, the proposed scale selective formulation eliminates scaling issues
without the need for a formal constraint.

We also introduce the pseudo-inverse reconstruction error (PRE) as a nonlinear constraint for estimation of rectangular non-invertible mixing matrices (also serving as a flexible alternative for data reduction), and utilize quasi-newton optimization based on the relative gradient for efficient optimization.

\section{Methodology}\label{sec:methodology}

\subsection{Objective Function}\label{sec:METH:objfun}

Equation~\eqref{eq:MISAprob} admits some simplifications following a few manipulations.
First, we note that $h(\y) = h(\W \x) = h(\x) + \ln\abs{\det(\W)}$, and $h(\x)$ can be discarded since it is constant with respect to $\W$.
Second, $\ln\abs{\det(\W)} = \sum_{m=1}^{M} \ln\abs{\det(\W_m)}$ since $\W$ is block diagonal.
Finally, when $V_m \ne C_m$, for any $m$, the determinant of $\W_m$ is undefined.
In order to circumvent this issue, we propose to substitute the determinant by the product of the singular values of $\W_m$, i.e., $\prod_{i = 1}^{C_m} \sigma_{mi}$, where $\sigma_{mi}$ are the diagonal elements of $\Lambdabf_m^{} = \mathbf{U}_m^\top \W_m^{} \mathbf{V}_m^{}$ originating from the singular value decomposition $\W_m^{} = \mathbf{U}_m^{} \Lambdabf_m^{} \mathbf{V}_m^{\top}$.
We note that $\abs{\det \W} = \prod_{i=1}^{C} \abs{\sigma_{mi}}$ when $\W$ is non-singular and square.
Altogether, the minimization problem can then be recast as:
\begin{IEEEeqnarray}{CCl}
	\min_{\W, \PP_k \forall k} && \>\> \check{I}(\y), \>\> \check{I}(\y) = -\sum_{m=1}^{M} J_{D_m} - \sum_{k=1}^{K} \mathbb{E} \left[ \ln p(\y_k) \right] \mathrm{,} \IEEEeqnarraynumspace \label{eq:MISAprob_check}%
\end{IEEEeqnarray}
where $J_{D_m} = \sum_{i = 1}^{C_m} \ln\abs{\sigma_{mi}}$, and $\y_k = \PP_k\W\x$.

This formulation is still incomplete because $p(\y_k)$ is undefined.
Here we choose to model each subspace \ac{pdf} as a multivariate Kotz distribution~\cite{NadarajahS2003_KotzDistApps,Kotz1975_MVDistCrossRoad}:
\begin{IEEEeqnarray}{CCl}
	p(\y_k) = \frac%
	{\beta_{k}^{} \lambda_k^{\nu_k} \Gamma \left( \frac{d_k}{2} \right) \left( \y_k^\top \mathbf{D}_k^{-1} \y_k^{} \right)^{\eta_k - 1}}%
	{\pi^{\frac{d_k}{2}} \left( \det \mathbf{D}_k \right)^{\frac{1}{2}} \Gamma \left( \nu_k \right)}%
	e^{-\lambda_k \left( \y_k^\top \mathbf{D}_k^{-1} \y_k^{} \right)^{\beta_k}} \IEEEeqnarraynumspace \label{eq:Kotzpdf}
\end{IEEEeqnarray}
where $d_k$ is the subspace dimensionality, $\beta_k > 0$ controls the shape of the \ac{pdf}, $\lambda_k > 0$ the kurtosis (i.e., the degree of peakedness), and $\eta_k > \frac{2-d_k}{2}$ the hole size, while $\nu_k \triangleq \frac{2\eta_k + d_k - 2}{2 \beta_k} > 0$ and
$\alpha_k \triangleq \frac
{\Gamma \left( \nu_k^{} + \beta_k^{-1} \right)}
{\lambda_k^{\beta_k^{-1}} d_k^{} \Gamma \left( \nu_k^{} \right)}$
for brevity.
$\Gamma \left( \cdot \right)$ denotes the gamma function.
The positive definite dispersion matrix $\mathbf{D}_k$ is related to the covariance matrix $\Sigmabf_k^{\y}$ by ${\mathbf{D}_k = \alpha_k^{-1}\Sigmabf_k^{\y}}$.
This is a good choice of \ac{pdf} since it generalizes and includes other classical distributions such as the multivariate Gaussian and the multivariate Laplace distributions by setting the \ac{pdf} parameter set $\bm{\psi}_k = [\beta_k, \lambda_k, \eta_k]$ to $\bm{\psi}_G = [1,\frac{1}{2},1]$ and $\bm{\psi}_L = [\frac{1}{2},1,1]$, respectively.
The second term in~\eqref{eq:MISAprob_check} admits the following form:
\begin{IEEEeqnarray}{CCl}
	\ln p(\y_k) = \ln \left( \frac%
	{\beta_{k}^{} \lambda_k^{\nu_k} \Gamma \left( \frac{d_k}{2} \right) }%
	{\pi^{\frac{d_k}{2}} \Gamma \left( \nu_k \right)} \right)
	+ \frac{(\eta_k - 1) J_{F_k}}{\frac{1}{2} J_{C_k}} -\lambda_k J_{E_k} \IEEEeqnarraynumspace \label{eq:Kotzpdf}
\end{IEEEeqnarray}
where $J_{C_k} = \ln \det \mathbf{D}_k$, $J_{F_k} = \ln \left( \y_k^\top \mathbf{D}_k^{-1} \y_k^{} \right)$, and ${J_{E_k} = \left( \y_k^\top \mathbf{D}_k^{-1} \y_k^{} \right)^{\beta_k}}$.

The minimization problem in (\ref{eq:MISAprob_check}) is equivalent to maximizing the likelihood of $\y_k$.
In the following, we estimate $\Sigmabf_k^\y$ from the data.
This is appealing because the sample average of $\bar{\Sigmabf}^{\x}$ is readily available and can be conveniently combined with $\W$ to produce an approximation of $\Sigmabf_k^{\y}$ for substitution in $\mathbf{D}_k$. %
This simple choice permits the reparameterization of $\Sigmabf_k^{\y}$ as a function of $\W$, specifically $\bar{\Sigmabf}_k^\y = \frac{1}{N-1}\PP_k^{}\W\X\X^\top\W^\top\PP_k^\top$.%

Two well-conceived dispersion matrix parameter choices are proposed for the Kotz distribution, one emphasizing invariance to source scales and the other not, resulting in two useful objective functions.
Accordingly, we present the two final forms of (\ref{eq:MISAprob_check}), using ${\Y_k = \PP_k \W \X}$ and $n$ to index each of the $N$ observations used in the sample mean approximation of the expected value $\mathbb{E}[\cdot]$. For the standard \emph{scale invariant} case, we have:
\begin{IEEEeqnarray}{Lcl}
	\check{I}(\y) & \> = \> &
	- \sum_{m=1}^M J_{D_m} + \frac{1}{2}\sum_{k=1}^K J_{C_k}
	- f_{\left(K,\beta_k,\lambda_k,\eta_k,d_k,\nu_k\right)} \nonumber
	\\
	&&
	- \sum_{k=1}^K \frac{\eta_k-1}{N}\sum_{n=1}^{N} J_{F_{kn}}
	+ \sum_{k=1}^K \frac{\lambda_k}{N}\sum_{n=1}^{N} J_{E_{kn}} \mathrm{,}
	\label{eq:MISAobjective}
\end{IEEEeqnarray}
\noindent where
\begin{IEEEeqnarray*}{Rll}
	f_{\left(K,\beta_k,\lambda_k,\eta_k,d_k,\nu_k\right)} & = \> &%
	\sum_{k=1}^K \left[
	\ln\beta_k + \nu_k \ln \lambda_k  + \ln\Gamma\left( \frac{d_k}{2} \right) \right. \nonumber \\
	&& \qquad \left. 
	\sizecorr{\ln\Gamma\left( \frac{d_k}{2} \right)}
	- \frac{d_k}{2} \ln \pi - \ln\Gamma\left( \nu_k \right)	\right] \mathrm{,} \nonumber
\end{IEEEeqnarray*}
\noindent with gradient given by:
\begin{IEEEeqnarray}{RCL}
	\nabla\check{I}(\W)_{mi_k} & \> = \> &
	\left[B_k^{}+\left[I-B_k^{}\mathbf{Y}_k^\top\right]A_k^{}\right]\X_m^\top - \left(\W_m^-\right)^\top \IEEEeqnarraynumspace \label{eq:MISAgrad}
	\\
	&& \>\> _{\forall k \> \in \> \{1,\ldots,K\}, \>\> \forall m \> \in \> \{1,\ldots,M\}} \nonumber
\end{IEEEeqnarray}
\noindent where $i_k$ represents all source indices (rows of $\nabla\check{I}(\W)_{m}$) assigned to subspace $k$, $\circ$ is the Hadamard product, and
\begin{IEEEeqnarray*}{RCL}
	A_k^{}  & \> = \> & \left[\bar{\Sigmabf}_{k}^{\y}\right]^{-1}\mathbf{Y}_k^{}\nonumber\\
	B_k  & \> = \> & A_k \>\mathrm{diag}\>(\mathbf{t}_k)\nonumber\\
	\mathbf{t}_k^{} & \> = \> & \left(\frac{2\beta_k\lambda_k}{N}\mathbf{z}_k^{\beta_k}+\frac{2\left(1-\eta_k\right)}{N}\right)\circ\mathbf{z}_k^{-1}\\
	\mathbf{z}_k & \> = \> & \left[z_{k1},z_{kn}, \>\ldots,\> z_{kN}\right]\\
	z_{kn}^{} & \> = \> & \mathbf{y}_{kn}^\top\left[\alpha_k^{-1}\Sigmabf_k^{\mathbf{y}}\right]^{-1}\mathbf{y}_{kn}^{} \mathrm{.}
\end{IEEEeqnarray*} 

We also propose a variant approach where $\mathbf{D}_k$ is the correlation matrix $\R_k^{\y} \triangleq \bm{\gamma}_k^\top \Sigmabf_k^{\y} \bm{\gamma}_k^{}$, with $\bm{\gamma}_k^{} \triangleq {\left( \eye_{d_k} \circ \Sigmabf_k^{\y} \right)}^{-\frac{1}{2}}$.
In this case, only correlations are estimated from the data while variances are fixed at $\alpha_k$.
The advantage of this choice is that it controls the scale of the sources rather than letting them be arbitrarily large/small. %
For this \emph{scale-controlled} case, we have:
\begin{IEEEeqnarray}{Lcl}
	\check{I}(\y) & \> = \> &
	- \sum_{m=1}^M J_{D_m} + \frac{1}{2}\sum_{k=1}^K J_{C_k}
	- f_{\left(K,\beta_k,\lambda_k,\eta_k,d_k,\nu_k\right)} \nonumber
	\\*
	&&
	- \sum_{k=1}^K \frac{\eta_k-1}{N}\sum_{n=1}^{N} J_{F_{kn}}
	+ \sum_{k=1}^K \frac{\lambda_k}{N}\sum_{n=1}^{N} J_{E_{kn}} \mathrm{,}
	\label{eq:MISAscobjective}
\end{IEEEeqnarray}

\begin{stretchpars}
\noindent where all terms are defined as in \eqref{eq:MISAobjective}, except
\end{stretchpars}

\noindent ${J_{C_k} = \ln\det\left(\bm{\gamma}_k^{} \Sigmabf_k^\y \bm{\gamma}_k^\top\right)}$, 
${J_{F_{kn}}  = \ln( \y_{kn}^\top {\left[ \bm{\gamma}_k^{}\Sigmabf_k^\y \bm{\gamma}_k^\top \right]}^{-1} \y_{kn}^{} )}$ and ${J_{E_{kn}} = {\left( \y_{kn}^\top {\left[ \bm{\gamma}_k^{}\Sigmabf_k^\y \bm{\gamma}_k^\top \right]}^{-1} \y_{kn}^{}\right)}^{\beta_k}}$,
with gradient:
\begin{IEEEeqnarray}{RCL}
	\nabla\check{I}(\W)_{mi_k} & \> = \> &
	\left[\bar{\bm{\gamma}}_{k}^{-1}B_k^{}+\left[\bar{\bm{\gamma}}_{k}^{}G_k^{}-B_k^{}A_k^\top \right. \right. \nonumber \\%
	&&  \left. \left. + \left[\mathbf{Z}^{-1}_{\Sigmabf}-\bar{\bm{\gamma}}_{k}^{2}\right] \right]\mathbf{Y}_k^{}\right]\X_m^\top - \left(\W_m^-\right)^\top \IEEEeqnarraynumspace \label{eq:MISAscgrad}
	\\
	&& \>\> _{\forall k \> \in \> \{1,\ldots,K\}, \>\> \forall m \> \in \> \{1,\ldots,M\}} \nonumber
\end{IEEEeqnarray}
\noindent where
\begin{IEEEeqnarray*}{RCL}   
	\bar{\bm{\gamma}}_{k}^{}  & \> = \> & \left(\eye\circ\mathbf{Z}_{\Sigmabf}^{}\right)^{-\frac{1}{2}}
	\\ \nonumber
	\mathbf{Z}_{\Sigmabf}^{}  & \> = \> & \PP_k^{}\W\X\X^{\top}\W^{\top}\PP_k^{\top} \\
	G_k^{}  & \> = \> & \I\circ\left(B_k^{}\mathbf{Y}_k^{\top}\right)\\
	B_k  & \> = \> & A_k \> \mathrm{diag}\>(\mathbf{t}_k)\\
	A_k^{}  & \> = \> & {\mathbf{Z}_{\Sigmabf}}^{-1}\bar{\bm{\gamma}}_{k}^{-1}\mathbf{Y}_k^{}\\
	\mathbf{t}_k^{} & \> = \> & \left(\frac{2\beta_k\lambda_k}{N}\mathbf{z}_k^{\beta_k}+\frac{2\left(1-\eta_k\right)}{N}\right)\circ\mathbf{z}_k^{-1}\\
	\mathbf{z}_k & \> = \> & \left[z_{k1},\>\ldots,\>z_{kn}, \>\ldots,\> z_{kN}\right]\\
	z_{kn}^{} & \> = \> & \y_{kn}^{\top}\left[\bm{\gamma}_k^{}\bar{\Sigmabf}_k^\y\bm{\gamma}_k^{\top} \right]^{-1}\y_{kn}^{} \mathrm{.}
	\label{eq:tk}
\end{IEEEeqnarray*}

While the equations presented above are general and support any choice of subspace specific parameters $\bm{\psi}_k$, in the examples presented here, we opted to use the same set $\bm{\psi}_k = \bm{\psi}_L$ for all subspaces, modeling subspaces as multivariate Laplace distributions \emph{with} correlation estimation.
The derivation of the gradients can be found in \emph{supplemental material} along with a description of the relative gradient update $\nabla \check{I}(\W) \W^{\top}\W$~\cite[Ch. 4]{ComonJutten2010Handbook}\cite{Cardoso1996_RelativeGrad} we used together with the \ac{L-BFGS-B}~\cite{ByrdRH1995_LBFGSB,ZhuC1997_LBFGSB_Alg} available in the non-linear constraint optimization function $\mathsf{fmincon}$ of MATLAB's Optimization Toolbox.
Nonlinear constraints such as those shown next can be easily incorporated in $\mathsf{fmincon}$'s interior-point barrier method~\cite[Ch. 19]{Nocedal2006_NumOpt}~\cite{Waltz2006_NLOpt_LS_TR_KNITRO}.

\subsection{Pseudoinverse Reconstruction Error}\label{sec:recmix}
In the overdetermined case, i.e., when $V_m > C_m$ and $\W$ is wide, it is necessary to constrain $\W$ in order to evade ill-conditioned solutions.
The error incurred by $\W$ in reconstructing the data samples can indirectly guide and constrain $\W$.
The \ac{MSE} between $\x$ and $\hat{\x}$ gives the following formulation of the \acf{RE}:
\begin{IEEEeqnarray}{CCl}
	E &=& %
	\mathbb{E} \left[ \norm{\hat{\x} - \x}_2^2 \right]
	\approx \frac{1}{N} \sum_{n=1}^{n=N} \norm{\hat{\x}_n - \x_n}_2^2 \mathrm{.}
	\label{eq:REWT}
\end{IEEEeqnarray}
Firstly, the optimal \emph{linear} estimator of $\x$ based on $\y$ for a system with estimation error $\e\text{'}$, such as $\y = \W\x + \e\text{'}$, is 
$\hat{\A}\y$, where $\hat{\A}$ is the minimizer of \ac{MSE}:
\begin{IEEEeqnarray}{RCl}
	\hat{\A} &=& \Sigmabf^{\x} \W^{\top} \left( \W\Sigmabf^{\x}\W^{\top} + \Sigmabf^{\e\text{'}} \right)^{-1} \mathrm{,}
	\label{eq:A_Wiener}
\end{IEEEeqnarray}
and $\Sigmabf^{\x}$ is the data covariance.
In the high \ac{SNR} regime, ${\mathrm{diag}\left( \W\Sigmabf^{\x}\W^{\top} \right) \gg \mathrm{diag}\left( \Sigmabf^{\e\text{'}} \right)}$ element-wise and, as discussed in~\cite{Haufe2014_BackToFWmodels_PLSreconstruction}, yields
\begin{IEEEeqnarray}{CCl}
	\hat{\A} = \Sigmabf^{\x} \W^{\top} \left( \W\Sigmabf^{\x}\W^{\top} \right)^{-1}%
	= \Sigmabf^{\x} \W^{\top} {\Sigmabf^{\y}}^{-1} \mathrm{.}
	\label{eq:A_Haufe}
\end{IEEEeqnarray}
This choice of $\hat{\A}$ \emph{always} minimizes the error no matter how far $\W$ is from the true $\W_\star$ and serves little as a constraint. %

Assuming unit source variances and data whitened such that $\Sigmabf^{\x} =  \mathbb{E \left[ \x\x^{\top} \right]} = \eye$, in \ac{ICA} problems $\W_\star$ must be \emph{row} orthonormal, i.e., $\W_\star \W^{\top}_\star = \eye$.
Our previous work~\cite{SilvaRF_MISA_ICIP} utilized $\hat{\A} = \W^{\top}$ to reconstruct $\x$ as $\hat{\x} = \W^{\top}\W\x$ instead.
Under the whitening assumption, this can be implemented in~\eqref{eq:REWT} as a soft regularizer provably equivalent to regularization by either the Frobenius norm $\frob{\W^{\top}\W - \eye}^2$ or $\frob{\W\W^{\top} - \eye}^2$, when the regularizer constant approaches infinity~\cite{LeQ_ReconstructErrorENIPS2011}.
Therefore, this approach effectively penalizes non-orthogonal $\W$. %

Here, our investigation of the \ac{SVD} of $\W$ reveals that, if the matrix has orthonormal rows, then its singular values are all $1$ and $\W = USV^{\top} = UV^{\top}$, where $S = \eye$, $U$ are the left singular vectors of $\W$, and $V$ its right singular vectors.
Therefore, $\W^{\top}\W = VU^{\top}UV^{\top} = VV^{\top}$.
Since $\W$ is wide, $V$ is tall, which implies $VV^{\top} \neq \eye$, in general.
Thus, using $\hat{\x}_n = \W^{\top}\W\x_n$, the \ac{RE} simplifies as:
\begin{IEEEeqnarray}{CCl}
	E_{\top} &\approx& \frac{1}{N} \sum_{n=1}^{n=N} \norm{(VV^{\top} - \eye)\x_n}_2^2 \mathrm{.}
	\label{eq:REWT_constraint}
\end{IEEEeqnarray}
This clearly shows that \ac{RE} with $\hat{\A} = \W^{\top}$ implicitly acts as a constraint on the right singular vectors of $\W$, selecting those whose outer product approximates the identity matrix $\eye$.

If not orthonormal, $\W^{\top}\W = VS^2V^{\top}$ since $S \ne \eye$.
Thus, we propose to use the pseudoinverse $\W^{-} = \W^{\top}(\W\W^{\top})^{-1}$ in lieu of $\W^{\top}$, with $\hat{\x}_n = \W^{-}\W\x_n$.
Then, this \ac{PRE} ($E_{-}$) also simplifies as~\eqref{eq:REWT_constraint}.
This result follows from the \ac{SVD} of the pseudoinverse ${\W^{-} = VS^{-1}U^{\top}}$ and $\W^{-}\W = VS^{-1}U^{\top}USV^{\top} = VV^{\top}$.
Unlike before, this formulation effectively constrains $V$ in the general case.
Note that since $\Sigmabf^{\x} = \eye$ in the case of white data, the optimal estimator~\eqref{eq:A_Haufe} simplifies to $\hat{\A} = \W^{\top} \left( \W\W^{\top} \right)^{-1} = \W^{-}$, i.e., the \emph{pseudoinverse gives the least error when the data is white} (if the \ac{SNR} is high), regardless of the values contained in $\W$.
Thus, for white data, we conclude that the \ac{RE} formulation ($E_{\top}$) is more appropriate than \ac{PRE} ($E_{-}$). %
Our experience, however, suggests that $\W$ is far more likely non-orthogonal in real noisy, non-white data, justifying our preference for $E_{-}$. 

Furthermore, we introduce a normalization term, dividing $E_{-}$ by $\x_{\mathrm{norm}}$, the average power in the data, and we get \emph{the proportion of power missed}:
\begin{IEEEeqnarray}{CCl}
	E &\approx& \frac{1}{\x_{\mathrm{norm}}} \frac{1}{N} \sum_{n=1}^{n=N} \norm{\W^{\top} \left(\W\W^{\top}\right)^{-1} \W\x_n - \x_n}_2^2 \mathrm{,} \IEEEeqnarraynumspace
	\label{eq:REPINV_normalized}
\end{IEEEeqnarray}
where $\x_{\mathrm{norm}} \approx \frac{1}{N} \sum_{n=1}^{n=N} \norm{\x_n}_2^2$.
Its gradient has the form:
\begin{IEEEeqnarray}{RCL}
	\nabla E(\W) & \> = \> &
	C - C\W^{-}\W%
	\label{eq:REPINV_norm_grad}
\end{IEEEeqnarray}
\noindent where  
\begin{IEEEeqnarray*}{RCL}
	C &=& \frac{2}{\x_{\mathrm{norm}}~N} \left[\W^{-}\right]^\top B \nonumber\\
	B &=& \X Z^\top + Z \X^\top \nonumber\\
	Z &=& \W^{-}\W\X - \X \nonumber \mathrm{.}
\end{IEEEeqnarray*}
Since $\X$ and $\W$ are block-diagonal, these operations can be computed separately on each dataset by replacing $\X$ with $\X_m$ and $\W$ with $\W_m$.
This can be used both as a data reduction approach or a nonlinear constraint for optimization.

Finally, in \ac{MDU} problems, when there is prior knowledge supporting \emph{linear} dependence (i.e., correlation) within subspaces, then one useful and popular approach is to use \emph{group} PCA projection to initialize all blocks of $\W$~\cite{GIFT2015_v4a}.
It works by performing a single data reduction step on datasets concatenated along the $V$ dimension.
We have investigated this approach in a separate work~\cite{2016Rachakonda_MemEffGroupPCA}, offering efficient algorithms to enable this procedure when the number of datasets is very large ($M > 10000$).
For comparison purposes, we also considered the use of \ac{gPCA} as an alternate initialization approach for $\W$ in our experiments.

\subsection{MISA with Greedy Permutations (SDM Case)}\label{sec:isaperm}
We present a greedy optimization approach to counter local minima resulting from arbitrary source permutations.
When these occur, the numerical optimization in Section~\ref{sec:METH:objfun} stops early, at the newly found local minimum.
At that point, we propose to check whether another permutation of sources would attain a lower objective value.
This entails two challenges: 1) given the combinatorial nature of the task, even mild numbers of sources lead to huge numbers of candidate permutations, and 2) when the optimization stops early, most sources are still mixed and there is not enough \emph{refinement} to establish which sources are dependent and belong in the same subspace.
The low refinement precludes the combinatorial problem since it hinders the ability to distinguish between dependent and independent sources in the first place.

Firstly, therefore, we propose to transform
the \acf{SDM} \ac{ISA} task into \acf{SDU} \ac{ICA}.
We do that by temporarily voiding and replacing \emph{all} subspaces ($d_k \ge 2$) with multiple sources (each $d_k = 1$), and then restarting the numerical optimization from the current $\W$ estimate (local minimum).
This pushes all sources towards being independent from each other.
However, dependent sources will only be \emph{as independent as possible} and will retain some of their dependence.
Partly motivated by~\cite{Szabo2012ISAbyICAPerm}, this approach secures enough refinement to distinguish among subspaces.
Thus, given sources that are as independent as possible, we propose a greedy search for any residual dependence among them.
The greedy solution is valid because the specific ordering within subspaces is irrelevant and it suffices to simply identify which sources go together.
Unlike~\cite{Szabo2012ISAbyICAPerm}, our approach does not require accessory objective functions to detect dependent sources.
Instead, it uses the same scale invariant objective defined in \eqref{eq:MISAobjective}.

The procedure is 1) switch to the \ac{ICA} model (effectively, make $\PP = \eye$), 2) numerically optimize it, 3) reassign sources into subspaces one at a time.
In the latter, as indicated in Algorithm~\ref{alg:GP} ($\GP$), each source is assigned sequentially to each subspace (if two or more are assigned to the same subspace, they are reassigned together thereafter).
Thus, the model changes with every assignment, and simple evaluation of the objective $\cost (\cdot)$ (without numerical optimization) produces a value for each particular assignment.
\emph{The scale invariant formulation ensures source variances do not influence the estimation}.
The assignment minimizing the objective function determines to which subspace a source belongs.
Here, assume that $k = K + 1$ inserts one more row in $\PP$ for a new subspace; $[:,p]$ are the contents of columns indexed by $p$ (conversely for rows); $\find(\cdot)$ recovers the indexes of all non-zero elements; $\rer(\PP)$ removes rows from $\PP$ containing only zero entries; $\eps$ is the machine's precision.
	\begin{algorithm}[!t]
		\small
		\removelatexerror
		\caption{Greedy Permutations ~~ $\GP$}
		\label{alg:GP}
		\begin{algorithmic}[1]
			\Require{dataset $\X \in \mathbb{R}^{V \times N}$, subspace assignment matrix $\PP \in \{0,1\}^{K \times C}$, unmixing matrix $\W \in \mathbb{R}^{C \times V}$}
			\State $K, C = \mdim(\PP)$
			\For{$c = 1$ \textbf{to} $C$} \Comment{loop over sources}
			\State $\kurrent = \find(\PP[:,c])$ \Comment{index of current  subspace}
			\State $p = \find(\PP[\kurrent,:])$ \Comment{source indices}
			\State $\PP[:,p] = 0$ \Comment{erase source assignments to subspace}
			\State $\vals = [~]$ \Comment{cost values array}
			\For{$k = 1$ \textbf{to} $K+1$} \Comment{loop over subspaces}
			\If{$k > 1$}
			\State $\PP[k-1,p] = 0$ \Comment{undo previous assignment}
			\EndIf
			\State $\PP[k,p]=1$ \Comment{assign sources to subspace $k$}
			\State $\PP_{nu} = \rer(\PP)$
			\State $\vals[k] = \cost(\X, \PP_{nu}, \W, \scc = \false)$ \NextLineComment{evaluate Equation~\eqref{eq:MISAobjective}}
			\EndFor
			\State $\PP[k,p] = 0$ \Comment{undo previous assignment}
			\State $k = \argmin(\vals)$ \Comment{assignment with lowest cost}
			\parIf{$k \ne \kurrent$ \myAnd \\ ${\abs{\vals[k] - \vals[\kurrent]} < \sqrt{\eps}}$}
			
			\State $k = \kurrent$ \Comment{ignore tiny change improvements}
			\EndparIf
			\State $\PP[k,p] = 1$
			\State $\PP = \rer(\PP)$
			\EndFor
			\State \textbf{return} $\PP$
		\end{algorithmic}
	\end{algorithm}

After repeating this procedure for all sources, in an attempt to solve the original model, we order the identified subspaces so as to match the original prescribed subspace structure $\PP$ as closely as possible.
This final sorting ($\match(\cdot)$) defines a specific permutation of the sources, which we then use to reorder the rows of the local minimum solution $\W$ for the original \ac{ISA} problem, effectively moving that solution out of the local minimum.
After that, we resume the numerical optimization of the original \ac{ISA} problem until another minimum is found.
In our experiments, repeating this procedure just twice in a row ($T=2$) and taking the best out of three solutions sufficed to drastically improve results.
In Algorithm~\ref{alg:MISAGP_SDM} ($\MISAGPSDM$), $\MISA(\cdot)$ represents the numerical optimization (Section~\ref{sec:METH:objfun}).

A direct benefit of this approach is that more dependence tends to be retained within subspaces as compared to~\cite{Szabo2012ISAbyICAPerm}, which is a desirable property because it leaves room for further post-processing and investigation.
Another advantage of our approach is that it can match source assignments to user-prescribed subspace priors ($\PP$) when they are available.
\begin{algorithm}[!t]
	\small
	\caption{MISA-GP for SDM Problems $\MISAGPSDM$} \label{alg:MISAGP_SDM}
	\begin{algorithmic}[1]
		\Require{dataset $\X \in \mathbb{R}^{V \times N}$, user-defined (UD) subspace assignment matrix $\PP_{\mathrm{UD}} \in \{0,1\}^{K \times C}$, initial unmixing matrix $\W_0 \in \mathbb{R}^{C \times V}$}, maximum number of greedy iterations $T$
		\State $\W = \MISA(\X, \PP_{\mathrm{UD}}, \W_0, \scc = \true)$ \NextLineComment{minimize Equation~\eqref{eq:MISAscobjective}}
		\State $\vals[0] = \cost(\X, \PP_{\mathrm{UD}}, \W, \scc = \true)$
		\State $\W_{\mathrm{opt}}[0] = \W$; ~~$t=1$; ~~$\vals[t]=\mathrm{Inf}$
		\While{$t \le T$ \textbf{and} $\vals[t] \ne \vals[t-1]$} %
		\State $\PP = \eye$ \Comment{switch to \ac{SDU} model}
		\State $\W_{\mathrm{SDU}} = \MISA(\X, \PP, \W, \scc = \true)$ \NextLineComment{solve \ac{SDU}}
		\State $\PP = \GP(\X,\PP,\W_{\mathrm{SDU}})$ \Comment{Algorithm~\ref{alg:GP}}
		\parState{$\ix = \match(\PP,\PP_{\mathrm{UD}})$}{\Comment{find source ordering best}}
		\NextLineComment{{matching prescribed $\PP_{\mathrm{UD}}$}}
		\State $\W = \W[\ix,:]$ \Comment{reorder sources (escape local min)}
		\State $\W = \MISA(\X, \PP_{\mathrm{UD}}, \W, \scc = \true)$ \NextLineComment{restart \ac{SDM}}
		\State $\vals[t] = \cost(\X, \PP_{\mathrm{UD}}, \W, \scc = \true)$
		\State $\W_{\mathrm{opt}}[t] = \W$
		\State $t=t+1$
		\EndWhile
		\State $t = \argmin(\vals)$ \Comment{find best solution}
		\State $\mathbf{return}$ $\W_{\mathrm{opt}}[t]$
	\end{algorithmic}
\end{algorithm}

\subsection{MISA with Greedy Permutations (MDM Case)}\label{sec:misaperm}
The previous approach addresses cross-subspace interference issues due to incorrect allocation of the \emph{sources} and, therefore is appropriate for  \ac{SDM} problems.
However, it is not sufficient to perform such procedure in \ac{MDM} problems since ambiguities may also occur at the \emph{subspace} level, i.e., incorrect allocation of the dataset-specific subspaces.

Consider the following example for a model with three subspaces spanning two datasets, each dataset containing five sources.
Assume the correct assignment of sources is as follows:
$p_1(y_{11},y_{21},y_{22})
p_2(y_{12},y_{13},y_{23})
p_3(y_{14},y_{15},y_{24},y_{25})$,
where the notation $y_{mi}$ refers to source $i$ from dataset $m$, and $p_k(\cdot)$ is the joint \ac{pdf} of subspace $k$.
Since $\MISAGPSDM$ is designed for single datasets, at best, it produces
$p_1(y_{11})
p_2(y_{12},y_{13})
p_3(y_{14},y_{15})$
for $m=1$ and
$p_1(y_{21},y_{22})
p_2(y_{23})
p_3(y_{24},y_{25})$
for $m=2$.
Then, from a global perspective, these solutions would yield the correct subspace assignment above, thus solving the \ac{MDM} problem.
However, it is equally acceptable for \ac{SDM} solvers to produce either
$p_1(y_{11})
p_2(y_{1\mathbf{4}},y_{1\mathbf{5}})
p_3(y_{1\mathbf{2}},y_{1\mathbf{3}})$
for $m=1$ or
$p_1(y_{2\mathbf{4}},y_{2\mathbf{5}})
p_2(y_{23})
p_3(y_{2\mathbf{1}},y_{2\mathbf{2}})$
for $m=2$ if the datasets are evaluated separately (notice the bold subscripts).
Together they imply
$p_1(y_{11},y_{2\mathbf{4}},y_{2\mathbf{5}})
p_2(y_{1\mathbf{4}},y_{1\mathbf{5}},y_{23})
p_3(y_{1\mathbf{2}},y_{1\mathbf{3}},y_{2\mathbf{1}},y_{2\mathbf{2}})$,
which does not match the correct assignment and, thus, fails to produce a solution for the \ac{MDM} problem.
What we have illustrated here is that within-dataset permutations of  \emph{equal-sized} subspaces may induce mismatches across datasets if the datasets are processed separately.
Another complicating factor are subspaces absent from a particular dataset.

\begin{algorithm}[!t]
	\small
	\caption{MISA-GP for MDM Problems ~~ $\MISAGP$} \label{alg:MISAGP}
	\begin{algorithmic}[1]
		\Require{dataset $\X = \{\X_m \in \mathbb{R}^{V_m \times N} : m \in M\}$, user-defined subspace assignment matrices $\PP_{\mathrm{UD}} = \{\PP_{\mathrm{UD},m} \in \{0,1\}^{K \times C_m} : m \in M\}$, initial unmixing matrix $\W_0 = \{\W_{0,m} \in \mathbb{R}^{C_m \times V_m} : m \in M\}$}, maximum number of greedy iterations $T$
		\State $\W = \MISA(\X, \PP_{\mathrm{UD}}, \W_0, \scc = \true)$ \NextLineComment{minimize Equation~\eqref{eq:MISAscobjective}}
		\State $\vals[0] = \cost(\X, \PP_{\mathrm{UD}}, \W, \scc = \true)$
		\State $\W_{\mathrm{opt}}[0] = \W$; ~~$t=1$; ~~$\vals[t]=\mathrm{Inf}$
		\While{$t \le T$ \textbf{and} $\vals[t] \ne \vals[t-1]$} %
		\For{$m = 1$ \textbf{to} $M$} \Comment{For each dataset}
		\State $\PP_m = \eye$ \Comment{switch to \ac{SDU} model}
		\State $\W_{\mathrm{SDU}} = \MISA(\X_m, \PP_m, \W_m, \scc = \true)$ \NextLineComment{solve \ac{SDU}}
		\State $\PP_m = \GP(\X_m,\PP_m,\W_{\mathrm{SDU}})$ \Comment{Algorithm~\ref{alg:GP}}
		\parState{$\ix = \match(\PP_m,\PP_{\mathrm{UD},m})$}{\Comment{find source ordering}}
		\NextLineComment{{best matching prescribed $\PP_{\mathrm{UD},m}$}}
		\parState{$\W_m = \W_m[\ix,:]$}{\Comment{reorder sources (escape local min)}}
		\EndFor
		\parState{$\W = \osp(\X, \PP_{\mathrm{UD}}, \W,$ $\scc = \false)$}{}
		\State $\W = \MISA(\X, \PP_{\mathrm{UD}}, \W, \scc = \true)$ \NextLineComment{restart \ac{MDM}}
		\State $\vals[t] = \cost(\X, \PP_{\mathrm{UD}}, \W, \scc = \false)$
		\State $\W_{\mathrm{opt}}[t] = \W$
		\State $t=t+1$
		\EndWhile
		\State $t = \argmin(\vals)$ \Comment{find best solution}
		\State $\mathbf{return}$ $\W_{\mathrm{opt}}[t]$
	\end{algorithmic}
\end{algorithm}

Borrowing from the ideas in Section~\eqref{sec:isaperm}, we propose three approaches to address these issues.
The first, extends the greedy search to \emph{all} datasets by sequentially assigning each source (in every dataset) to every subspace and accepting the assignments that reduce the objective function.
This would yield a complexity of at least $O(\bar{C}K)$, and $O(\bar{C}^2)$ in the (unlikely) worst case of $K = \bar{C}$.
The second, processes each dataset separately (as in the previous example) and then applies the same greedy strategy at the level of subspaces instead.
Effectively, this approach cycles through each subspace sequentially, trying to determine which of them can be combined to form a larger subspace.
This yields a complexity of $O(C_m K M) + O(K^2 M)$.
The final approach is to test all possible permutations of subspaces with the same size, after processing each dataset separately, which yields $O((K!)^M)$.
While this can quickly become computationally prohibitive, it can also identify better solutions since it evaluates all subspace permutations of interest.
In this work, we elected to use the third approach when the number of sources is small and the second when that number becomes larger ($\osp(\cdot)$).
Full procedures are indicated in Algorithm~\ref{alg:MISAGP} ($\MISAGP$).

\section{Results}\label{ch:exp}
We present results on multiple experiments that follow the principles outlined in~\cite{SilvaRF_MMSimNIMG2014}, including a summary of various controlled simulations on carefully crafted synthetic data, as well as hybrid data and comparisons with several algorithms.
\subsection{General Simulation Setup and Evaluation} \label{sec:simsetup}
In the following, we consider the problem of identifying statistically independent subspaces.
Thus, in all experiments, each subspace $\y_k$ is a random sample with $N$ observations from Laplace distribution. %
Subspace observations are linearly mixed via a random $\A$ as $\x = \A\y + \e$, where $\e$ is additive sensor white noise. %
The condition number $c_m$ of block $\A_m$ (the ratio between its largest and smallest singular values) is prescribed as follows.
First generate a $V_m \times C_m$ random Gaussian sample $\A_m$ with zero mean and unit variance.
By \ac{SVD}, $\A_m = USV^\top$, with $\sigma_{\max} = \max(S)$ and $\sigma_{\min} = \min(S)$.
Finally, new singular values are defined as $\bar{S} = S + \frac{\sigma_{\max} - c_m \sigma_{\min}}{c_m-1}$,
and the resulting $\A = U\bar{S}V^\top$ will have $\textrm{cond}(\A) = c_m$.

The white Gaussian noise $\e$ (zero mean and unit variance) is scaled by a value $a$ in order to attain a prescribed \ac{SNR}.
The \ac{SNR} is the power ratio between the noisy signal $\x$ and the noise $\e$.
The average power of $\x$ is ${P_{\x} = \mathbb{E}\left[ \x^\top \x \right]}$.
Substituting $\x = \A\y + a\e$, and using the identities $\mathbb{E}\left[ \z^\top \z \right] = \mathbb{E}\left[ \mathrm{tr}\left( \z \z^\top \right) \right] = \mathrm{tr}\left( \mathbb{E}\left[ \z \z^\top \right] \right)$, ${\mathbb{E}\left[ \e \e^\top \right] = \eye_{V_m}}$, and $\mathbb{E}\left[ \y \y^\top \right] = \eye_{C_m}$, it is easy to show that $P_{\e} = a^2 V_m$, and $P_{\x} = \mathrm{tr}\left( \A\A^\top \right) + a^2 V_m$, where $\mathrm{tr}(\cdot)$ is the trace operator.
Their ratio gives the SNR.
Then, %
$a = \sqrt{\frac{\mathrm{tr}\left( \A\A^\top \right)}{V_m \left(\mathrm{SNR}-1\right)}}$,
which we use to prescribe a SNR to synthetic data.
The equality $\mathrm{SNR} = 10^{\frac{\mathrm{SNRdB}}{10}}$ permits decibel (dB) specifications.

The quality of results is evaluated using the normalized \ac{MISI}~\eqref{eq:MISI}, which extends the ISI~\cite{RN44,Macchi1995} to multiple datasets.

{{
\small
\begin{IEEEeqnarray}{rCl}
\mathrm{MISI}(\mathbf{H}) &=& \frac{0.5}{K(K-1)}%
	\left[\sum_{i=1}^{K}\left(-1+\sum_{j=1}^{K}\frac{\left|h_{ij} \right|}{\mathop{\max }\limits_{k} \left|h_{ik} \right|}  \right) \right. %
	\nonumber%
	\\%
	&& \qquad \quad ~~~%
	\left. +\sum_{j=1}^{K}\left(-1+\sum _{i=1}^{K}\frac{\left|h_{ij} \right|}{\mathop{\max }\limits_{k} \left|h_{kj} \right|}  \right) \right] ~~~~%
\label{eq:MISI} 
\end{IEEEeqnarray}
}}%
where $\mathbf{H}$ is a matrix with elements $h_{ij} = \mathbf{1}^\top \abs{\PP_i \hat{\W} \A \PP_j^\top} \mathbf{1}$, with $(i,j) = 1 \ldots K$, i.e., the sum of absolute values from all elements of the interference matrix $\hat{\W} \A$ corresponding to subspaces $i$ and $j$, and $\hat{\W}$ is the solution being evaluated.

For fairness, all algorithms are initialized with the same $\W_0$.
See optimization parameters in supplemental material.

\subsection{Summary of Synthetic Data Simulations}\label{sec:synthdata}
The performance \ac{MISA} in a series of synthetic data experiments with different properties is summarized below (Table~\ref{tab:results_sim}).
Complete details are available as supplemental material online.
\subsubsection{ICA 1 ($\bar{V}>N$)}\label{sec:ICA1}
effects of additive noise (a) and condition number (b) are assessed in a moderately large ICA problem ($\bar{C}=75$, $M=1$) with rectangular mixing matrix $\A$ ($\bar{V}=8000$) at a fairly small sample size regime ($N=3500$).
Under low \ac{SNR} (b), MISA outperforms Infomax when ${\cond{\A} \ne 1}$. At high SNR (a), MISA outperforms Infomax more often than not.

\subsubsection{IVA 1 ($V_m<N$, $V_m=C_m$)}
MISA performance is assesed in an \ac{IVA} problem, in which subspaces span all of $M=10$ datasets but all sources are unidimensional in each dataset. Specifically, we study the case when no data reduction is required (i.e., $V_m=C_m=16$), noise is absent, and observations are abundant ($N=32968$) (c).
The striking feature observed here is that the performance of IVA-GL~\cite{Anderson2012_JBSSGaussianAlgPerf} is much more variable than that from MISA, especially with high correlation within the subspaces.
MISA performs well even at low within-subspace correlation levels and is highly stable when these correlations are larger than 0.2.

\subsubsection{IVA 2 ($V_m<N$)}
Effects of additive noise (a) and condition number (b) are assessed in a larger IVA problem ($C_m = 75$, $M = 16$) with rectangular mixing matrix $\A$ ($V_m = 250$) and an abundant number of observations $N=32968$.
Data reduction with either \acf{gPCA} or \acf{PRE} produced equivalent results in this large $N$ scenario. Under low SNR, increasing the condition number had a fairly small detrimental effect on the performance of both IVA-L~\cite{KimT_IVA2006} (modified for backtracking, see supplement) and MISA. More importantly, while both IVA-L and \ac{MISA} performed very well at mild-to-high SNR levels, the performance of MISA on extremely noisy scenarios ($\mathrm{SNRdB} = 0.0043$) is remarkable ($0.1< \mathrm{MISI} < 0.01$), irrespective of using \ac{PRE} or \ac{gPCA}.

\begin{table}[!t]
	\centering
	\renewcommand{\arraystretch}{.8}
	\setlength{\tabcolsep}{4.5pt}
	\caption{Summary of simulation results.
		$\mathrm{\mathbf{(a, b)}}$ Median (over 10 dataset instances) of best  MISI (over 10 initializations per dataset).
		$\mathrm{\mathbf{(c, d)}}$ Median MISI (over 10 initializations, 1 dataset instance).}
	\label{tab:results_sim}
	\captionsetup[subfloat]{justification=centering}
	\subfloat[Varying SNRdB, Fixed cond($\A$) $=7$]{
		\begin{tabular}{llccccc}
			\toprule
			SNRdB & & 30 & 10 & 3 & 0.46 & 0.0043 \\ 
			\toprule
			ICA1 & PRE+Infomax & 0.0222 & 0.0292 & 0.0685 & \textbf{0.0928} & 0.2576 \\ 
			& PRE+MISA & \textbf{0.0145} & \textbf{0.0165} & \textbf{0.0261} & 0.1932 & 0.2743 \\ 
			\midrule
			IVA2 & PRE+IVA-L & 0.0111 & 0.0136 & 0.0197 & 0.0277 & 0.5158 \\ 
			& PRE+MISA & 0.0059 & 0.0088 & 0.0113 & 0.0151 & 0.0338 \\ 
			& gPCA+IVA-L & 0.0081 & 0.0090 & 0.0095 & 0.0094 & 0.1271 \\ 
			& gPCA+MISA & \textbf{0.0044} & \textbf{0.0045} & \textbf{0.0049} & \textbf{0.0065} & \textbf{0.0205} \\ 
			\midrule
			ISA3 & PRE+JBD-SOS & 0.2700 & 0.2804 & 0.2996 & 0.3255 & 0.3712 \\ 
			& PRE+MISA & 0.1153 & 0.1275 & 0.1320 & 0.1495 & 0.3404 \\ 
			& PRE+MISA-GP & \textbf{0.0366} & \textbf{0.0670} & \textbf{0.0794} & \textbf{0.1140} & \textbf{0.3404} \\ 
			\bottomrule
		\end{tabular}%
		\label{tab:results_SNR}}
	\\
	\subfloat[Fixed SNRdB $=3$, Varying cond($\A$)]
	{\begin{tabular}{llcccc}
			\toprule
			cond($\A$) & & 1 & 3 & 7 & 15 \\ 
			\toprule
			ICA1 & PRE+Infomax & \textbf{0.0804} & 0.0188 & 0.0216 & 0.0493 \\ 
			& PRE+MISA & 0.1934 & \textbf{0.0148} & \textbf{0.0161} & \textbf{0.0267} \\ 
			\midrule
			IVA2 & PRE+IVA-L & 0.1923 & 0.1013 & 0.0749 & 0.0505 \\ 
			& PRE+MISA & \textbf{0.0052} & \textbf{0.0045} & \textbf{0.0049} & 0.0078 \\ 
			& gPCA+IVA-L & 0.0090 & 0.0086 & 0.0092 & 0.0095 \\ 
			& gPCA+MISA & \textbf{0.0052} & \textbf{0.0045} & \textbf{0.0049} & \textbf{0.0067} \\ 
			\midrule
			ISA3 & PRE+JBD-SOS & 0.2905 & 0.2792 & 0.2815 & 0.2962 \\ 
			& PRE+MISA & 0.1008 & 0.1065 & 0.1202 & 0.1351 \\ 
			& PRE+MISA-GP & \textbf{0.0395} & \textbf{0.0330} & \textbf{0.0612} & \textbf{0.0743} \\ 
			\bottomrule
		\end{tabular}%
		\label{tab:results_Acond}}
	\\
	\subfloat[IVA1: Increasing max. subspace correlation $\rho_{k,max}$]
	{\begin{tabular}{lcccccc}
			\toprule
			$\rho_{k,max}$ & 0 & 0.1 & 0.23 & 0.39 & 0.5 & 0.65 \\
			\toprule
			IVA-GL & 0.4767 & 0.0361 & 0.0114 & 0.0199 & 0.0184 & 0.0186 \\
			MISA & \textbf{0.0273} & \textbf{0.0098} & \textbf{0.0072} & \textbf{0.0062} & \textbf{0.0061} & \textbf{0.0049} \\
			\bottomrule
		\end{tabular}%
		\label{tab:results_IVA1}}
	\\
	\subfloat[Varying vs Fixed subspace dimensionality $d_k$]
	{\begin{tabular}{lcccc}
			\toprule
			& \multicolumn{2}{c}{ISA1 ($\rho_k = 0$)} & \multicolumn{2}{c}{ISA2 ($\rho_k > 0.2$)}  \\
			\cmidrule(l{1pt}r{1pt}){2-3} \cmidrule(l{1pt}r{1pt}){4-5}
			& $d_k = k$ & $d_k = 4$ & $d_k = k$ & $d_k = 4$ \\
			\toprule
			EST\_ISA & -- &.0.7557 & -- & 0.7766 \\
			JBD-SOS & 0.2600 & 0.3496 & 0.2826 & 0.3739 \\
			MISA & \textbf{0.0239} & \textbf{0.0162} & \textbf{0.0369} & \textbf{0.0326} \\
			\bottomrule
		\end{tabular}%
		\label{tab:results_ISA1_2}}
	\small
\end{table}

\subsubsection{ISA 1 and 2 ($\bar{V}<N$, $\bar{V}=\bar{C}$)}\label{subsubsec:ISA}
MISA performance is assessed in \ac{ISA} problems, in which subspaces are multidimensional, with $M=1$. Specifically, we study the case when no data reduction is required (i.e., $\bar{V}=\bar{C}=28$), noise is absent, and the number of observations $N$ is abundant (d).
Fixed and varying configurations of $K=7$ subspaces are considered, at two subspace correlation $\rho_k$ settings.
The striking feature observed here is that the performance of both JBD-SOS~\cite{LahatD2012_MICA_SOS} and EST\_ISA~\cite{LeQ2011_ISA} is very poor in all cases, even when within-subspace correlations are present.
MISA-GP is the only method with good performance, highlighting the large benefit of  our approach for evasion of local minima.

\subsubsection{ISA 3 ($\bar{V}>N$)}
Effects of additive noise (a) and condition number (b) are assessed in a mildly large ISA problem ($\bar{C}=51$, $M=1$) with variable subspace dimensionalities $d_k$, rectangular mixing matrix $\A$ ($\bar{V}=8000$) at a fairly small sample size regime ($N=5250$).
Under a challenging \ac{SNR}, JBD-SOS and MISA fail in virtually all cases ($\mathrm{MISI} > 0.1$). Inclusion of combinatorial optimization enables MISA-GP to perform quite well at mild-to-high SNR levels ($\mathrm{SNRdB} \ge 3$).

\subsection{Hybrid Data Experiments}\label{sec:hybdata}
We present three major results on novel applications of \ac{BSS} to brain image analysis, open sourcing realistic hybrid data standards (\href{https://github.com/rsilva8/MISA}{\url{https://github.com/rsilva8/MISA}}) that test estimation limits at small sample size.
The first pushes the conditions of experiment ICA 1 and emulates a single-subject temporal ICA of functional MRI (fMRI).
The second investigates the use of IVA with $V_m > N$ for multimodal fusion of brain MRI-derived data.
Finally, the last experiment evaluates the value of MDM models without data reduction for fusion of \ac{fMRI} and \acs{EEG} neural signals.

\subsubsection{Single-Subject Temporal \ac{ICA} of fMRI}\label{subsec:ssdata}
Here we consider temporal ICA of fast acquisition fMRI.
The dimensionality of the data is $\bar{V} = \mathrm{voxels} \approx 60\mathrm{k}$ and $N = \mathrm{time~points} \approx 1300$.
In order to better assess the performance of MISA in a realistic scenario, we propose to set the mixing matrix $\A$ as the \emph{real} part of the data.
First, we let $\bar{C}=20$ sources.
Then, $\A$ must be a $60\mathrm{k} \times 20$ matrix.
In order to have it correspond to real data, we assign to it the first twenty well-established aggregate spatial maps (3D volumes) published in~\cite{Allen2011baseline}.

\begin{figure}[!t]
	\centering
	\includegraphics[height=5cm]{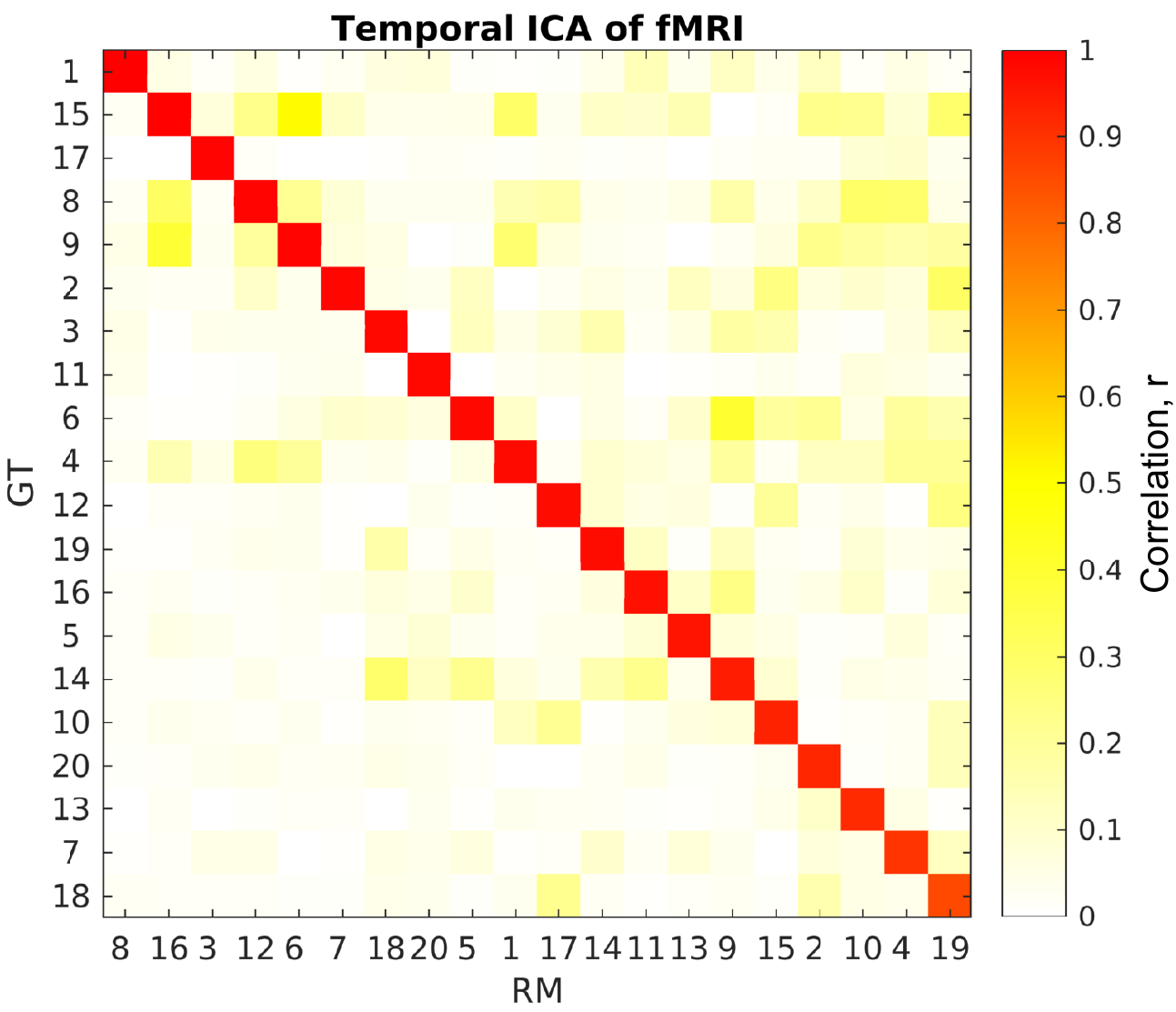}
	\caption{\textbf{Correlation with the ground-truth (hybrid temporal ICA)}. The correlation between the spatial map estimates from \ac{MISA} with \ac{PRE} (RM) and the ground-truth (GT) is very high with little residual similarity across sources, suggesting the analysis was successful.}
	\label{fig:hybICAtemporal_corr}
\end{figure}
For the synthetic part of the data, we propose to simulate a $20 \times 1334$ matrix of timecourses $\y$ by generating realistic autocorrelated samples that mimic observed fMRI timecourses to a good extent.
Sampling 20 such timecourses that retain independence with respect to each other is challenging because independently sampled autocorrelated time series tend to be correlated with one another.
Building on the simulation principles outlined in~\cite{SilvaRF_MMSimNIMG2014}, we seek to avoid randomly correlated timecourses (sources) in order to prevent mismatches to the underlying ICA model we wish to test.
In the same spirit, we also wish to have sources sampled from the same distribution used in the model, here a Laplace distribution.
We developed the following steps in order to meet all these requirements:
\begin{figure}[!t]
	\centering
	\includegraphics[width=\linewidth]{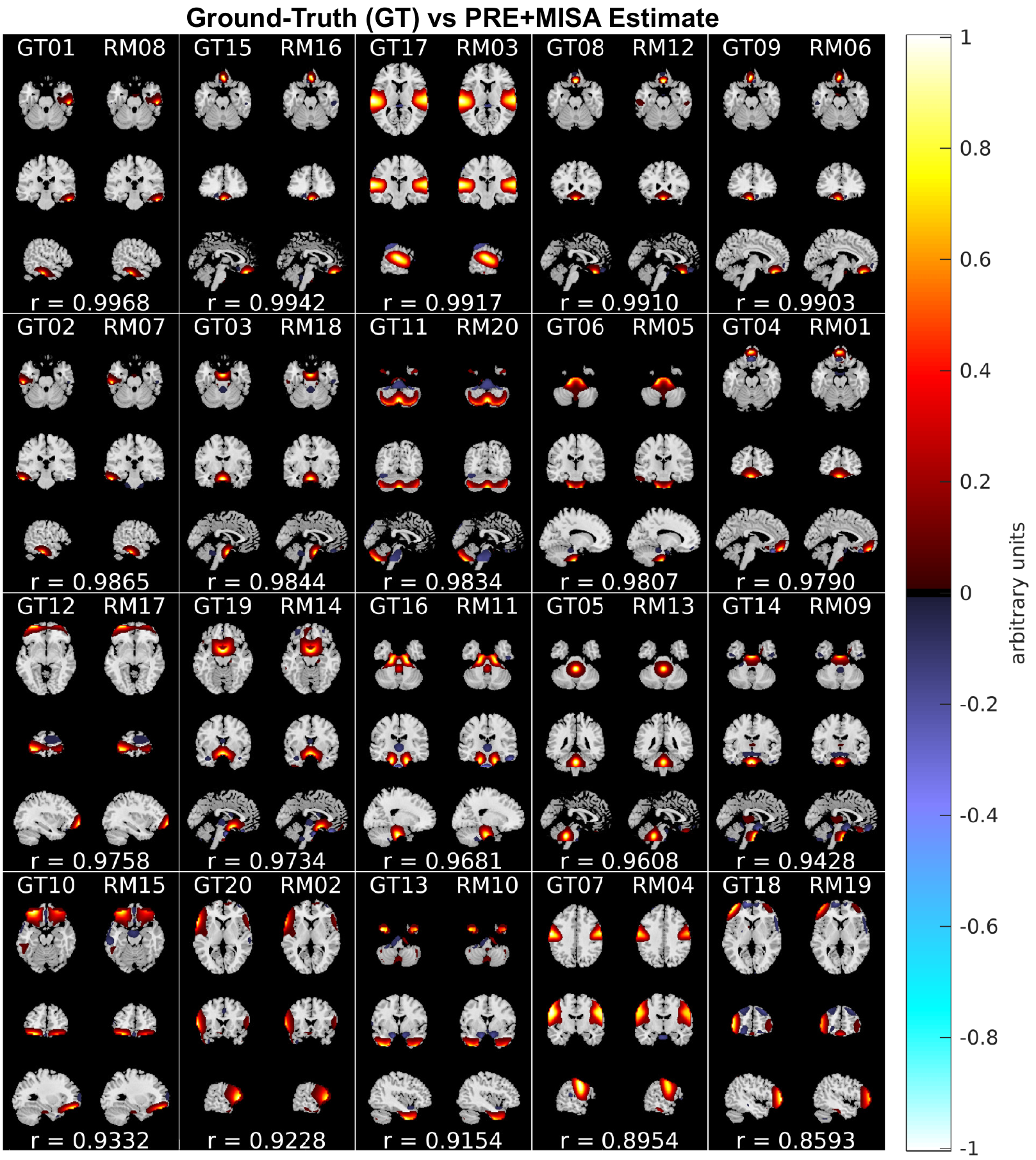}
	\caption{\textbf{Side-by-side comparison with the ground-truth (hybrid temporal ICA)}. The clear resemblance to the ground-truth maps suggests a successful recovery of the mixing matrix $\A$. The sample correlation $r$ is shown below each matched pair. Maps are sorted from highest to lowest correlation.}
	\label{fig:hybICAtemporal_maps}
\end{figure}
\begin{IEEEdescription}[\IEEEsetlabelwidth{0)}]
	\item[1)] Design a joint \emph{autocorrelation} matrix $\R^{\y\y}$ for all sources.  For the example above, this means a $\bar{C}N \times \bar{C}N$ block-diagonal correlation matrix ($\bar{C}N = 26680$) with $\bar{C}$ blocks of size $N \times N$.
	Each block is designed with an exponentially decaying autocorrelation function such that the autocorrelation between timepoint $n$ and $n-1$ is in the order of 0.85, and between $n$ and $n-10$ is in the order of 0.2. 
	This structure retains autocorrelation within each $N$-long section of an observation while retaining uncorrelation/independence among sections.
	\item[2)] Generate $50\mathrm{k}$ $\bar{C}N$-dimensional observations using a Gaussian copula~\cite{RN2609} and the autocorrelation matrix $\R^{\y\y}$ from step 1.
	Using copulas enables transformation of the marginal distributions while retaining their correlation/dependence.
	\item[3)] For each of the $50\mathrm{k}$ copula-sampled observations, transform the sample into a Laplace distribution.
	\item[4)] For each of the $50\mathrm{k}$ transformed $\bar{C}N$-dimensional observations, reshape them into a $\bar{C} \times N$ matrix and compute the resulting $\bar{C} \times \bar{C}$ $\R^\y$ correlation matrix.
	\item[5)] Compute the median correlation matrix $\R_{med}^\y$ over the $50\mathrm{k}$ observed $\R^\y$.
	\item[6)] Retain the transformed observation whose $\R^\y$ is closest to $\R_{med}^\y$ and reject the rest.
\end{IEEEdescription}
This type of rejection sampling effectively produces the desired outcome. %
Finally, Gaussian noise is added to the mixture for a \emph{low} $\mathrm{SNRdB} = 3$.
The condition number of $\A$ was 4.59.

In the results, the data was reduced using \ac{PRE} and then processed with MISA to obtain independent \emph{timecourses}.
The correlation between ground-truth (GT) and PRE+MISA spatial map estimates (RM) is presented in Fig.~\ref{fig:hybICAtemporal_corr}, and the spatial maps (estimating $\A$ from $\hat{\W}^{-}$) in Fig.~\ref{fig:hybICAtemporal_maps}. MISI $= 0.0365$.

\subsubsection{Multimodal IVA of sMRI, fMRI, and FA}\label{subsubsec:MMIVA}
In this multimodal fusion of \ac{sMRI}, \ac{fMRI}, and Fractional Anisotropy (FA) diffusion MRI data, the dimensionalities are $V_1 = \mathrm{voxels} \approx 300\mathrm{k}$, $V_2 = \mathrm{voxels} \approx 67\mathrm{k}$, $V_3 = \mathrm{voxels} \approx 15\mathrm{k}$, respectively, and 
$N = \mathrm{subjects} = 600$ (each modality measured on the same subject).
We pursue a hybrid setting where only the mixing matrices $\A_m$ are taken from real datasets to overcome typically small $N$ in patient population studies.
First, we let $C_m=20$ sources in each dataset.
Then, $\A_1$, $\A_2$, and $\A_3$ must be $300\mathrm{k} \times 20$, $67\mathrm{k} \times 20$, and $15\mathrm{k} \times 20$, respectively. To each, we assign the first twenty aggregate 3D spatial maps published in~\cite{judith2012},~\cite{Allen2011baseline},~\cite{HBM:HBM22945}, respectively.

\begin{figure}[!t]
	\centering
	\captionsetup[subfloat]{justification=centering}
	\subfloat[Structural MRI]{\vtop{\vskip -9pt \hbox{\includegraphics[height=3cm,trim={0 0 2cm 0},clip]{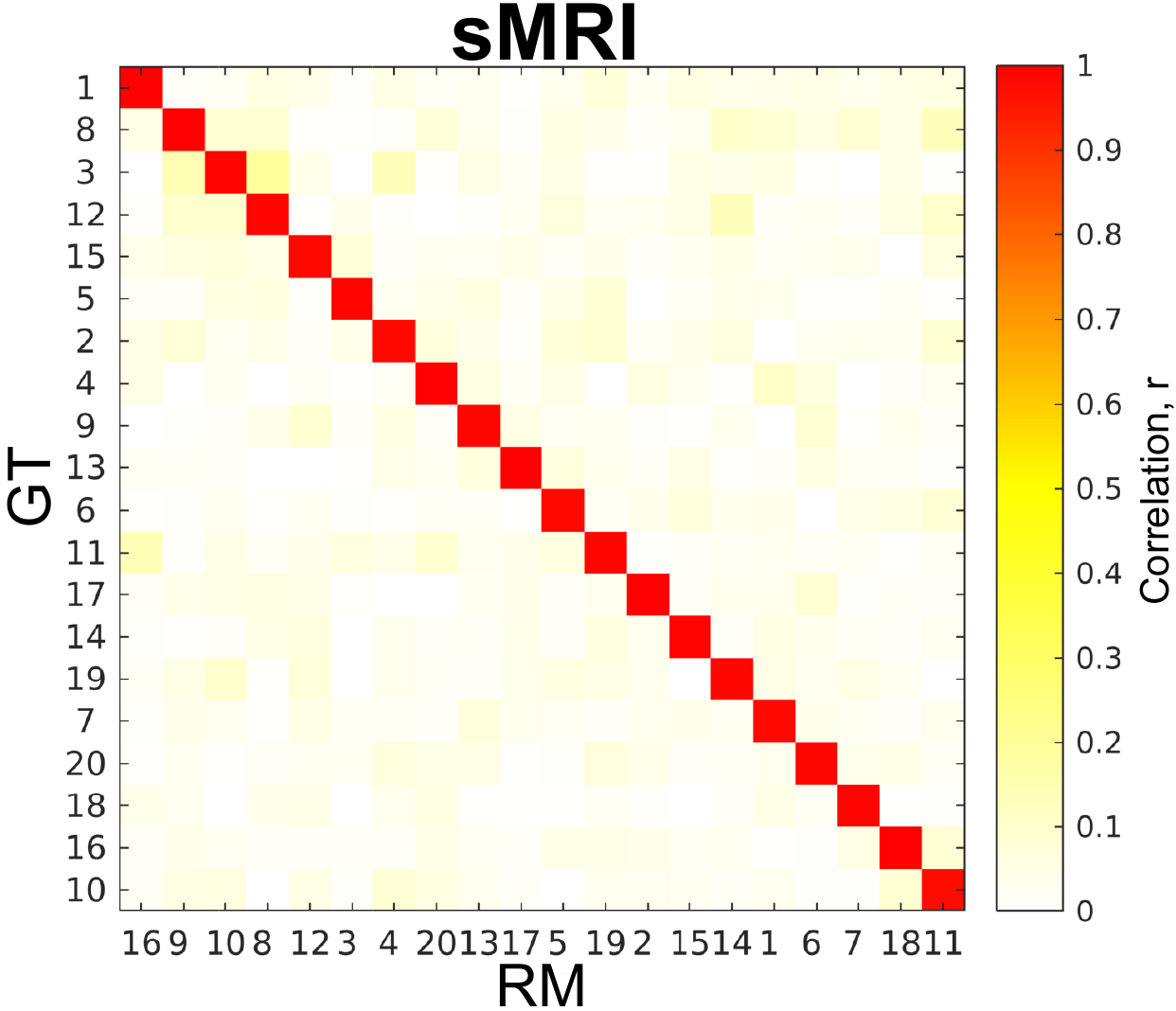}}}%
		\label{fig:IVAcorr1}}
	\hfil
	\subfloat[Functional MRI]{\vtop{\vskip -9pt \hbox{\includegraphics[height=3cm,trim={1cm 0 2cm 0},clip]{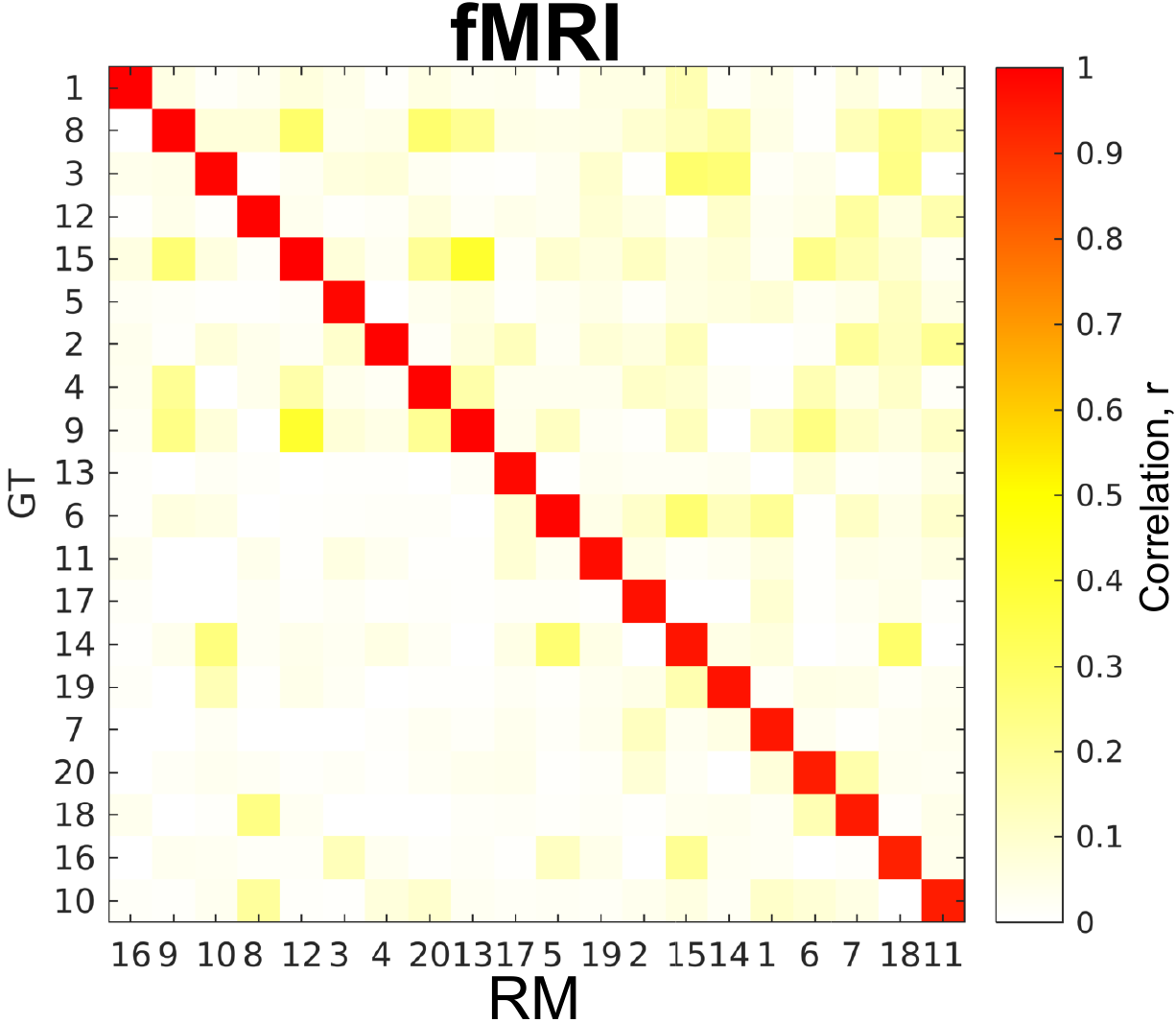}}}%
		\label{fig:IVAcorr2}}
	\hfil
	\subfloat[Fractional Anisotropy]{\vtop{\vskip -9pt \hbox{\includegraphics[height=3cm,trim={1cm 0 0 0},clip]{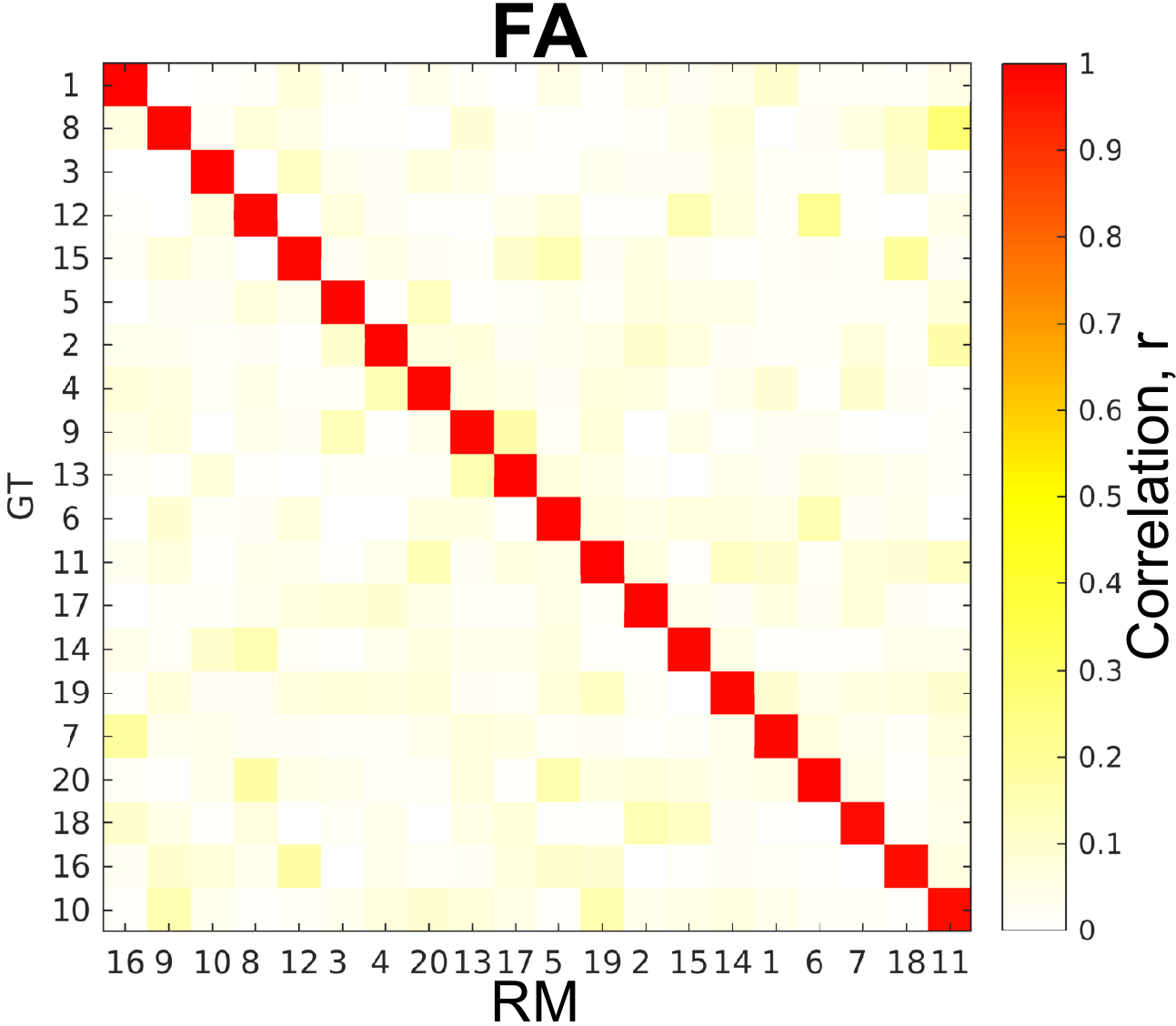}}}%
		\label{fig:IVAcorr3}}
	\caption{\textbf{Correlation with the ground-truth (multimodal IVA).} The correlation between the spatial map estimates from \ac{MISA} with \ac{PRE} (RM) and the ground-truth (GT) is very high in all modalities, with little residual similarity across sources, suggesting the analysis was successful.}
	\label{fig:hybMMIVAcorr}
\end{figure}

For the simulated part of the data, we generate three $20 \times 600$ matrices of subject expression levels $\y$.
$K=20$ subspaces, each with $d_k=3$ and $N = 600$ observations, were sampled independently from a Gaussian copula, using an inverse exponential autocorrelation function with maximal correlation varying from $0.65$ to $0.85$ for each subspace.
These were transformed to Laplace distribution marginals (\emph{not} multivariate Laplace) so as to induce a \emph{controlled mismatch} between the data (only \ac{SOS} dependence) and the model subspace distributions (multivariate Laplace---all-order dependence).
Finally, Gaussian noise was added separately in each dataset for a low $\mathrm{SNRdB} = 3$.
The condition numbers of $\A_1$, $\A_2$, and $\A_3$ were 1.52, 4.59, 1.63, respectively.

\begin{figure}[!t]
	\centering
	\includegraphics[width=.9\linewidth]{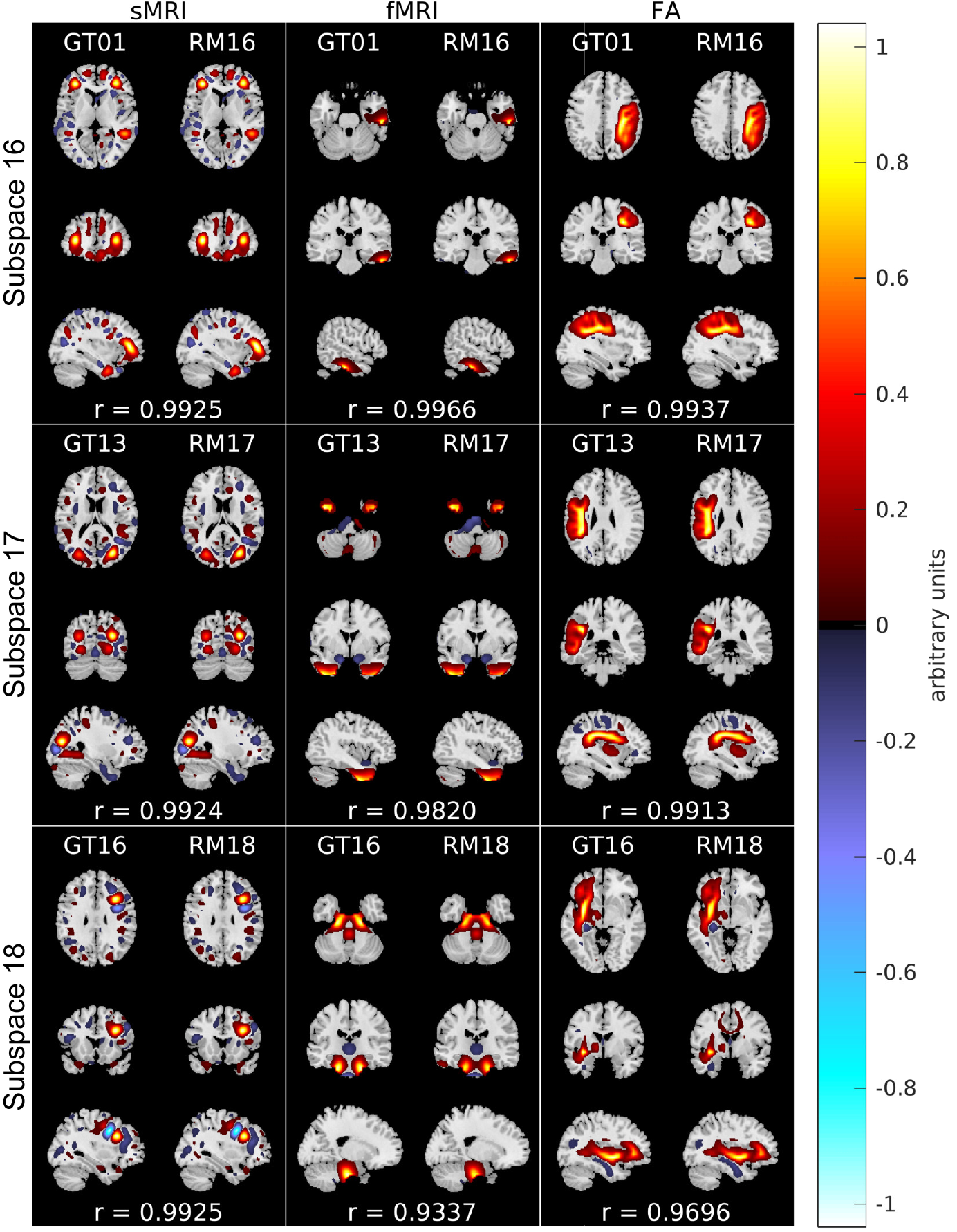}
	\caption{\textbf{Summary of multimodal IVA maps}. In each panel, ground-truth (GT) maps are presented on the left and maps estimated from MISA with \ac{PRE} (RM) on the right. Each subspace represents the multimodal set of maps (joint features) with highest, median, and minimum correlation with the GT, from top to bottom, respectively. \emph{No IVA-L comparison available since it failed to converge, likely due to the small sample size ($N=600$).}}
	\label{fig:hybMMIVAsummary}
\end{figure}
In the results, the data was reduced using \ac{PRE} and then processed with MISA to obtain independent \emph{subject expression levels}.
Per-modality correlation between ground-truth and PRE+MISA spatial maps are presented in Fig.~\ref{fig:hybMMIVAcorr}, and spatial maps (estimating $\A$ from $\hat{\W}^{-}$) in Fig.~\ref{fig:hybMMIVAsummary}. MISI $= 0.0273$.
\subsubsection{Multimodal MISA of fMRI, and EEG}\label{subsubsec:MMMISA}
We show the value of MDM models without data reduction for fusion of \ac{EEG} \ac{ERP} and \ac{fMRI} datasets with dimensionality $V_1 = \mathrm{time~points} \approx 600$, $V_2 = \mathrm{voxels} \approx 67\mathrm{k}$, respectively,
and $N = \mathrm{subjects} = 1001$.
Let $C_1 = 4$ and $C_2 = 6$ sources in the ERP and fMRI datasets, respectively, organized into $K=4$ subspaces ($y_{mi}$ represents source $i$ from dataset $m$):
\begin{IEEEdescription}[\IEEEsetlabelwidth{$k=9$:}]
	\item[$k=1$:] IVA-type, sources $y_{11}$ and $y_{21}$ ($d_k=2$);
	\item[$k=2$:] MISA-type, sources $y_{12}$, $y_{22}$ and $y_{23}$ ($d_k=3$);
	\item[$k=3$:] MISA-type, sources $y_{13}$, $y_{14}$ and $y_{24}$ ($d_k=3$);
	\item[$k=4$:] ISA-type, sources $y_{25}$ and $y_{26}$ ($d_k=2$).
\end{IEEEdescription}
Utilizing real spatial maps and timecourses, $\A_1$ and $\A_2$ must be $600 \times 4$ and $67\mathrm{k} \times 6$, respectively, this time ensuring they form column-orthogonal mixings (with Gram-Schmidt).

For the simulated part of the data, we generate $4 \times 1001$ and $6 \times 1001$ matrices of subject expression levels for ERP and fMRI datasets, respectively.
A total of $K=4$ $d_k$-dimensional subspaces with $N = 600$ observations each were sampled from a multivariate Laplace distribution, using an inverse exponential autocorrelation function with maximal correlation of $0.65$ for each subspace.
Noise was absent in both datasets. The condition number was 1.00 for both $\A_1$ and $\A_2$.

\begin{figure}[!t]
	\centering
	\captionsetup[subfloat]{justification=centering}
	\subfloat{\vtop{\vskip -8pt \hbox{\includegraphics[width=.545\linewidth,trim={0 14.3cm 0 0},clip]{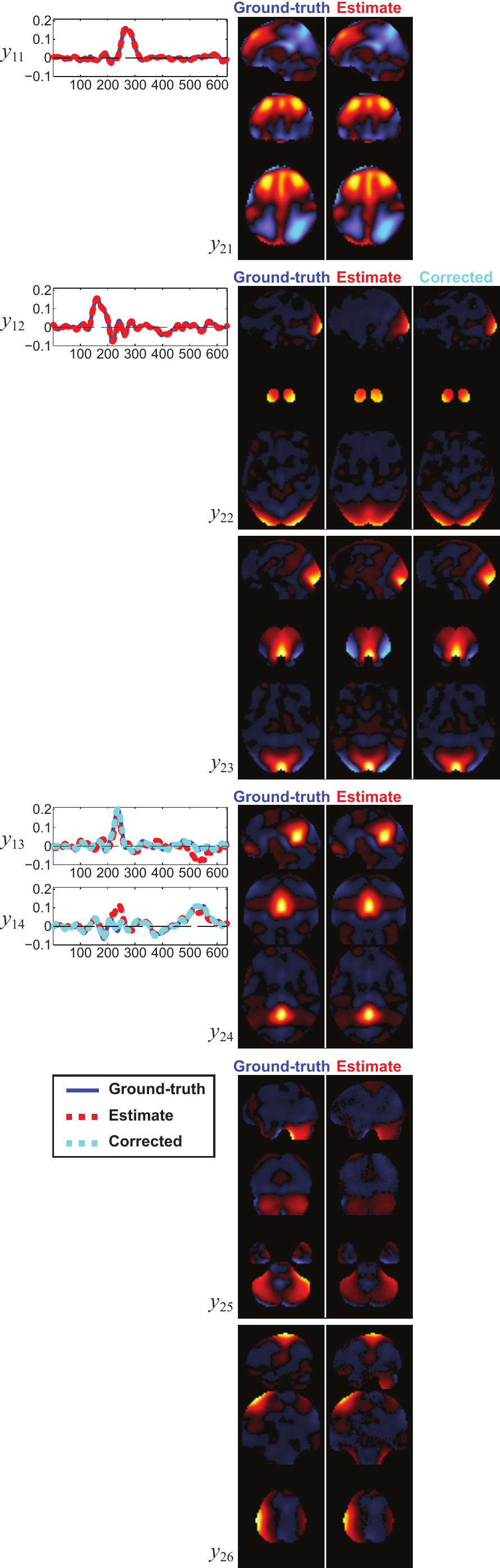}}}}
	\hfil
	\subfloat{\vtop{\vskip -8pt \hbox{\includegraphics[width=.45\linewidth,trim={0 0 45 14.25cm},clip]{MISA_MISA_EEGfMRI}}}}
	\caption{\textbf{Multimodal MISA of fMRI and ERP}. GT maps are presented on the left of each panel, MISA-GP estimates in the middle, and corrected MISA-GP estimates on the right.  GT ERPs are presented in blue, MISA-GP ERPs in dashed red, and corrected MISA-GP ERPs in dashed cyan.}
	\label{fig:hybMMMISA_maps}
\end{figure}

Fig.~\ref{fig:hybMMMISA_maps} shows the results obtained from constrained MISA-GP, i.e., with $\hat{\A} = \W^{\top}$ \ac{RE} constraint using~\eqref{eq:REWT}. %
No data reduction was performed on the data.
The spatial fMRI maps and ERP timecourses were produced by estimating $\A$ from $\hat{\W}^{\top}$.
Since subspace independence is invariant to linear transformations (arbitrary basis) within any subspace~\cite{CardosoMICA98}, the estimation yields timecourses (red) and maps (middle) that do not match the GT exactly.
In an attempt to correct for that, we performed additional within-modality ICAs on the columns of $\A_m$ corresponding to subspaces.
This effectively selected for a particular basis within each subspace (right maps and cyan timecourses).
The ability to choose a particular representation demonstrates the kinds of post-processing enabled by MDM models.
Overall, this result validates and illustrates the benefit of a constrained optimization approach.

\section{Conclusion}\label{sec:ConclusionMain}

We have presented \ac{MISA}, an approach that solves multiple \ac{BSS} problems (including \ac{ICA}, \ac{IVA}, \ac{ISA}, and more) under the same framework, with remarkable performance and improved robustness even at low \ac{SNR}.
In particular, we derived a general formulation that controls for source scales, leveraging the flexible Kotz distribution in an interior point non-linear constraint optimization, with \ac{PRE} as a general and flexible formulation for either direct subspace estimation or dimensionality reduction, in conjunction with combinatorial optimization for evasion of local minima, permitting self-correction to the closest subspace structures supported by the data (MISA-GP).
Altogether, the proposed methods permit all-order statistics linkage across multidatasets as well as features of higher complexity to be identified and fully exploited in a direct, principled, and synergistic way, even at sample sizes as low as $N=600$.
Flexible approaches like MISA are key to meet the growing complexity of multidataset tasks.
These complexities are incorporated in the hybrid dataset standards we open source here, built from relevant results published in the brain imaging \ac{BSS} literature.
Generalizations building on this work could be easily developed exploring other divergence families.
Future work will focus on compiling real multimodal datasets to validate MISA's ability to capture reliable modes of shared and unique variability across and within modalities.

\ifCLASSOPTIONcaptionsoff
  \newpage
\fi

\bibliographystyle{IEEEtran}
\bibliography{IEEEabrv,rogers_refs}

 \newcommand{\noop}[1]{}
\begin{thebibliography}{10}
\providecommand{\url}[1]{#1}
\csname url@samestyle\endcsname
\providecommand{\newblock}{\relax}
\providecommand{\bibinfo}[2]{#2}
\providecommand{\BIBentrySTDinterwordspacing}{\spaceskip=0pt\relax}
\providecommand{\BIBentryALTinterwordstretchfactor}{4}
\providecommand{\BIBentryALTinterwordspacing}{\spaceskip=\fontdimen2\font plus
\BIBentryALTinterwordstretchfactor\fontdimen3\font minus
  \fontdimen4\font\relax}
\providecommand{\BIBforeignlanguage}[2]{{%
\expandafter\ifx\csname l@#1\endcsname\relax
\typeout{** WARNING: IEEEtran.bst: No hyphenation pattern has been}%
\typeout{** loaded for the language `#1'. Using the pattern for}%
\typeout{** the default language instead.}%
\else
\language=\csname l@#1\endcsname
\fi
#2}}
\providecommand{\BIBdecl}{\relax}
\BIBdecl

\bibitem{Silva2016_UnifyBSSReview}
R.~F. Silva, S.~M. Plis, J.~Sui, M.~S. Pattichis, T.~Adal{\i}, and V.~D.
  Calhoun, ``Blind source separation for unimodal and multimodal brain
  networks: A unifying framework for subspace modeling,'' \emph{IEEE J Sel
  Topics Signal Process}, vol.~10, no.~7, pp. 1134--1149, 2016.

\bibitem{ComonJutten2010Handbook}
P.~Comon and C.~Jutten, \emph{Handbook of Blind Source Separation},
  1st~ed.\hskip 1em plus 0.5em minus 0.4em\relax Oxford, UK: Academic Press,
  2010.

\bibitem{YiL2016_ChemoMethodsReview}
L.~Yi, N.~Dong, Y.~Yun, B.~Deng, D.~Ren, S.~Liu, and Y.~Liang, ``Chemometric
  methods in data processing of mass spectrometry-based metabolomics: A
  review,'' \emph{Anal. Chim Acta}, vol. 914, pp. 17--34, 2016.

\bibitem{SaitoS2015_BSSAJDSpeech}
S.~{Saito}, K.~{Oishi}, and T.~{Furukawa}, ``{Convolutive Blind Source
  Separation Using an Iterative Least-Squares Algorithm for Non-Orthogonal
  Approximate Joint Diagonalization},'' \emph{IEEE/ACM Trans Audio Speech Lang
  Process}, vol.~23, no.~12, pp. 2434--2448, 2015.

\bibitem{NielsenA2002_MCCA_MultispectralTemporal}
A.~A. {Nielsen}, ``{Multiset canonical correlations analysis and multispectral,
  truly multitemporal remote sensing data},'' \emph{IEEE Trans Image Process},
  vol.~11, no.~3, pp. 293--305, 2002.

\bibitem{AmmanouilR2014_BSSSparseHyperspectral}
R.~{Ammanouil}, A.~{Ferrari}, C.~{Richard}, and D.~{Mary}, ``{Blind and Fully
  Constrained Unmixing of Hyperspectral Images},'' \emph{IEEE Trans Image
  Process}, vol.~23, no.~12, pp. 5510--5518, 2014.

\bibitem{Calhoun2009ReviewICAJoint}
V.~D. Calhoun, J.~Liu, and T.~Adal{\i}, ``A review of group {ICA} for {fMRI}
  data and {ICA} for joint inference of imaging, genetic, and {ERP} data,''
  \emph{NeuroImage}, vol.~45, no. 1, Supplement 1, pp. S163--S172, 2009.

\bibitem{2016CalhounVD_FusionReview}
V.~D. Calhoun and J.~Sui, ``Multimodal fusion of brain imaging data: A key to
  finding the missing link(s) in complex mental illness,'' \emph{Biol
  Psychiatry Cogn Neurosci Neuroimag}, vol.~1, no.~3, pp. 230--244, 2016.

\bibitem{BhingeS_IVA_MultiCameraObject}
S.~{Bhinge}, Y.~{Levin-Schwartz}, and T.~{Adali}, ``Data-driven fusion of
  multi-camera video sequences: Application to abandoned object detection,'' in
  \emph{{Proc IEEE ICASSP 2017}}, 2017, pp. 1697--1701.

\bibitem{NicolaouM2014_DynProbCCA_VideoAnnotation}
M.~A. {Nicolaou}, V.~{Pavlovic}, and M.~{Pantic}, ``{Dyn. Probabilistic CCA
  Analysis of Affective Behavior and Fusion of Conts. Annotations},''
  \emph{IEEE Trans Pattern Anal Mach Intell}, vol.~36, no.~7, pp. 1299--1311,
  2014.

\bibitem{Lahat2015_MultimodalDataFusionReview}
D.~Lahat, T.~Adal{\i}, and C.~Jutten, ``Multimodal data fusion: An overview of
  methods, challenges, and prospects,'' \emph{Proc IEEE}, vol. 103, no.~9, pp.
  1449--1477, 2015.

\bibitem{MillerKL2016_Multimodal_UKBiobank5K}
K.~L. Miller, F.~Alfaro-Almagro, N.~K. Bangerter, D.~L. Thomas, E.~Yacoub,
  J.~Xu, A.~J. Bartsch, S.~Jbabdi, S.~N. Sotiropoulos, J.~L.~R. Andersson,
  L.~Griffanti, G.~Douaud, T.~W. Okell, P.~Weale, I.~Dragonu, S.~Garratt,
  S.~Hudson, R.~Collins, M.~Jenkinson, P.~M. Matthews, and S.~M. Smith,
  ``Multimodal population brain imaging in the {UK Biobank} prospective
  epidemiological study,'' \emph{Nat Neurosci}, vol.~19, no.~11, pp.
  1523--1536, 2016.

\bibitem{Calhoun2012ReviewNeurodicoveryICA}
V.~D. Calhoun and T.~Adal{\i}, ``Multisubject independent component analysis of
  {fMRI}: A decade of intrinsic networks, default mode, and neurodiagn.
  discovery,'' \emph{IEEE Rev Biomed Eng}, vol.~5, pp. 60--73, 2012.

\bibitem{SeghouaneA2017_SequentialDicLearn}
A.~{Seghouane} and A.~{Iqbal}, ``{Sequential Dictionary Learning From
  Correlated Data: Application to fMRI Data Analysis},'' \emph{IEEE Trans Image
  Process}, vol.~26, no.~6, pp. 3002--3015, 2017.

\bibitem{MohammadiNAR2017_ssCCAfusion}
A.~R. Mohammadi-Nejad, G.~A. Hossein-Zadeh, and H.~Soltanian-Zadeh,
  ``Structured and sparse canonical correlation analysis as a brain-wide
  multi-modal data fusion approach,'' \emph{IEEE Trans Med Imaging}, vol.~36,
  no.~7, pp. 1438--1448, 2017.

\bibitem{LeeIVAfMRI_NIMG08}
J.-H. Lee, T.-W. Lee, F.~Jolesz, and S.-S. Yoo, ``Independent vector analysis
  ({IVA}): Multivariate approach for {fMRI} group study,'' \emph{NeuroImage},
  vol.~40, no.~1, pp. 86--109, 2008.

\bibitem{BhingeS_ContrainIVA}
S.~{Bhinge}, R.~{Mowakeaa}, V.~D. {Calhoun}, and T.~{Adalı}, ``{Extraction of
  time-varying spatio-temporal networks using parameter-tuned constrained
  IVA},'' \emph{IEEE Trans Med Imaging}, 2019.

\bibitem{PakravanM2018_JIMDM}
M.~{Pakravan} and M.~B. {Shamsollahi}, ``{Extraction and Automatic Grouping of
  Joint and Individual Sources in Multi-Subject fMRI Data Using Higher Order
  Cumulants},'' \emph{IEEE J Biomed Health Inform}, 2018.

\bibitem{YuM2018_HarmonizationMultisiteFMRI}
M.~Yu, K.~A. Linn, P.~A. Cook, M.~L. Phillips, M.~McInnis, M.~Fava, M.~H.
  Trivedi, M.~M. Weissman, R.~T. Shinohara, and Y.~I. Sheline, ``{Statistical
  harmonization corrects site effects in functional connectivity measurements
  from multi-site fMRI data},'' \emph{Hum Brain Mapp}, vol.~39, no.~11, pp.
  4213--4227, 2018.

\bibitem{MirzaalianH2016_HarmonizationMultisiteDiffusion}
H.~Mirzaalian, L.~Ning, P.~Savadjiev, O.~Pasternak, S.~Bouix, O.~Michailovich,
  G.~Grant, C.~Marx, R.~Morey, L.~Flashman, M.~George, T.~McAllister,
  N.~Andaluz, L.~Shutter, R.~Coimbra, R.~Zafonte, M.~Coleman, M.~Kubicki,
  C.~Westin, M.~Stein, M.~Shenton, and Y.~Rathi, ``{Inter-site and
  inter-scanner diffusion MRI data harmonization},'' \emph{NeuroImage}, vol.
  135, pp. 311--323, 2016.

\bibitem{AlamF2017_IoTFusion}
F.~{Alam}, R.~{Mehmood}, I.~{Katib}, N.~N. {Albogami}, and A.~{Albeshri},
  ``{Data Fusion and IoT for Smart Ubiquitous Environments: A Survey},''
  \emph{IEEE Access}, vol.~5, pp. 9533--9554, 2017.

\bibitem{ElmadanyN2018_GlobalLocalCCAFusion}
N.~E.~D. {Elmadany}, Y.~{He}, and L.~{Guan}, ``{Information Fusion for Human
  Action Recognition via Biset/Multiset Globality Locality Preserving Canonical
  Correlation Analysis},'' \emph{IEEE Trans Image Process}, vol.~27, no.~11,
  pp. 5275--5287, 2018.

\bibitem{UzairM2015_PLSRHyperspectralFace}
M.~{Uzair}, A.~{Mahmood}, and A.~{Mian}, ``{Hyperspectral Face Recognition With
  Spatiospectral Information Fusion and PLS Regression},'' \emph{IEEE Trans
  Image Process}, vol.~24, no.~3, pp. 1127--1137, 2015.

\bibitem{YaoJ2019_NMFHyperspectral}
J.~{Yao}, D.~{Meng}, Q.~{Zhao}, W.~{Cao}, and Z.~{Xu}, ``{Nonconvex-sparsity
  and Nonlocal-smoothness Based Blind Hyperspectral Unmixing},'' \emph{IEEE
  Trans Image Process}, 2019.

\bibitem{VillaA2011_ICAHyperspectralJutten}
A.~{Villa}, J.~A. {Benediktsson}, J.~{Chanussot}, and C.~{Jutten},
  ``{Hyperspectral Image Classification With Indep. Component Discriminant
  Analysis},'' \emph{IEEE Trans Geosci Remote Sens}, vol.~49, no.~12, pp.
  4865--4876, 2011.

\bibitem{XuH2018_DomainAdaptiveDictLearn}
\BIBentryALTinterwordspacing
H.~Xu, J.~Zheng, A.~Alavi, and R.~Chellappa, ``Cross-domain visual recognition
  via domain adaptive dictionary learning,'' \emph{arXiv preprint}, 2018.
  [Online]. Available: \url{http://arxiv.org/abs/1804.04687}
\BIBentrySTDinterwordspacing

\bibitem{FanJ2017_VisRecog_Groups}
J.~Fan, T.~Zhao, Z.~Kuang, Y.~Zheng, J.~Zhang, J.~Yu, and J.~Peng, ``{HD-MTL:
  Hierarchical Deep Multi-Task Learning Large-Scale Visual Recog.}'' \emph{IEEE
  Trans Image Process}, vol.~26, no.~4, pp. 1923--1938, 2017.

\bibitem{LuH_EmbarassDomainAdapt_ClassMean}
H.~Lu, C.~Shen, Z.~Cao, Y.~Xiao, and A.~van~den Hengel, ``An embarrassingly
  simple approach to visual domain adaptation,'' \emph{IEEE Trans Image
  Process}, vol.~27, no.~7, pp. 3403--3417, 2018.

\bibitem{LongM_JointMatchDomainAdapt}
M.~Long, J.~Wang, G.~Ding, J.~Sun, and P.~S. Yu, ``Transfer joint matching for
  unsupervised domain adaptation,'' in \emph{Proc {IEEE CVPR} 2014}, 2014, pp.
  1410--1417.

\bibitem{PatelVM_DomainAdaptOverview}
V.~M. Patel, R.~Gopalan, R.~Li, and R.~Chellappa, ``Visual domain adaptation: A
  survey of recent advances,'' \emph{IEEE Signal Process Mag}, vol.~32, no.~3,
  pp. 53--69, 2015.

\bibitem{CaiZ2014_CCA_Multi-viewVideo}
Z.~{Cai}, L.~{Wang}, X.~{Peng}, and Y.~{Qiao}, ``Multi-view super vector for
  action recognition,'' in \emph{{Proc IEEE CVPR 2014}}, 2014, pp. 596--603.

\bibitem{TangL2018_CCAMultiViewObjRecog}
L.~{Tang}, Z.~{Yang}, and K.~{Jia}, ``{Canonical Correlation Analysis
  Regularization: An Effective Deep Multi-View Learning Baseline for RGB-D
  Object Recognition},'' \emph{IEEE Trans Cogn Devel Syst}, 2018.

\bibitem{SarginM2007_CCA_AudioVisualFusion}
M.~E. {Sargin}, Y.~{Yemez}, E.~{Erzin}, and A.~M. {Tekalp}, ``{Audiovisual
  Synchronization and Fusion Using Canonical Correlation Analysis},''
  \emph{IEEE Trans Multimedia}, vol.~9, no.~7, pp. 1396--1403, 2007.

\bibitem{GaoL2019_LabeledMCCA_AudioVisual}
L.~{Gao}, R.~{Zhang}, L.~{Qi}, E.~{Chen}, and L.~{Guan}, ``{The Labeled
  Multiple Canonical Correlation Analysis for Information Fusion},'' \emph{IEEE
  Trans Multimedia}, vol.~21, no.~2, pp. 375--387, 2019.

\bibitem{PierreN_IVAAudioVisual}
P.~Narvor, B.~Rivet, and C.~Jutten, ``Audiovisual speech separation based on
  independent vector analysis using a visual voice activity detector,'' in
  \emph{Proc LVA/ICA 2017}, Grenoble, France, 2017, pp. 247--257.

\bibitem{NestaF2017_IVAAudioVisual}
F.~{Nesta}, S.~{Mosayyebpour}, Z.~{Koldovský}, and K.~{Paleček},
  ``{Audio/video supervised independent vector analysis through multimodal
  pilot dependent components},'' in \emph{{Proc EUSIPCO 2017}}, 2017, pp.
  1150--1164.

\bibitem{Silva2014MISA_OHBM}
R.~F. Silva, S.~M. Plis, T.~Adal{\i}, and V.~D. Calhoun, ``Multidataset
  independent subspace analysis,'' in \emph{Proc {OHBM} 2014}, Hamburg,
  Germany, 2014, {Poster 3506}.

\bibitem{SilvaRF_MISA_ICIP}
------, ``Multidataset independent subspace analysis extends independent vector
  analysis,'' in \emph{Proc {IEEE ICIP} 2014}, France, 2014, pp. 2864--2868.

\bibitem{Silva2015MISASOS_OHBM}
R.~F. Silva, S.~M. Plis, M.~S. Pattichis, T.~Adal{\i}, and V.~D. Calhoun,
  ``Incorporating second-order statistics in multidataset independent subspace
  analysis,'' in \emph{Proc {OHBM} 2015}, Honolulu, HI, 2015, {Poster 3743}.

\bibitem{Comon1994ICANewConcept}
P.~Comon, ``Independent component analysis, a new concept?'' \emph{Signal
  Process}, vol.~36, no.~3, pp. 287--314, 1994.

\bibitem{CardosoMICA98}
J.-F. Cardoso, ``Multidimensional independent component analysis,'' in
  \emph{Proc {IEEE ICASSP} 1998}, vol.~4, Seattle, WA, 1998, pp. 1941--1944.

\bibitem{KimT_IVA2006}
T.~Kim, T.~Eltoft, and T.-W. Lee, ``Independent vector analysis: An extension
  of {ICA} to multivariate components,'' in \emph{Proc {ICA} 2006}, Charleston,
  SC, 2006, vol. 3889, pp. 165--172.

\bibitem{SilvaRF_MMSimNIMG2014}
R.~F. Silva, S.~M. Plis, T.~Adal{\i}, and V.~D. Calhoun, ``A statistically
  motivated framework for simulation of stochastic data fusion models applied
  to multimodal neuroimaging,'' \emph{NeuroImage}, vol. 102, Part 1, pp.
  92--117, 2014.

\bibitem{BellSejnowski1995Infomax}
A.~Bell and T.~Sejnowski, ``An information-maximization approach to blind
  separation and blind deconvolution.'' \emph{Neural Comput}, vol.~7, no.~6,
  pp. 1129--1159, 1995.

\bibitem{Amari1998_NaturalGrad}
S.-I. Amari, ``Natural gradient works efficiently in learning,'' \emph{Neural
  Comput}, vol.~10, no.~2, pp. 251--276, 1998.

\bibitem{Anderson2012_JBSSGaussianAlgPerf}
M.~Anderson, T.~Adal{\i}, and X.~L. Li, ``Joint blind source separation with
  multivariate gaussian model: Algorithms and performance analysis,''
  \emph{IEEE Trans Signal Process}, vol.~60, no.~4, pp. 1672--1683, 2012.

\bibitem{LahatD2012_MICA_SOS}
D.~Lahat, J.~Cardoso, and H.~Messer, ``Second-order multidimensional {ICA}:
  Performance analysis,'' \emph{IEEE Trans Signal Process}, vol.~60, no.~9, pp.
  4598--4610, 2012.

\bibitem{LeQ2011_ISA}
Q.~V. {Le}, W.~Y. {Zou}, S.~Y. {Yeung}, and A.~Y. {Ng}, ``Learning hierarchical
  invariant spatio-temporal features for action recognition with independent
  subspace analysis,'' in \emph{{Proc CVPR 2011}}, 2011, pp. 3361--3368.

\bibitem{SilvaRF2019_BookPPP_Ch8}
R.~F. Silva and S.~M. Plis, \emph{How to Integrate Data from Multiple
  Biological Layers in Mental Health?}\hskip 1em plus 0.5em minus 0.4em\relax
  Springer, 2019, pp. 135--159.

\bibitem{HyvarinenNIS09}
A.~Hyv{\"a}rinen, J.~Hurri, and P.~Hoyer, \emph{Natural Image Statistics: A
  Probabilistic Approach to Early Computational Vision}, 1st~ed., ser.
  Computational Imaging and Vision.\hskip 1em plus 0.5em minus 0.4em\relax
  Springer, 2009, vol.~39.

\bibitem{BouwmansT2018_RPCA_OtazoR}
T.~{Bouwmans}, S.~{Javed}, H.~{Zhang}, Z.~{Lin}, and R.~{Otazo}, ``{On the
  Applications of Robust PCA in Image and Video Processing},'' \emph{Proc
  IEEE}, vol. 106, no.~8, pp. 1427--1457, 2018.

\bibitem{PontTusetJ2017_CombolGroupingSegment}
J.~{Pont-Tuset}, P.~{Arbeláez}, J.~{T. Barron}, F.~{Marques}, and J.~{Malik},
  ``Multiscale combinatorial grouping for image segmentation and object
  proposal generation,'' \emph{IEEE Trans Pattern Anal Mach Intell}, vol.~39,
  no.~1, pp. 128--140, 2017.

\bibitem{ElZehiryN2016_ComboElasticaSegmentation}
N.~Y. {El-Zehiry} and L.~{Grady}, ``Contrast driven elastica image segmnt.''
  \emph{IEEE Trans Image Process}, vol.~25, no.~6, pp. 2508--2518, 2016.

\bibitem{Szabo2012ISAbyICAPerm}
Z.~Szab\'{o}, B.~P\'{o}czos, and A.~L\H{o}rincz, ``Separation theorem for
  independent subspace analysis and its consequences,'' \emph{Pattern
  Recognit}, vol.~45, no.~4, pp. 1782--1791, 2012.

\bibitem{LeeTW_ExtendedICA}
T.-W. Lee and T.~J. Sejnowski, ``Independent component analysis for mixed
  sub-gaussian and super-gaussian sources,'' in \emph{Proc 4th {JSNC}}.\hskip
  1em plus 0.5em minus 0.4em\relax INC, 1997, pp. 132--139.

\bibitem{HyvarinenESANN06}
A.~Hyv{\"a}rinen and U.~K{\"o}ster, ``{FastISA}: A fast fixed-point algorithm
  for independent subspace analysis,'' in \emph{Proc {ESANN}}, 2006, pp.
  371--376.

\bibitem{AndersonM_2012IVAKOTZ}
M.~Anderson, G.-S. Fu, R.~Phlypo, and T.~Adal{\i}, ``{Independent vector
  analysis, the Kotz distribution, and performance bounds},'' in \emph{Proc
  {IEEE ICASSP} 2013}, Vancouver, Canada, 2013, pp. 3243--3247.

\bibitem{Kotz1975_MVDistCrossRoad}
S.~Kotz, ``Multivariate distributions at a cross road,'' in \emph{Proc {NATO
  Advanced Study Institute, Statistical Distributions in Scientific
  Work}}.\hskip 1em plus 0.5em minus 0.4em\relax Calgary, Canada: Springer,
  1974, pp. 247--270.

\bibitem{EltoftT2006_MultivariateK}
T.~Eltoft, T.~Kim, and T.-W. Lee, ``Multivariate scale mixture of gaussians
  modeling,'' in \emph{{Proc ICA 2006}}, Charleston, USA, 2006, pp. 799--806.

\bibitem{AndersonM2013_IVAthesis}
M.~Anderson, ``{Independent vector analysis: Theory, algorithms, and
  applications},'' Doctoral Dissertation, University of Maryland, Baltimore
  County, Baltimore, NM, USA, 2013.

\bibitem{LiX_ICAEBM}
X.~{Li} and T.~{Adali}, ``Independent component analysis by entropy bound
  minimization,'' \emph{IEEE Transactions on Signal Processing}, vol.~58,
  no.~10, pp. 5151--5164, 2010.

\bibitem{Anderson2010IVAG}
M.~Anderson, X.-L. Li, and T.~Adal{\i}, ``Nonorthogonal independent vector
  analysis using multivariate gaussian model,'' in \emph{Proc {LVA/ICA} 2010},
  ser. Lecture Notes in Computer Science.\hskip 1em plus 0.5em minus
  0.4em\relax France: Springer, 2010, vol. 6365, pp. 354--361.

\bibitem{Lahat2016_JISASOS_IEEETSP}
D.~Lahat and C.~Jutten, ``Joint independent subspace analysis using
  second-order statistics,'' \emph{IEEE Trans Signal Process}, vol.~64, no.~18,
  pp. 4891--4904, 2016.

\bibitem{NadarajahS2003_KotzDistApps}
S.~Nadarajah, ``The {K}otz-type distribution with applications,''
  \emph{Statistics}, vol.~37, no.~4, pp. 341--358, 2003.

\bibitem{Cardoso1996_RelativeGrad}
J.~F. Cardoso and B.~H. Laheld, ``Equivariant adaptive source separation,''
  \emph{IEEE Trans Signal Process}, vol.~44, no.~12, pp. 3017--3030, 1996.

\bibitem{ByrdRH1995_LBFGSB}
R.~H. Byrd, P.~Lu, J.~Nocedal, and C.~Zhu, ``A limited memory algorithm for
  bound constrained optimization,'' \emph{SIAM J Sci Comput}, vol.~16, no.~5,
  pp. 1190--1208, 1995.

\bibitem{ZhuC1997_LBFGSB_Alg}
C.~Zhu, R.~H. Byrd, P.~Lu, and J.~Nocedal, ``Algorithm 778: {L-BFGS-B}: Fortran
  subroutines for large-scale bound-constrained optimization,'' \emph{ACM Trans
  Math Softw}, vol.~23, no.~4, pp. 550--560, 1997.

\bibitem{Nocedal2006_NumOpt}
J.~Nocedal and S.~Wright, \emph{Numerical Optimization}, 2nd~ed.\hskip 1em plus
  0.5em minus 0.4em\relax New York, NY: Springer, 2006.

\bibitem{Waltz2006_NLOpt_LS_TR_KNITRO}
R.~Waltz, J.~Morales, J.~Nocedal, and D.~Orban, ``An interior algorithm for
  nonlinear optimization that combines line search and trust region steps,''
  \emph{Math Prog}, vol. 107, no.~3, pp. 391--408, 2006.

\bibitem{Haufe2014_BackToFWmodels_PLSreconstruction}
S.~Haufe, F.~Meinecke, K.~G{\"o}rgen, S.~D{\"a}hne, J.-D. Haynes, B.~Blankertz,
  and F.~Bie{\ss}mann, ``On the interpretation of weight vectors of linear
  models in multivar. neuroimag.'' \emph{NeuroImage}, vol.~87, pp. 96--110,
  2014.

\bibitem{LeQ_ReconstructErrorENIPS2011}
Q.~Le, A.~Karpenko, J.~Ngiam, and A.~Ng, ``{ICA} with reconstruction cost for
  efficient overcomplete feature learning,'' in \emph{Proc {NIPS} 2011},
  Granada, Spain, 2011, pp. 1017--1025.

\bibitem{GIFT2015_v4a}
\BIBentryALTinterwordspacing
MIALAB, ``{Group ICA of fMRI Toolbox (GIFT)},'' 2015. [Online]. Available:
  \url{http://trendscenter.org/trends/software/gift/index.html}
\BIBentrySTDinterwordspacing

\bibitem{2016Rachakonda_MemEffGroupPCA}
S.~Rachakonda, R.~F. Silva, J.~Liu, and V.~D. Calhoun, ``Memory efficient {PCA}
  methods for large group {ICA},'' \emph{Front Neurosci}, vol.~10, p.~17, 2016.

\bibitem{RN44}
S.-I. Amari, A.~Cichocki, and H.~H. Yang, ``A new learning algorithm for blind
  signal separation,'' \emph{Proc NIPS 1996}, vol.~8, pp. 757--763, 1996.

\bibitem{Macchi1995}
O.~Macchi and E.~Moreau, ``Self-adaptive source separation by direct or
  recursive networks,'' in \emph{Proc {ICDSP} 1995}, Cyprus, 1995, pp.
  122--129.

\bibitem{Allen2011baseline}
E.~A. Allen, E.~B. Erhardt, E.~Damaraju, W.~Gruner, J.~M. Segall, R.~Silva,
  M.~Havlicek, S.~Rachakonda, J.~Fries, R.~Kalyanam, and {et al.}, ``A baseline
  for the multivariate comparison of resting-state networks,'' \emph{Front Syst
  Neurosci}, vol.~5, 2011.

\bibitem{RN2609}
R.~Nelsen, \emph{An Introduction to Copulas}, 2nd~ed., ser. Springer Series in
  Statistics.\hskip 1em plus 0.5em minus 0.4em\relax New York, NY: Springer New
  York, 2006, vol.~1.

\bibitem{judith2012}
J.~Segall, E.~Allen, R.~Jung, E.~Erhardt, S.~Arja, K.~Kiehl, and V.~Calhoun,
  ``Correspondence between structure and function in the human brain at rest,''
  \emph{Front Neuroinform}, vol.~6, p.~10, 2012.

\bibitem{HBM:HBM22945}
L.~Wu, V.~D. Calhoun, R.~E. Jung, and A.~Caprihan, ``Connectivity-based whole
  brain dual parcellation by group {ICA} reveals tract structures and decreased
  connectivity in schizophrenia,'' \emph{Hum Brain Mapp}, vol.~36, no.~11, pp.
  4681--4701, 2015.

\end{thebibliography}


 \newcommand{\noop}[1]{}
\begin{thebibliography}{10}
\providecommand{\url}[1]{#1}
\csname url@samestyle\endcsname
\providecommand{\newblock}{\relax}
\providecommand{\bibinfo}[2]{#2}
\providecommand{\BIBentrySTDinterwordspacing}{\spaceskip=0pt\relax}
\providecommand{\BIBentryALTinterwordstretchfactor}{4}
\providecommand{\BIBentryALTinterwordspacing}{\spaceskip=\fontdimen2\font plus
\BIBentryALTinterwordstretchfactor\fontdimen3\font minus
  \fontdimen4\font\relax}
\providecommand{\BIBforeignlanguage}[2]{{%
\expandafter\ifx\csname l@#1\endcsname\relax
\typeout{** WARNING: IEEEtran.bst: No hyphenation pattern has been}%
\typeout{** loaded for the language `#1'. Using the pattern for}%
\typeout{** the default language instead.}%
\else
\language=\csname l@#1\endcsname
\fi
#2}}
\providecommand{\BIBdecl}{\relax}
\BIBdecl

\bibitem{Cardoso1996_RelativeGrad}
J.~F. Cardoso and B.~H. Laheld, ``Equivariant adaptive source separation,''
  \emph{IEEE Trans Signal Process}, vol.~44, no.~12, pp. 3017--3030, 1996.

\bibitem{Amari1998_NaturalGrad}
S.-I. Amari, ``Natural gradient works efficiently in learning,'' \emph{Neural
  Comput}, vol.~10, no.~2, pp. 251--276, 1998.

\bibitem{ComonJutten2010Handbook}
P.~Comon and C.~Jutten, \emph{Handbook of Blind Source Separation},
  1st~ed.\hskip 1em plus 0.5em minus 0.4em\relax Oxford, UK: Academic Press,
  2010.

\bibitem{Nocedal2006_NumOpt}
J.~Nocedal and S.~Wright, \emph{Numerical Optimization}, 2nd~ed.\hskip 1em plus
  0.5em minus 0.4em\relax New York, NY: Springer, 2006.

\bibitem{ByrdRH1995_LBFGSB}
R.~H. Byrd, P.~Lu, J.~Nocedal, and C.~Zhu, ``A limited memory algorithm for
  bound constrained optimization,'' \emph{SIAM J Sci Comput}, vol.~16, no.~5,
  pp. 1190--1208, 1995.

\bibitem{ZhuC1997_LBFGSB_Alg}
C.~Zhu, R.~H. Byrd, P.~Lu, and J.~Nocedal, ``Algorithm 778: {L-BFGS-B}: Fortran
  subroutines for large-scale bound-constrained optimization,'' \emph{ACM Trans
  Math Softw}, vol.~23, no.~4, pp. 550--560, 1997.

\bibitem{Waltz2006_NLOpt_LS_TR_KNITRO}
R.~Waltz, J.~Morales, J.~Nocedal, and D.~Orban, ``An interior algorithm for
  nonlinear optimization that combines line search and trust region steps,''
  \emph{Math Prog}, vol. 107, no.~3, pp. 391--408, 2006.

\bibitem{SilvaRF_MMSimNIMG2014}
R.~F. Silva, S.~M. Plis, T.~Adal{\i}, and V.~D. Calhoun, ``A statistically
  motivated framework for simulation of stochastic data fusion models applied
  to multimodal neuroimaging,'' \emph{NeuroImage}, vol. 102, Part 1, pp.
  92--117, 2014.

\bibitem{RN44}
S.-I. Amari, A.~Cichocki, and H.~H. Yang, ``A new learning algorithm for blind
  signal separation,'' \emph{Proc NIPS 1996}, vol.~8, pp. 757--763, 1996.

\bibitem{Macchi1995}
O.~Macchi and E.~Moreau, ``Self-adaptive source separation by direct or
  recursive networks,'' in \emph{Proc {ICDSP} 1995}, Cyprus, 1995, pp.
  122--129.

\bibitem{Nordhausen2011_PerfBSS_ISI_MD_MSE}
K.~Nordhausen, E.~Ollila, and H.~Oja, ``On the performance indices of {ICA} and
  blind source separation,'' in \emph{Proc {IEEE SPAWC} 2011}, San Francisco,
  CA, 2011, pp. 486--490.

\bibitem{AssignmentProblems_Book}
R.~Burkard, M.~Dell'Amico, and S.~Martello, \emph{Assignment Problems},
  revised~ed.\hskip 1em plus 0.5em minus 0.4em\relax Philadelphia, PA: SIAM,
  2012.

\bibitem{BellSejnowski1995Infomax}
A.~Bell and T.~Sejnowski, ``An information-maximization approach to blind
  separation and blind deconvolution.'' \emph{Neural Comput}, vol.~7, no.~6,
  pp. 1129--1159, 1995.

\bibitem{GIFT2015_v4a}
\BIBentryALTinterwordspacing
MIALAB, ``{Group ICA of fMRI Toolbox (GIFT)},'' 2015. [Online]. Available:
  \url{http://trendscenter.org/trends/software/gift/index.html}
\BIBentrySTDinterwordspacing

\bibitem{KimT_IVA2006}
T.~Kim, T.~Eltoft, and T.-W. Lee, ``Independent vector analysis: An extension
  of {ICA} to multivariate components,'' in \emph{Proc {ICA} 2006}, Charleston,
  SC, 2006, vol. 3889, pp. 165--172.

\bibitem{Anderson2012_JBSSGaussianAlgPerf}
M.~Anderson, T.~Adal{\i}, and X.~L. Li, ``Joint blind source separation with
  multivariate gaussian model: Algorithms and performance analysis,''
  \emph{IEEE Trans Signal Process}, vol.~60, no.~4, pp. 1672--1683, 2012.

\bibitem{Anderson2010IVAG}
M.~Anderson, X.-L. Li, and T.~Adal{\i}, ``Nonorthogonal independent vector
  analysis using multivariate gaussian model,'' in \emph{Proc {LVA/ICA} 2010},
  ser. Lecture Notes in Computer Science.\hskip 1em plus 0.5em minus
  0.4em\relax France: Springer, 2010, vol. 6365, pp. 354--361.

\bibitem{LahatD2012_MICA_SOS}
D.~Lahat, J.~Cardoso, and H.~Messer, ``Second-order multidimensional {ICA}:
  Performance analysis,'' \emph{IEEE Trans Signal Process}, vol.~60, no.~9, pp.
  4598--4610, 2012.

\bibitem{LeQ2011_ISA}
Q.~V. {Le}, W.~Y. {Zou}, S.~Y. {Yeung}, and A.~Y. {Ng}, ``Learning hierarchical
  invariant spatio-temporal features for action recognition with independent
  subspace analysis,'' in \emph{{Proc CVPR 2011}}, 2011, pp. 3361--3368.

\bibitem{2016Rachakonda_MemEffGroupPCA}
S.~Rachakonda, R.~F. Silva, J.~Liu, and V.~D. Calhoun, ``Memory efficient {PCA}
  methods for large group {ICA},'' \emph{Front Neurosci}, vol.~10, p.~17, 2016.

\bibitem{HyvarinenESANN06}
A.~Hyv{\"a}rinen and U.~K{\"o}ster, ``{FastISA}: A fast fixed-point algorithm
  for independent subspace analysis,'' in \emph{Proc {ESANN}}, 2006, pp.
  371--376.

\bibitem{CardosoMICA98}
J.-F. Cardoso, ``Multidimensional independent component analysis,'' in
  \emph{Proc {IEEE ICASSP} 1998}, vol.~4, Seattle, WA, 1998, pp. 1941--1944.

\end{thebibliography}

\end{document}


\title{Supplemental Material for \\Multidataset Independent Subspace Analysis \\with Application to Multimodal Fusion}

\author{Rogers~F.~Silva$^\star$,~\IEEEmembership{Member,~IEEE,}
        Sergey~M.~Plis,
        T\"{u}lay~Adal{\i},~\IEEEmembership{Fellow,~IEEE,}
        Marios~S.~Pattichis,~\IEEEmembership{Senior~Member,~IEEE,}
        and~Vince~D.~Calhoun,~\IEEEmembership{Fellow,~IEEE}%
\thanks{$^\star$Corresponding author.}%
}

\markboth{Under Review}%
{Rogers F. Silva \MakeLowercase{\textit{et al.}}: {Supplemental Material for} Multidataset Independent Subspace Analysis with Application to Multimodal Fusion}

\maketitle

\begin{abstract}

Here we present a series of supplemental material in support of our paper titled Multidataset Independent Subspace Analysis with Application to Multimodal Fusion.
The first part presents the results and complete details about our experiments on synthetic data simulations. It also includes details about the optimization parameters used for each algorithm we tested. The second part includes a guide to the derivation of the gradient for equation (9) (derivation for equations (7) and (18) are left as an exercise for the reader and can be obtained using the material presented here).

\end{abstract}

\IEEEpeerreviewmaketitle

\begin{acronym}[L-BFGS-B]
	\setlength{\parskip}{-1ex}
	\input{myacronyms.acr}
\end{acronym}

\section{Optimization Considerations}\label{sec:meth}
In order to identify independent subspaces, we pursue the minimization (with respect to $\W$) of an information functional reflective of the shared information amongst subspaces, utilizing numerical optimization via either regular, relative, or natural gradient descent approaches.
The regular gradient $\nabla I(\W) = \vecti{(\nabla I(\w))}$
is simply the matrix of partial derivatives of the objective function $I(\W)$ with respect to each element of $\W$.
It is an essential part of the typical textbook definition of the \emph{directional derivative} in an arbitrary direction $\U$, a key concept at the core of multivariable numerical optimization.
The first order approximation of a continuously differentiable multivariable function $f(\w)$ associated with an update $\w + \alpha\mathbf{u}$ is $f(\w + \alpha\mathbf{u}) \approx f(\w) + D_{\mathbf{u}}f(\w)$, where the directional derivative is
\begin{IEEEeqnarray}{CCC}
	D_{\mathbf{u}}f(\w) =%
	\lim_{h \rightarrow 0}{\frac{f(\w + h\mathbf{u}) - f(\w)}{h}} =%
	\nabla f(\w)^{\top} \mathbf{u}.
\end{IEEEeqnarray}
In minimization problems with iterative updates of the form $\W \leftarrow \W + \alpha \U$, where $\alpha$ is either the step-length or the learning rate, $\U$ should be a \emph{descent} direction, i.e., the directional derivative along $\U$ should be negative.
Thus, in steepest descent algorithms, $\U = -\nabla I(\W)$, i.e., the negative of the gradient at $\W$, while in Newton or quasi-Newton algorithms, $\U = -\vecti{\left( \HH^{-1} \nabla I(\w) \right)}$, i.e., 
the negative of the vectorized gradient ($\nabla I(\w) = \vect{(\nabla I(\W))}$) \emph{scaled by the inverse of its Hessian} $\HH = \nabla^2 I(\w)$ at $\w$ (or by an approximation of it).
The Hessian captures information about the local curvature of the objective function and tends to produce results in fewer iterations than steepest descent, at the cost of heavy memory requirements.

Based on the concept of \emph{relative variation}, defined as $\y \gets \y + \boldsymbol{\epsilon}\y$, early work in the neural networks literature led to the discovery of the relative gradient~\cite{Cardoso1996_RelativeGrad} and relative Hessian~\cite{Amari1998_NaturalGrad} for a multivariable function $f(\y)$.
Here, $\bm{\epsilon}$ is a matrix of infinitesimal change.
Noting that $f(\W\x) = I(\W)$, the relative gradient is defined as~\cite[Ch. 4]{ComonJutten2010Handbook}:
\begin{IEEEeqnarray}{CCl}
	\nabla_{rel} f(\y) &\triangleq&%
	\left.\frac%
	{\partial f(\y + \bm{\epsilon}\y)}%
	{\partial \bm{\epsilon}}%
	\right|_{\bm{\epsilon} = \mathbf{0}}%
	\mathrm{,} \label{eq:relgrad}
\end{IEEEeqnarray}
at  $\y = \W\x$. %
The first order approximation of a relative variation then is $f(\y + \bm{\epsilon}\y) \approx f(\y) + \nabla_{rel} f(\y)^{\top} \bm{\epsilon}$.
Letting $\bm{\epsilon} = - \alpha \nabla_{rel} f(\y)$, the relative variation becomes $\y + \bm{\epsilon}\y = \y - \alpha \nabla_{rel} f(\y) \y$.
From Equation~\eqref{eq:relgrad}, when $I(\W) = f(\W\x)$ then $\nabla_{rel} f(\y) = \nabla I(\W) \W^{\top}$, relating the relative gradient back to the regular gradient and giving the final form of the relative gradient descent update: $\W \gets \W - \alpha \nabla I(\W) \W^{\top} \W$.
Effectively, $\U = - \nabla I(\W) \W^{\top} \W$.

Lastly, the natural gradient~\cite{Amari1998_NaturalGrad} is simply the Newton version of the relative gradient.
It uses the relative Hessian, defined as
\begin{IEEEeqnarray}{CCl}
	\nabla_{rel}^2 f(\y) &=&%
	\left.\frac%
	{\partial^2 f(\y + \bm{\epsilon}\y)}%
	{\partial \bm{\epsilon}^2}%
	\right|_{\bm{\epsilon} = \mathbf{0}} \mathrm{.}
\end{IEEEeqnarray}
It approximates the relative Hessian with its mean value, leading to the method of Fisher scoring (the statistical variant of the Newton approach), with the \ac{FIM} being defined as
\begin{IEEEeqnarray}{CCl}
	\F &\triangleq&%
	\mathbb{E}_{\bm{\epsilon}} \left[ \nabla_{rel} f(\y) \nabla_{rel} f(\y)^{\top} \right]%
	=%
	-\mathbb{E}_{\bm{\epsilon}} \left[ \nabla_{rel}^2 f(\y) \right] \mathrm{.}%
\end{IEEEeqnarray}
Careful analysis~\cite[Ch. 4]{ComonJutten2010Handbook} shows that the Fisher matrix is sparse and decomposes into $2\times2$ blocks in the case of \ac{ICA}.
As a result, it is separable and grants an update rule of the form:
\begin{IEEEeqnarray}{CCl}
	\W \gets \W - \alpha \D(\W) \W%
	\nonumber
	\\ [2ex]%
	\begin{bmatrix}
		\D(\W)_{ij} \\
		\D(\W)_{ji}
	\end{bmatrix}%
	=%
	\F_{ij}^{-1}%
	\begin{bmatrix}
		\nabla_{rel} I(\W)_{ij} \\
		\nabla_{rel} I(\W)_{ji}
	\end{bmatrix} \mathrm{.}
\end{IEEEeqnarray}
Effectively, $\U = \F^{-1} \nabla I(\W) \W^{\top} \W$.
Note that $\F$ ($\bar{C} \times \bar{C}$) is considerably more memory efficient than $\HH$ ($\bar{V}\bar{C} \times \bar{V}\bar{C}$).
The relative gradient has been shown to produce better and more efficient results than the regular gradient.
It can also be shown that the natural gradient converges to the relative gradient when the sources are highly non-Gaussian.

An attractive alternative for $\U$ which we consider here is the \ac{L-BFGS}~\cite[Ch. 7]{Nocedal2006_NumOpt} to approximate $\HH^{-1}$.
This is a quasi-Newton method that approximates the Hessian efficiently as a low-rank matrix directly derived from the past $m$ gradient differences ($\U - \U_{\textrm{prev}}$) as well as the corresponding $\W$ differences ($\W - \W_{\textrm{prev}}$).
The low-memory aspect results from never directly evaluating the Hessian.
Instead, a series of memory efficient inner products and vector summations take place such that the result is implicitly equivalent to directly computing the matrix-vector product $\B \nabla I(\w)$, where $\B$ is the inverse Hessian-approximate.

In this work, we opt to use the relative gradient together with the  \ac{L-BFGS-B}~\cite{ByrdRH1995_LBFGSB,ZhuC1997_LBFGSB_Alg} available in the non-linear constraint optimization function $\mathsf{fmincon}$ of MATLAB's Optimization Toolbox.
Its approach for non-linear constraint optimization is very efficient and utilizes non-linear programing based on an interior-point barrier method~\cite[Ch. 19]{Nocedal2006_NumOpt}~\cite{Waltz2006_NLOpt_LS_TR_KNITRO} that combines line search and trust region steps.
The utility of non-linear constraint optimization for \ac{BSS} is discussed and illustrated in the main paper.

\section{Results}\label{ch:exp}
In this section, we present a series of results on multiple experiments that follow the principles outlined in~\cite{SilvaRF_MMSimNIMG2014}, including a summary of various controlled simulations on carefully crafted synthetic data, as well as comparisons with several algorithms.
\subsection{General Simulation Setup and Evaluation} \label{sec:simsetup}
In the following, we consider the problem of identifying statistically independent subspaces.
Thus, in all experiments, each subspace $\y_k$ is a random sample with $N$ observations, which are organized into a source matrix $\Y$.
A random mixing matrix $\A$ is then used to linearly mix subspace observations as $\x = \A\y + \e$, where $\e$ is additive sensor white noise and $\y$ is a column (i.e., an observation) of $\Y$.
The condition number of the mixing matrix $\A$ is defined as the ratio between its largest and smallest singular values.
In order to generate a matrix with prescribed condition number $c_m$, we first generate a $V_m \times C_m$ random Gaussian sample $\A$ with zero mean and unit variance.
The \ac{SVD} of $\A$ is then computed as $USV^\top$, with $\sigma_{\max} = \max(S)$ and $\sigma_{\min} = \min(S)$.
Finally, new singular values are defined as:
\begin{IEEEeqnarray}{CCl}
	\bar{S} &=& S + \frac{\sigma_{\max} - c_m \sigma_{\min}}{c_m-1} \mathrm{,}%
	\nonumber
\end{IEEEeqnarray}
and the resulting $\A = U\bar{S}V^\top$ will have the desired condition number $\textrm{cond}(\A) = c_m$.

The additive sensor white Gaussian noise $\e$ with zero mean and unit variance is scaled by a value $a$ in order to attain a prescribed \ac{SNR}.
The \ac{SNR} is defined as the power ratio between the noisy signal $\x$ and the noise $\e$.
The power of $\x$ is defined as its expected squared $L_2$-norm, i.e., ${P_{\x} = \mathbb{E}\left[ \x^\top \x \right]}$.
Substituting $\x = \A\y + a\e$, and using the identities $\mathbb{E}\left[ \z^\top \z \right] = \mathbb{E}\left[ \mathrm{tr}\left( \z \z^\top \right) \right] = \mathrm{tr}\left( \mathbb{E}\left[ \z \z^\top \right] \right)$, ${\mathbb{E}\left[ \e \e^\top \right] = \eye_{V_m}}$, and $\mathbb{E}\left[ \y \y^\top \right] = \eye_{C_m}$, it is easy to show that $P_{\e} = a^2 V_m$, and $P_{\x} = \mathrm{tr}\left( \A\A^\top \right) + a^2 V_m$, where $\mathrm{tr}(\cdot)$ is the trace operator.
Their ratio gives the SNR.
Then, %
\begin{IEEEeqnarray}{CCl}
	a &=& \sqrt{\frac{\mathrm{tr}\left( \A\A^\top \right)}{V_m \left(\mathrm{SNR}-1\right)}} \mathrm{,}%
	\nonumber
\end{IEEEeqnarray}
which we use to prescribe a SNR to synthetic data.
The equality $\mathrm{SNR} = 10^{\frac{\mathrm{SNRdB}}{10}}$ permits decibel (dB) specifications.

The presence of additive sensor noise $\e$ implies that $\A$ and $\y$ are recovered with different accuracies.
The quality with respect to $\A$ is evaluated using the normalized \ac{MISI}~\eqref{eq:MISI}, which extends the ISI~\cite{RN44,Macchi1995} to multiple datasets.
\begin{IEEEeqnarray}{rCl}
MISI(\mathbf{H}) &=& \frac{0.5}{K(K-1)}%
	\left[\sum_{i=1}^{K}\left(-1+\sum_{j=1}^{K}\frac{\left|h_{ij} \right|}{\mathop{\max }\limits_{k} \left|h_{ik} \right|}  \right) \right. %
	\nonumber%
	\\%
	&& \qquad \quad ~~~%
	\left. +\sum_{j=1}^{K}\left(-1+\sum _{i=1}^{K}\frac{\left|h_{ij} \right|}{\mathop{\max }\limits_{k} \left|h_{kj} \right|}  \right) \right] ~~~~%
\label{eq:MISI} 
\end{IEEEeqnarray}
where $\mathbf{H}$ is a matrix with elements $h_{ij} = \mathbf{1}^\top \abs{\PP_i \hat{\W} \A \PP_j^\top} \mathbf{1}$, with $(i,j) = 1 \ldots K$, i.e., the sum of absolute values from all elements of the interference matrix $\hat{\W} \A$ corresponding to subspaces $i$ and $j$, where $\hat{\W}$ is the solution being evaluated.
Likewise, the quality of the resulting $\y$ was evaluated using the \ac{MMSE}, which extends the \ac{MSE}~\cite{Nordhausen2011_PerfBSS_ISI_MD_MSE}:
\begin{IEEEeqnarray}{CCl}
	\mathrm{MMSE}(\R^{\hat{\y}\y}) &=& 2 - \frac{2}{K}\mathrm{tr}\left( \abs{\T \R^{\hat{\y}\y}} \right) \mathrm{,}%
	\label{eq:MMSE}
\end{IEEEeqnarray}
where $\R^{\hat{\y}\y}$ is the cross correlation between estimated and ground-truth sources, $\T = \LSAP(\B)$ is a permutation matrix that solves the \emph{\ac{LSAP}}~\cite{AssignmentProblems_Book}, and $\B$ is the \ac{LSAP} cost matrix with elements of the form $\B_{ij} = \sum_{k=1}^{K}\left(\eye_{jk}^{} - \abs{\R^{\hat{\y}\y}_{ik}}\right)$.
The \ac{LSAP} is easily solved using the Hungarian algorithm~\cite{AssignmentProblems_Book}.

\subsection{Numerical Optimization Strategy}\label{sec:numopt}
As discussed in the main paper, \ac{RE} can be either used as a data reduction technique or a constraint for the \ac{MISA} optimization procedure.
When, \ac{RE} is used for data reduction, the following optimization is performed:
\begin{IEEEeqnarray}{CCl}
	\min_{\B} && \>\> E(\B) \mathrm{,} \label{eq:MISAprob}%
\end{IEEEeqnarray}
where $E$ could be either $E = \frac{E_{\top}}{\x_{\mathrm{norm}}} = \frac{\mathrm{MSE} \left( \B^{\top}\B\x - \x \right)}{\x_{\mathrm{norm}}}$ or $E = \frac{E_{-}}{\x_{\mathrm{norm}}} = \frac{\mathrm{MSE} \left( \B^{\top}(\B\B^{\top})^{-1}\B\x - \x \right)}{\x_{\mathrm{norm}}}$.
In this instance, we reduce the data as $\Z = \B_\star\X$ and follow with unconstrained optimization of equation (3) in the main paper, treating $\Z$ as the data in order to estimate sources as $\Y = \W_{\mathrm{red}} \Z.$
The final model is then $\Y = \W_{\mathrm{red}} \B_\star \X$, and $\W = \W_{\mathrm{red}} \B_\star$.
The initial $\B_0$ is random, while the initial $\W_{\mathrm{red},0} = \eye$.
The precisions for the objective function and optimization variables (elements of $\W$) were set as TolFun = TolX = $\frac{10^{\floor{\log_{10} \norm{\nabla E(\B_0)}}}}{b}$ in MATLAB's Optimization Toolbox, where $b$ is a precision parameter.

When \ac{RE} is used as a constraint, we formulate the optimization problem as:
\begin{IEEEeqnarray}{CCl}
	\min_{\W} && \>\> \check{I}(\W) \\
	s.t.      && \>\> E(\W) \le \Delta, \nonumber %
\end{IEEEeqnarray}
where $\Delta$ is a user-specified threshold, and we initialize $\W_0 = \B_\star$ above to ensure $\W_0$ lies within the feasible region.
For fairness, all algorithms are initialized with the same $\B_0$.

Specific optimization parameters are described along with each experiment below.

All experiments were run on MATLAB R2014a.

\subsection{External Code References}
The numerical optimizer utilized to execute MISA is a barrier-type interior-point non-linear constraint optimizer that combines line search and trust region steps as described in~\cite[Ch. 19]{Nocedal2006_NumOpt}~\cite{Waltz2006_NLOpt_LS_TR_KNITRO}, and available in MATLAB's Optimization Toolbox as a function called fmincon.m.

The code for Infomax is described in~\cite{BellSejnowski1995Infomax,Amari1998_NaturalGrad}, and is available as a MATLAB function called icatb\_runica.m in~\cite{GIFT2015_v4a}.

The code for IVA-L is described in~\cite{KimT_IVA2006,Anderson2012_JBSSGaussianAlgPerf},
and is available as a MATLAB function called icatb\_iva\_laplace.m in~\cite{GIFT2015_v4a}.

The code for IVA-G is described in~\cite{Anderson2010IVAG,Anderson2012_JBSSGaussianAlgPerf},
and is available as a MATLAB function called icatb\_iva\_second\_order.m in~\cite{GIFT2015_v4a}.

The code for JBD-SOS is described in~\cite{LahatD2012_MICA_SOS}, and is available as a MATLAB function called jbd.m at \href{https://www.irit.fr/~Dana.Lahat/jbd.zip}{\url{https://www.irit.fr/~Dana.Lahat/jbd.zip}}.

The code for EST\_ISA was the same used in~\cite{LeQ2011_ISA}, which is available as a MATLAB function called isa\_est.m at \href{http://ai.stanford.edu/~quocle/video_release.tar.gz}{\url{http://ai.stanford.edu/~quocle/video\_release.tar.gz}}.
Here, we used the default function parameters.

\subsection{Synthetic Data Simulations}\label{sec:synthdata}
Below we present a series of synthetic data experiments to investigate the performance of the proposed \ac{MISA} algorithm in different scenarios.

\vfill
\pagebreak
\subsubsection{ICA 1 ($V>N$)}\label{sec:ICA1}
\begin{figure}[!h]
	\centering
	\captionsetup[subfloat]{justification=centering}
		\subfloat[Fixed \ac{SNR}, varying condition number]{\includegraphics[height=14.2cm,trim={0 0 1cm 0},clip]{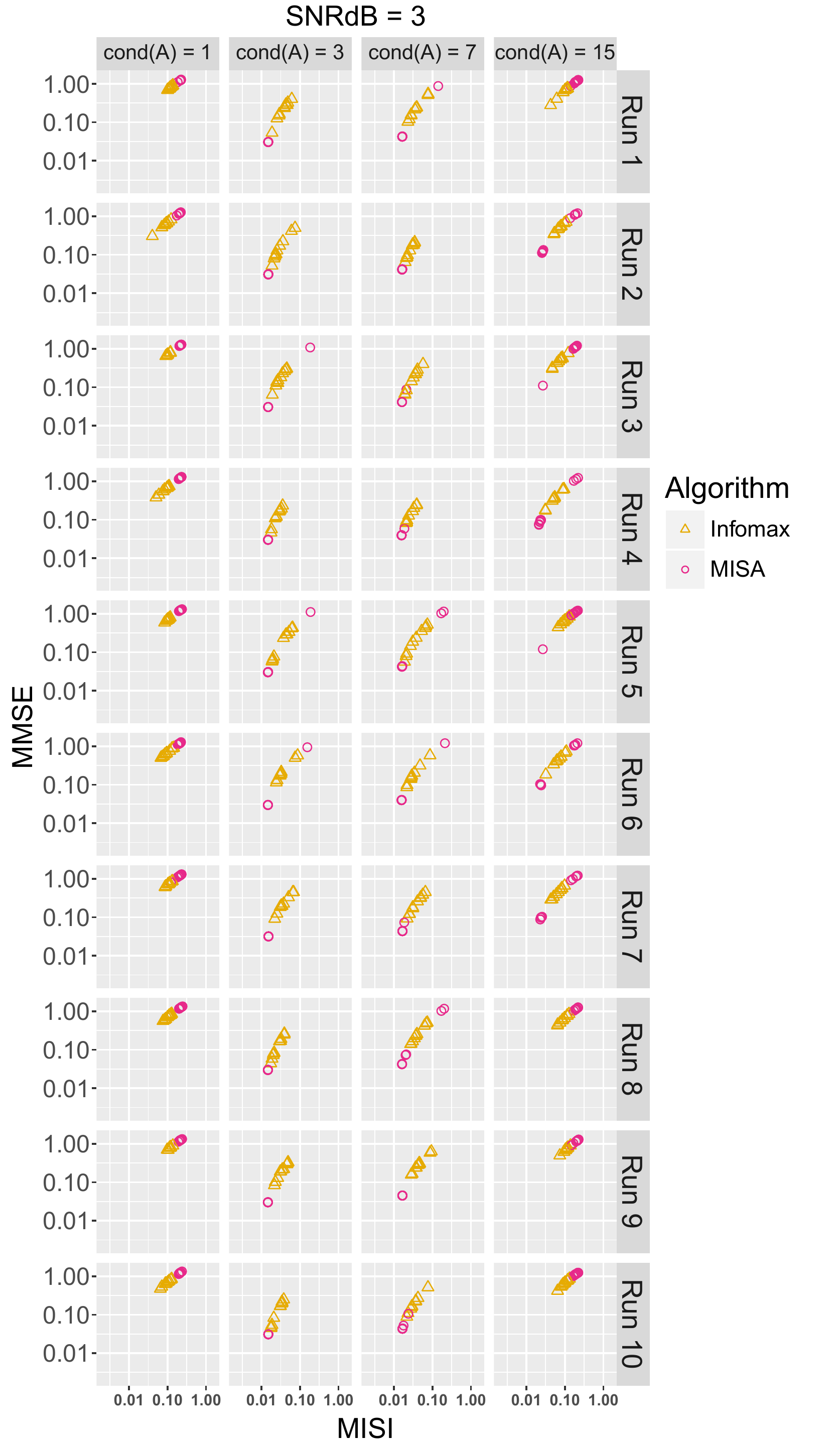}%
			\label{fig:ICASim_part_a}}
		\hfil
		\subfloat[Fixed condition number, varying \ac{SNR}]{\includegraphics[height=14.2cm,trim={0 0 3.75cm 0},clip]{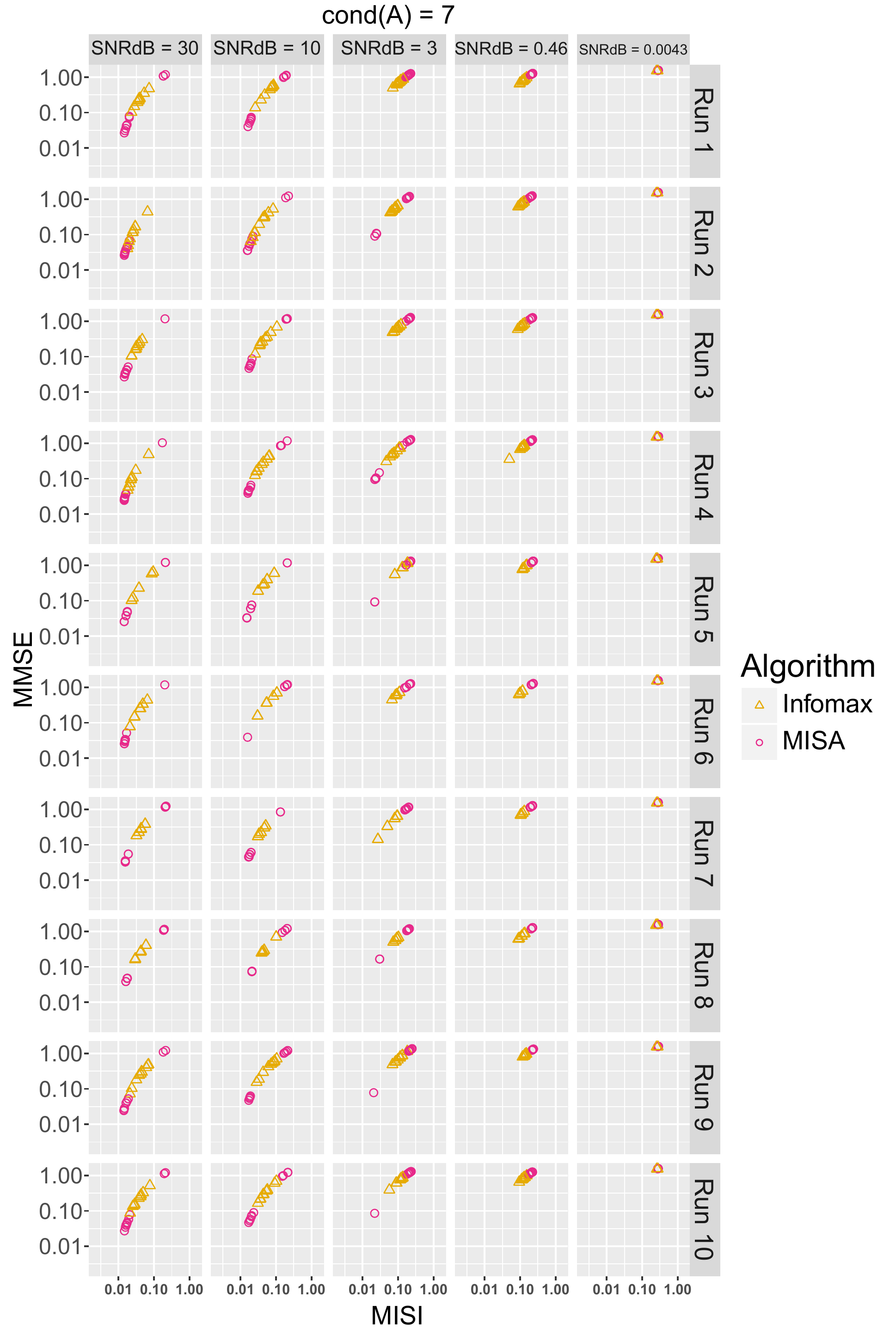}%
			\label{fig:ICASim_part_b}}
	\caption{\textbf{ICA 1: $V>N$}. Part (a): Under low \ac{SNR}, MISA outperforms Infomax when $1 < \cond{\A} < 15$. Part (b): At $\mathrm{SNRdB} > 3$, MISA outperforms Infomax more often than not. Each Instance and condition uses a different random mixing matrix $\A$ and source sample $\y$ for data generation. For each $\A$, the algorithms run ten times from different initial $\W_0$. MISI measures the quality of the mixing matrix estimate while MMSE measures the quality of the  source estimates (lower is better in both cases). Following~\cite{SilvaRF_MMSimNIMG2014}, \ac{MISI} values below 0.1 indicate good performance. NOTE: The stopping condition for the norm of the \ac{PRE} gradient was more strict (lower) in Part (a) than in Part (b), meaning Part (b) relied less on good \ac{PRE} performance.}
	\label{fig:ICA1}
\end{figure}

Here we assess the effect of additive sensor noise and condition number in a moderately large ICA problem with rectangular mixing matrix $\A$ and fairly low number of observations $N$.
The experiment was setup with $M=1$ dataset containing $\bar{C}=C_1=75$ sources organized into $K=\bar{C}=75$ one-dimensional subspaces, and $N = 3500$ observations sampled independently from a Laplace distribution.
In each of the ten instances of this experiment, a new, unique $(V \times C)$ rectangular mixing matrix $\A$ ($V=8000$) was randomly generated.
For each instance, ten runs were performed, each with a different random row-orthogonal $\W_0$ initialization (thus, 100 runs per experimental condition).
The experiments were broken into two parts:
\begin{IEEEdescription}[\IEEEsetlabelwidth{Part b)}]
	\item[Part a)] Gaussian white noise $\e$ was added to the mixtures to yield a $\mathrm{SNRdB} = 3$, while varying the condition number: $\mathrm{cond}(\A) \in \{1,3,7,15\}$.
	\item[Part b)] the condition number was fixed at $\mathrm{cond}(\A) = 7$, while Gaussian white noise $\e$ was added to the mixtures with varying ${\mathrm{SNRdB} \in \{30,10,3,0.4,0.004\}}$.
	These \ac{SNR} values correspond to signal power~:~noise power ratios of $[999:1, 99:1, 1:1, 1:99, 1:999]$.
\end{IEEEdescription}
Here, \ac{PRE} was employed for data-reduction, followed by either Infomax or \ac{MISA} for final separation (Fig.~\ref{fig:ICA1}).

\textbf{Optimization parameters}:
For the data reduction, we utilized MATLAB's Optimization Toolbox function fmincon.m.
The bounds on the values taken by each element of $\W$ were set to -100 and 100, their typical value was set to TypicalX = 0.1, the maximal number of objective function evaluations was set to 50000, the maximum number of iterations was set to 10000, the initial barrier parameter was set to 0.1, and the number of past gradients utilized for L-BFGS was set to 5. TolX and TolFun were set as described in Section~\ref{sec:numopt} above, with $b = 80$ for part a), and $b = 4$ for part b).
For MISA, we utilized the same parameters, except: TolFun = $10^{-4}$, TolX = $10^{-9}$, the initial barrier parameter was set to 1, and the number of past gradients utilized for L-BFGS was set to 20.
For Infomax, we utilized all the defaults in~\cite{GIFT2015_v4a} and ``sphering'' was set to ``on''.

The experiments in part a) were setup with a challenging low \ac{SNR} (noise power on the same order as the signal power).
Under this condition, we notice that increasing the condition number above 7 significantly reduces the performance of both Infomax (using relative gradient)~\cite{BellSejnowski1995Infomax} and MISA.
Also, as predicted in Section III-B of the main paper, when $\mathrm{cond}(\A) = 1$ (i.e., it is column-orthogonal) and $\mathbb{E}[\y\y^{\top}] = \eye$, $\W$ is expected to be row-orthogonal and, thus, the proposed \ac{PRE} induces reduced performance ($\MISI > 0.1$).
Otherwise, both Infomax and \ac{MISA} perform very well ($\MISI < 0.1$) when ${1 < \cond{\A} < 15}$, with MISA outperforming Infomax in this range.

For part b), the performance of \ac{PRE} prior to Infomax/MISA was worse due to a less strict requirement in the gradient norm stopping condition.
As a result, more noise leaked into the Infomax/MISA stage, making the identification of the sources harder.
Overall, both Infomax and \ac{MISA} performed well at high SNR, with MISA outperforming Infomax more often than not.
Altogether, the results from a) and b) suggest the performance of \ac{PRE} as a data-reduction tool has a significant effect on the \ac{ICA} decomposition.

\vfill

\subsubsection{IVA 1 ($V<N$, $V=C$)}
\begin{figure}[!ht]
	\centering
	\includegraphics[width=.75\linewidth]{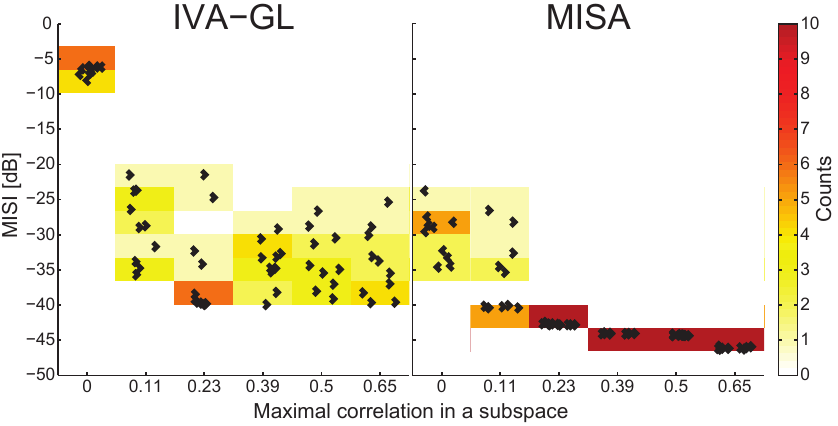}
	\caption{\textbf{IVA 1:} $V<N$, $V=C$. MISA performs well even at low correlation levels and is highly stable when within subspace correlations are larger than 0.2. Each run uses a different random mixing matrix $\A$ and source sample $\y$ for data generation. For each $\A$, the algorithms run once from a different initial $\W_0$. MISI measures the quality of the mixing matrix estimate (lower is better). Following~\cite{SilvaRF_MMSimNIMG2014}, MISI values below $0.1$ (here, $-20$dB) indicate good performance, and values below $0.01$ (here, $-40$dB) indicate excellent performance.}
	\label{fig:IVAsqcond3_AAAI}
\end{figure}

In this experiment, we want to assess the performance in an \ac{IVA} problem when no data reduction is required (i.e., $V=C$), noise is absent, and the number of observations $N$ is abundant.
The experiment was setup with $M=10$ datasets, each containing $C_m=16$ sources for a total of $K=C_m=16$ ten-dimensional subspaces, and $N = 32968$ observations sampled independently from a multivariate Laplace distribution ($\bm{\psi}_L = [\frac{1}{2},1,1]$).
The selected autocorrelation function within subspaces followed an inverse exponential, leading to a Toeplitz structure in the subspace correlation matrix $\R^\y_k$.
The maximal correlation in a subspace varied from $0$ to $0.65$, and $\R^\y_k$ was different for each subspace in the same experimental run.
In each of the ten runs for each of the six correlation cases considered, a new, unique square mixing matrix $\A$ ($V=C$) with condition number $\mathrm{cond}(\A) = 3$ was randomly generated.
Likewise, each run was initialized with a different random row-orthogonal $\W_0$ (60 runs total).
Finally, we compared the performance of IVA-GL (i.e., IVA-G with Newton step~\cite{Anderson2010IVAG} as an initializer for IVA-L with relative gradient~\cite{KimT_IVA2006}) and MISA in each setting (Fig.~\ref{fig:IVAsqcond3_AAAI}).

\textbf{Optimization parameters}:
No data reduction was employed.
For MISA, we utilized MATLAB's Optimization Toolbox function fmincon.m.
The bounds on the values taken by each element of $\W$ were set to -100 and 100, their typical value was set to TypicalX = 0.1, the maximal number of objective function evaluations was set to 10000, the maximum number of iterations was set to 10000, the initial barrier parameter was set to 0.1, and the number of past gradients utilized for L-BFGS was set to 1. Also, TolX = $10^{-30}$ and TolFun = $10^{-4}$.
For IVA-G, we utilized all the defaults in~\cite{GIFT2015_v4a} and ``opt\_approach'' was set to ``newton''.
For IVA-L, we utilized all the defaults in~\cite{GIFT2015_v4a}, ``whiten'' was set to ``false'',  and ``maxIter'' was set to 1024, and ``InitialW'' was set $\W_G$ (the unmixing matrix estimated with IVA-G).

The striking feature observed here is that the performance of IVA-GL is much more variable than that from MISA, especially with high correlation within the subspaces.
MISA performs well even at low within-subspace correlation levels and is highly stable when these correlations are larger than 0.2.

\newpage
\subsubsection{IVA 2 ($V<N$)}
\begin{figure}[!ht]
	\centering
	\captionsetup[subfloat]{justification=centering}
	\subfloat[Fixed \ac{SNR}, varying condition number]{\includegraphics[height=14.8cm,trim={0 0 1cm 0},clip]{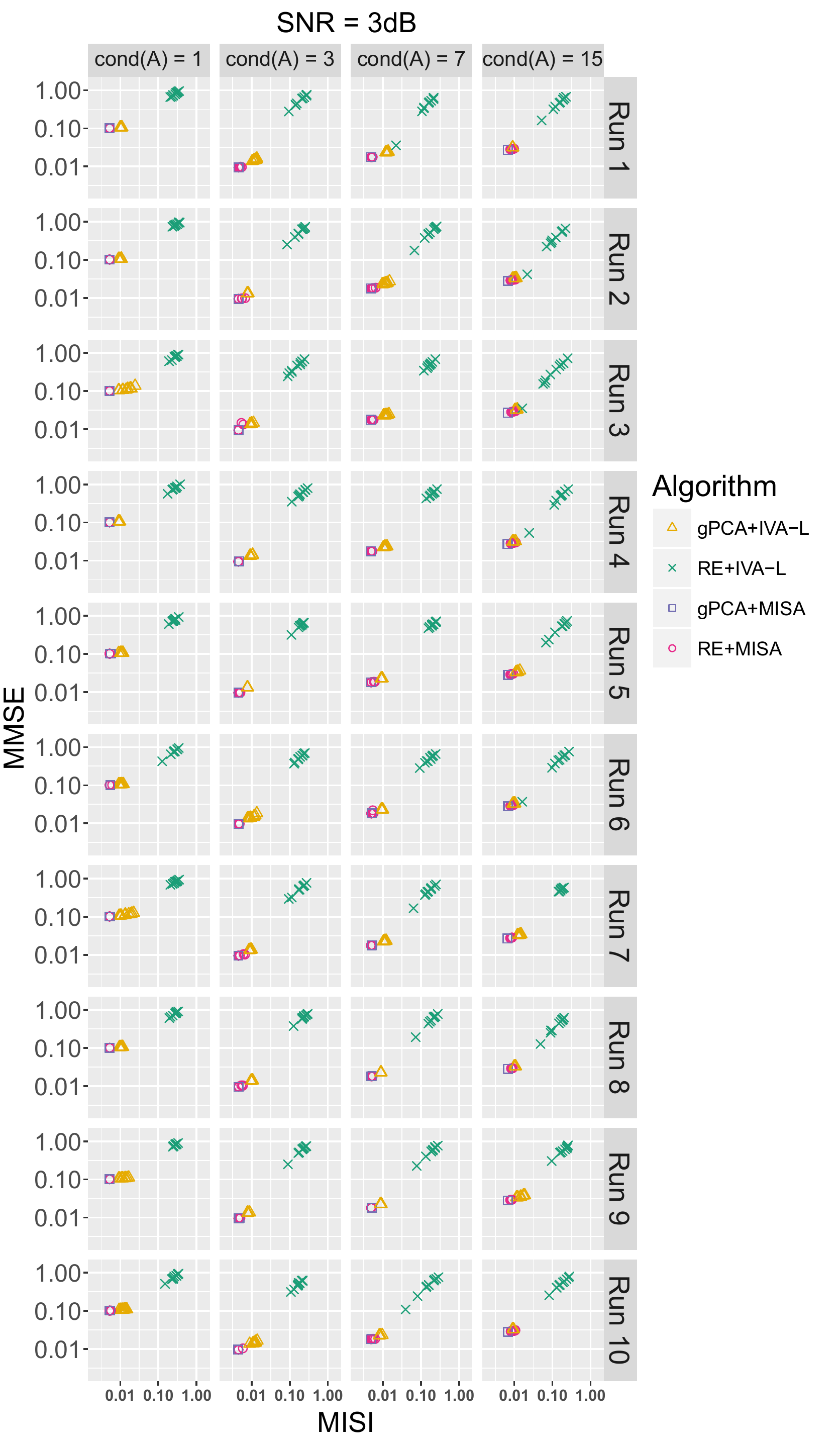}%
		\label{fig:IVASim_part_a}}
	\hfil
	\subfloat[Fixed condition number, varying \ac{SNR}]{\includegraphics[height=14.8cm,trim={0 0 3.75cm 0},clip]{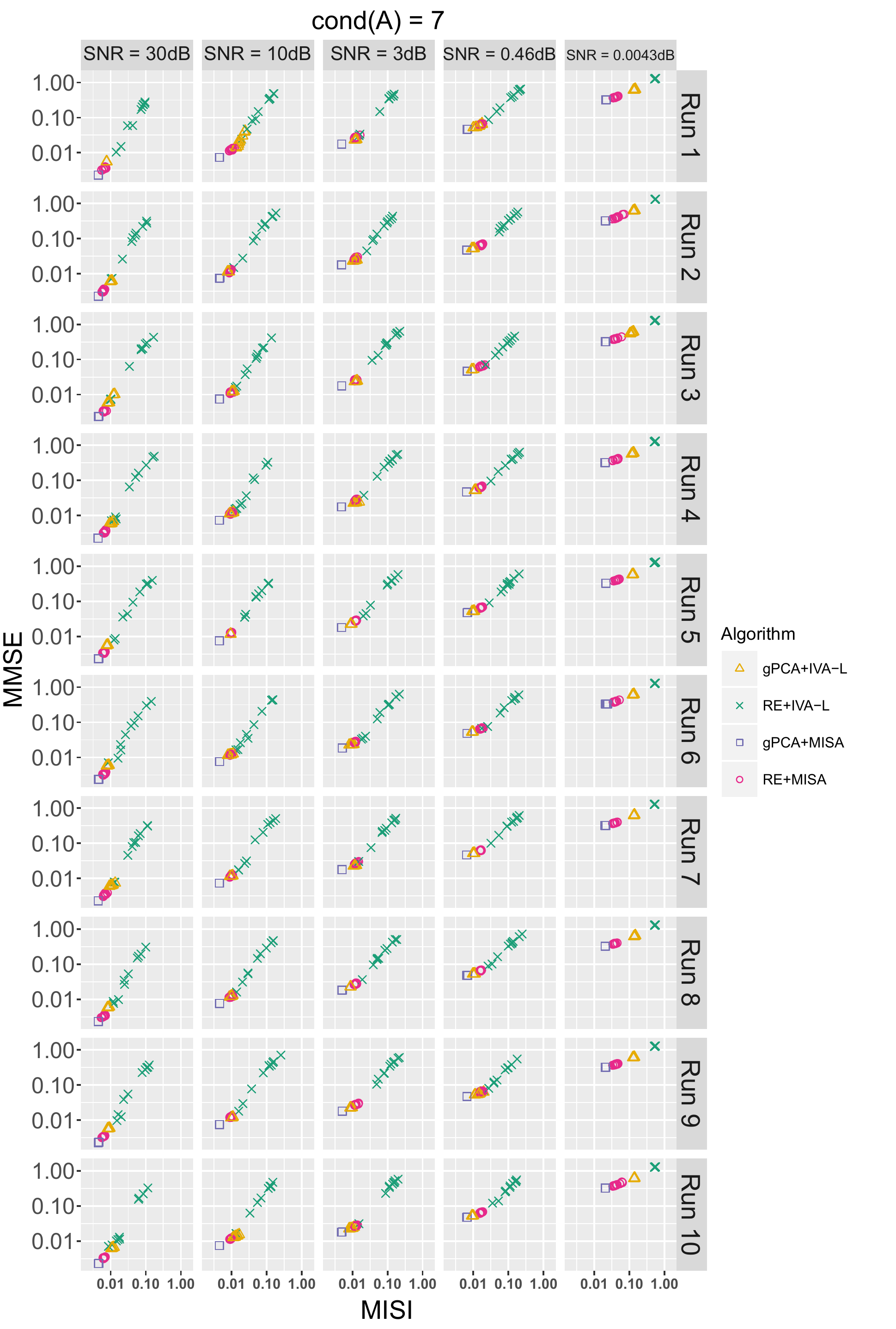}%
		\label{fig:IVASim_part_b}}
	\caption{\textbf{IVA 2:} $V<N$. Part (a): reducing the data with either \ac{gPCA} or \ac{PRE} produced equivalent results in this large $N$ scenario. Overall, increasing the condition number had a fairly small detrimental effect on the performance of both IVA-L and MISA. Part (b): Overall, both IVA-L and \ac{MISA} performed very well at mild SNR levels. The performance of MISA on extremely noisy scenarios is remarkable ($0.1< \mathrm{MISI} < 0.01$), irrespective of preprocessing with \ac{PRE} or \ac{gPCA}. Each Instance and condition uses a different random mixing matrix $\A$ and source $\y$ for data generation. For each $\A$, the algorithms run ten times from different initial $\W_0$.  MISI measures the quality of the mixing matrix estimate while MMSE measures the quality of the  source estimates (lower is better in both Following~\cite{SilvaRF_MMSimNIMG2014}, MISI values below 0.1 indicate good performance. NOTE: The stopping condition for the norm of the \ac{PRE} gradient was more strict (lower) in Part (a) than in Part (b), meaning Part (b) relied less on good \ac{PRE} performance.}
	\label{fig:IVA2}
\end{figure}

Here we assess the effect of additive sensor noise and condition number in a larger IVA problem with rectangular mixing matrix $\A$ and an abundant number of observations $N$.
The experiment was setup with $M=16$ datasets, each containing $C_m=75$ sources for a total of $K=C_m=75$ sixteen-dimensional subspaces, and ${N = 66000}$ observations sampled independently from a multivariate Laplace distribution.
Similarly to experiment IVA 1, an inverse exponential autocorrelation function with maximal correlation varying from $0$ to $0.5$ was selected.
In each of the ten instances of this experiment, a new, unique $(V \times C)$ rectangular mixing matrix $\A$ ($V=250$) was randomly generated.
For each instance, ten runs were performed, each with a different random row-orthogonal $\W_0$ initialization (thus, 100 runs per experimental condition).
The experiments were broken into two parts, a) and b), exactly as described in experiment ICA 1 (Section~\ref{sec:ICA1}).
We also evaluate the performance of data-reduction by \ac{gPCA}~\cite{2016Rachakonda_MemEffGroupPCA} or \ac{PRE}, followed by either IVA-L~\cite{KimT_IVA2006} or \ac{MISA} (Fig.~\ref{fig:IVA2}).

\textbf{Optimization parameters}:
For the data reduction, we utilized MATLAB's Optimization Toolbox function fmincon.m.
The bounds on the values taken by each element of $\W$ were set to -100 and 100, their typical value was set to TypicalX = 0.1, the maximal number of objective function evaluations was set to 50000, the maximum number of iterations was set to 10000, the initial barrier parameter was set to 0.1, and the number of past gradients utilized for L-BFGS was set to 5. TolX and TolFun were set as described in Section~\ref{sec:numopt} above, with $b = 80$ for part a), and $b = 4$ for part b).
For MISA, we utilized the same parameters, except: TolFun = $10^{-4}$, TolX = $10^{-9}$, the initial barrier parameter was set to 1, and the number of past gradients utilized for L-BFGS was set to 10.
For IVA-L, we utilized all the defaults in~\cite{GIFT2015_v4a} and ``whiten'' was set to ``false'', ``alpha0'' was set to 1, ``terminationCriterion'' was set to ``ChangeInW'', and ``maxIter'' was set to $min\left(10000, 4 \cdot MISAiter\right)$, where $MISAiter$ is the number of iterations until convergence for MISA on the same problem, from the same starting point.

Under a challenging low \ac{SNR} condition, the experiments in Part a) show that both \ac{gPCA} and \ac{PRE} provide good quality data-reduction when observations are abundant and $V < N$.
However, for PRE+IVA-L, the results were unsatisfactory because  we imposed a maximum number of iterations on IVA-L equal to four times that of MISA in order to compensate for its very long convergence time.
As a result, PRE+IVA-L stops prematurely before converging.
Apart from that, reducing the data with either \ac{gPCA} or \ac{PRE} produced equivalent results in this large $N$ scenario.
Note that this was not the case for gPCA on a separate experiment with $V > N$ (not shown).
Overall, increasing the condition number had a fairly small detrimental effect on the performance of both IVA-L and MISA.
Also, as we predicted in Section III-B of the main paper, the proposed \ac{PRE} induced a reduction in performance when $\cond{\A}=1$, although this was only noticeable in the MMSE measure.
Consequently, while the estimation quality of $\A$ was retained in this case, the same did not occur with the source estimates $\y$.
We believe this is also in part due to the way MMSE is computed, which is an adaptation of the MSE measure for \ac{SDU} problems.

For part b), the performance of \ac{PRE} is reduced due to a less strict requirement in the gradient norm stopping condition.
As a result, more noise leaked into the IVA-L/MISA stage, making the identification of the sources harder.
Overall, both IVA-L and \ac{MISA} performed very well at mild SNR levels, with MISA outperforming IVA-L at extremely low SNR.
The performance of MISA on extremely noisy scenarios is remarkable ($0.1< \mathrm{MISI} < 0.01$), irrespective of preprocessing with \ac{PRE} or \ac{gPCA}.
We believe this is due to the fairly small number of datasets ($M=16$).
In experiments with larger number of datasets (not shown), we observed an interaction between $M$ and $N$ on the performance.
Those experiments have suggested that the performances of IVA-L and MISA deteriorate with larger $M$ and lower $N$.
Further study is required to understand these performance limits.

\vfill
\pagebreak
\subsubsection{ISA 1 and 2 ($V<N$, $V=C$)}
\begin{figure}[!ht]
	\centering
	\captionsetup[subfloat]{justification=centering}
	\subfloat[ISA 1: Non-correlated sources]{\includegraphics[width=.75\linewidth]{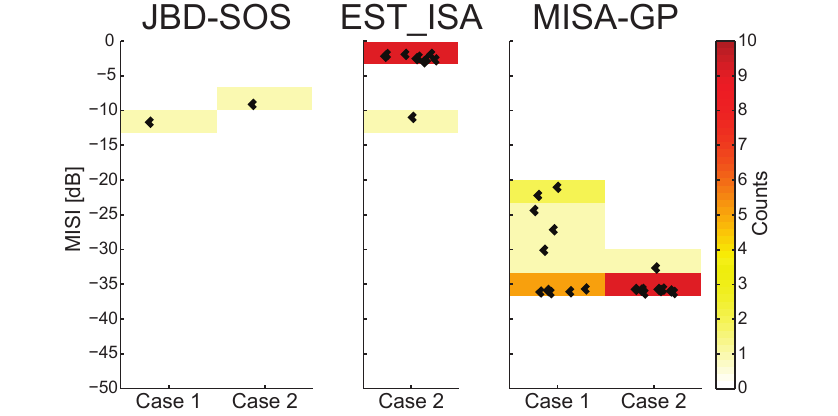}%
		\label{fig:ISA_nocorr_sqcond3_AAAI}}
	
	\subfloat[ISA 2: Correlated sources]{\includegraphics[width=.75\linewidth]{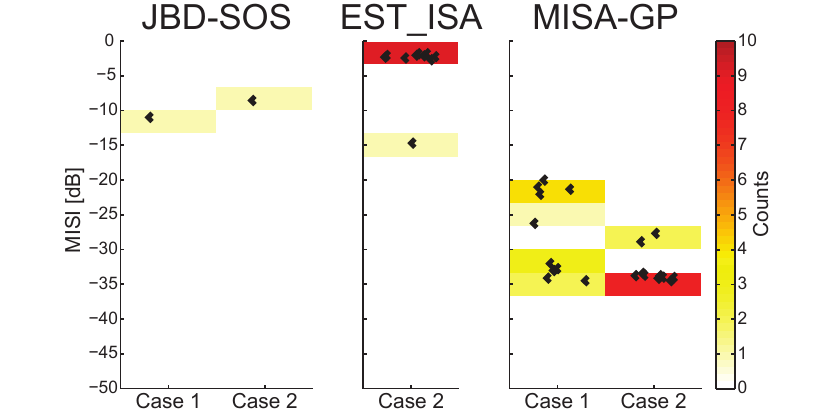}%
		\label{fig:ISA_corr_sqcond3_AAAI}}
	\caption{\textbf{ISA 1 and ISA 2:} $V<N$, $V=C$. Even when within-subspace correlations are present (panel (b)) MISA-GP is the only method with good performance, which highlights the large benefit of the permutation approach we proposed to escape local minima. Each run uses a different random mixing matrix $\A$ and source sample $\y$ for data generation. For each $\A$, the algorithms run once from a different initial $\W_0$. MISI measures the quality of the mixing matrix estimate (lower is better). Following~\cite{SilvaRF_MMSimNIMG2014}, MISI values below 0.1 (here $-20$dB) indicate good performance. All algorithms are given the same initial $\W_0$ except JBD-SOS, which always starts from $\W_0 = \eye$.}
	\label{fig:ISA1_2}
\end{figure}

In these experiments, we want to assess the performance in \ac{ISA} problems when no data reduction is required (i.e., $V=C$), noise is absent, and the number of observations $N$ is abundant. 
The ISA 1 experiment was setup with $M=1$ dataset containing $\bar{C}=C_1=28$ sources organized into $K=7$ $d_k$-dimensional subspaces (see $d_k$ values below), with $N = 32968$ observations sampled independently from a multivariate Laplace distribution ($\bm{\psi}_L = [\frac{1}{2},1,1]$).
Correlation was absent within all subspaces (i.e., only \ac{HOS} dependence was present).
In each of the ten runs for each of the two cases below, a new, unique square mixing matrix $\A$ ($V=C$) with condition number $\mathrm{cond}(\A) = 3$ was randomly generated.
Likewise, each run was initialized with a different random row-orthogonal $\W_0$ (20 runs total).
The experiments were broken into two cases:
\begin{IEEEdescription}[\IEEEsetlabelwidth{Part b)}]
	\item[Case 1:] subspace sizes $d_k = k$, $k = [1,2,3,4,5,6,7]$.
	\item[Case 2:] subspace sizes $d_k=4$ for all subspaces.
\end{IEEEdescription}
The ISA 2 experiment was identical to experiment ISA 1, except with a within-subspace correlation structure like the one described in experiment IVA 1, with maximal correlation varying from $0.2$ to $0.75$.
We also evaluate the performance of JBD-SOS~\cite{LahatD2012_MICA_SOS}, EST\_ISA (in Case 2 only)~\cite{LeQ2011_ISA,HyvarinenESANN06} and MISA-GP in each setting (Fig.~\ref{fig:ISA1_2}).

\textbf{Optimization parameters}:
No data reduction was employed.
For MISA-GP, we utilized MATLAB's Optimization Toolbox function fmincon.m.
The bounds on the values taken by each element of $\W$ were set to -100 and 100, their typical value was set to TypicalX = 0.1, the maximal number of objective function evaluations was set to 10000, the maximum number of iterations was set to 10000, the initial barrier parameter was set to 0.1, and the number of past gradients utilized for L-BFGS was set to 1. Also, TolX = $10^{-30}$ and TolFun = $10^{-4}$.
For JBD-SOS, we provided 500 data covariance matrices as input (computed after \ac{PRE} data reduction), ``threshold'' was set to $10^{-9}$, and ``max\_sweep'' was set to 512.

The striking feature observed here is that the performance of both JBD-SOS and EST\_ISA is very poor in all cases, even when within-subspace correlations are present (ISA 2, panel (b)).
MISA-GP is the only method with good performance, which highlights the large benefit of the permutation approach we proposed to escape local minima.

\newpage
\subsubsection{ISA 3 ($V>N$)}
\begin{figure}[!h]
	\centering
	\captionsetup[subfloat]{justification=centering}
	\subfloat[Fixed \ac{SNR}, varying condition number]{\includegraphics[height=14.8cm,trim={0 0 1cm 0},clip]{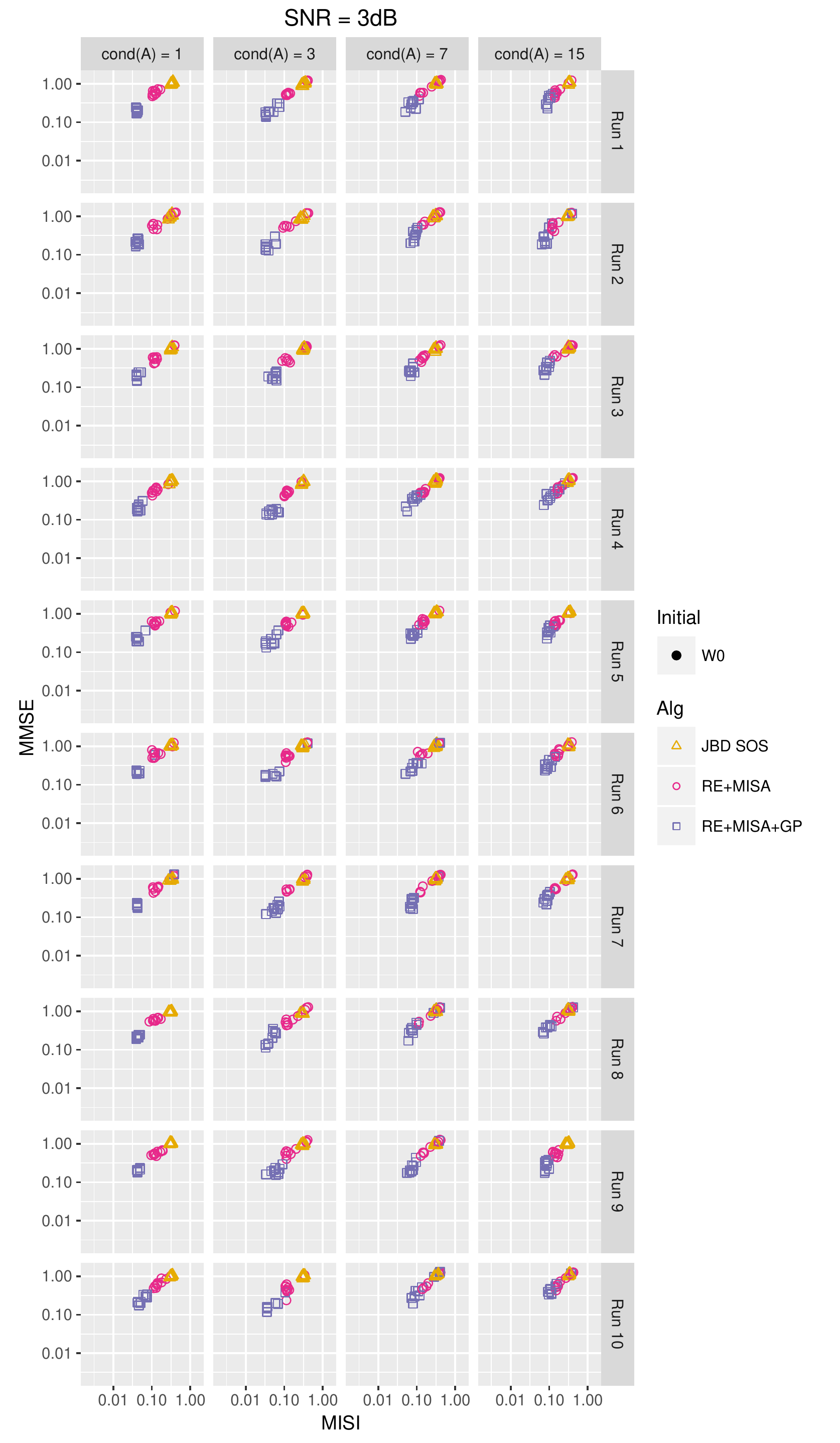}%
		\label{fig:ISASim_part_a}}
	\hfil
	\subfloat[Fixed condition number, varying \ac{SNR}]{\includegraphics[height=14.8cm,trim={0 0 3.75cm 0},clip]{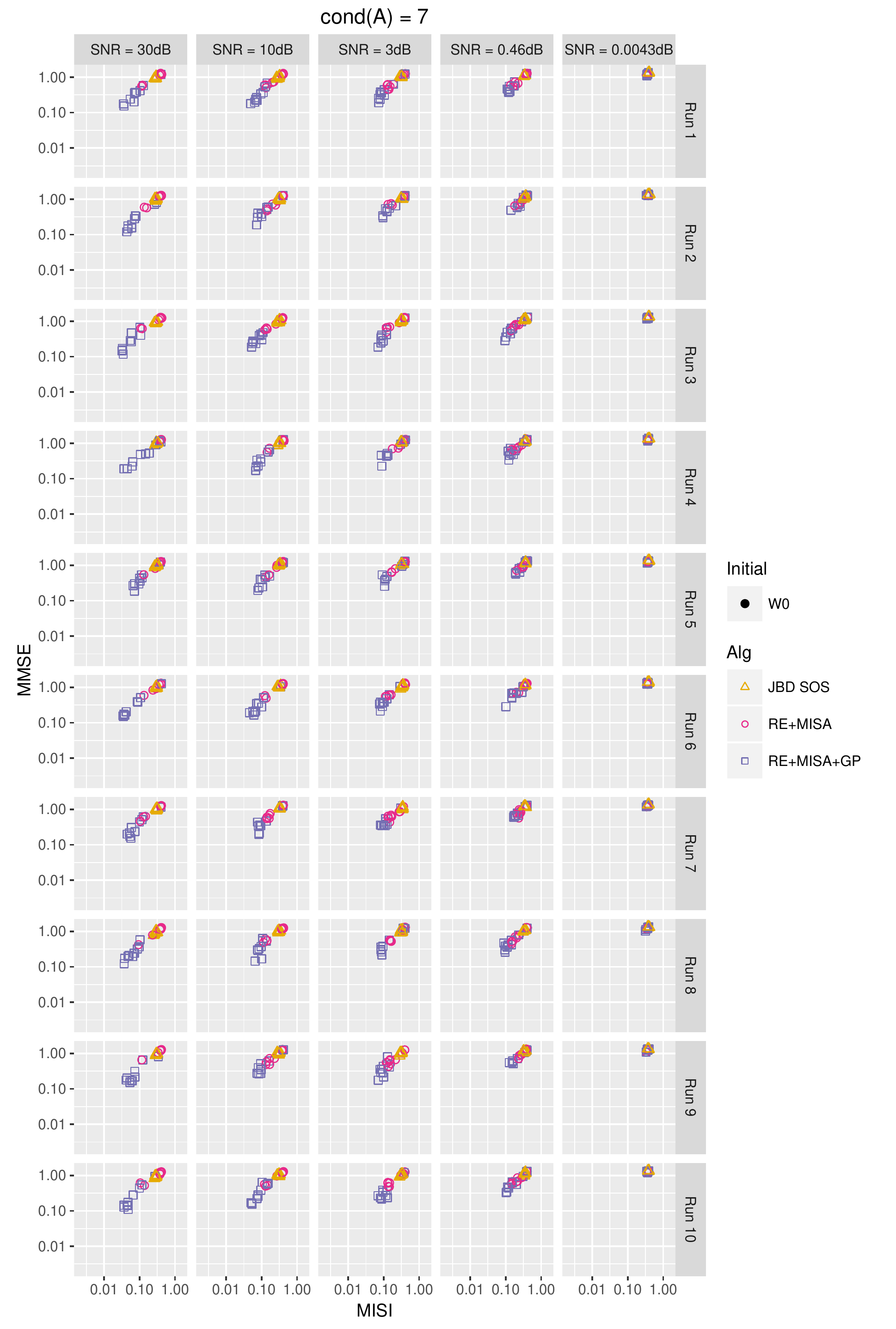}%
		\label{fig:ISASim_part_b}}
	\caption{\textbf{ISA 3: $V>N$}. Part (a): JDB-SOS and MISA fail in virtually all cases ($\mathrm{MISI} > 0.1$). Inclusion of greedy permutations (GP) enables MISA-GP to perform well in terms of \ac{MISI}. Part (b): MISA-GP performed well at mild-to-high SNR levels ($\mathrm{SNRdB} > 0.46$). Each Instance and condition uses a different random mixing matrix $\A$ and source $\y$ for data generation. For each $\A$, the algorithms run ten times from different initial $\W_0$.  MISI measures the quality of the mixing matrix estimate while MMSE measures the quality of the  source estimates (lower is better in both cases), though MMSE is an unreliable measure because \ac{ISA} problems are invariant to linear transformations within subspaces~\cite{CardosoMICA98}. Following~\cite{SilvaRF_MMSimNIMG2014}, MISI values below 0.1 indicate good performance. NOTE: The stopping condition for the norm of the \ac{PRE} gradient was more strict (lower) in Part (a) than in Part (b), meaning Part (b) relied less on good \ac{PRE} performance.
	\label{fig:ISA3}}
\end{figure}

Here we assess the effect of additive sensor noise and condition number in a mildly large ISA problem with rectangular mixing matrix $\A$ and a fairly low number of observations $N$.
The experiment was setup with $M=1$ dataset containing $\bar{C}=C_1=51$ sources organized into $K=18$ $d_k$-dimensional subspaces, with ${d_k = [1:5, 5:1, 1:5, 2, 2, 2]}$ (where $1:5$ means ``one through 5'') and $N = 5250$ observations sampled independently from a multivariate Laplace distribution.
Similarly to experiment IVA 1, an inverse exponential autocorrelation function with maximal correlation varying from $0$ to $0.5$ was selected.
In each of the ten instances of this experiment, a new, unique $(V \times C)$ rectangular mixing matrix $\A$ ($V=8000$) was randomly generated.
For each instance, ten runs were performed, each with a different random row-orthogonal $\W_0$ initialization (thus, 100 runs per experimental condition).
The experiments were broken into two parts, a) and b), exactly as described in experiment ICA 1 (Section~\ref{sec:ICA1}).
We also evaluate the performance of JBD-SOS, MISA, and MISA-GP in each setting (Fig.~\ref{fig:ISA3}).

\textbf{Optimization parameters}:
For the data reduction, we utilized MATLAB's Optimization Toolbox function fmincon.m.
The bounds on the values taken by each element of $\W$ were set to -100 and 100, their typical value was set to TypicalX = 0.1, the maximal number of objective function evaluations was set to 50000, the maximum number of iterations was set to 10000, the initial barrier parameter was set to 0.1, and the number of past gradients utilized for L-BFGS was set to 5. TolX and TolFun were set as described in Section~\ref{sec:numopt} above, with $b = 80$ for part a), and $b = 4$ for part b).
For MISA-GP, we utilized the same parameters, except: TolFun = $10^{-4}$, TolX = $10^{-9}$, the initial barrier parameter was set to 1, and the number of past gradients utilized for L-BFGS was set to 15.
For JBD-SOS, we provided 500 data covariance matrices as input (computed after \ac{PRE} data reduction), ``threshold'' was set to $10^{-6}$, and ``max\_sweep'' was set to 512.

Under a challenging low \ac{SNR} condition, the experiments in Part a) show that both JDB-SOS and MISA fail in virtually all cases ($\mathrm{MISI} > 0.1$).
Inclusion of greedy permutations (GP) enables MISA-GP to perform well in terms of \ac{MISI}.
Overall, increasing the condition number had a fairly small detrimental effect on the performance of MISA-GP, but enough to cause a few failing runs at $\cond{\A} \ge 7$.
Here, we note that the MMSE is an unreliable measure because \ac{ISA} problems are invariant to any linear transformation within subspaces~\cite{CardosoMICA98}, meaning that sources within subspaces are not uniquely identifiable even if the subspaces are well separated.
This also explains why the proposed \ac{PRE} did not induce a reduction in performance when $\cond{\A}=1$ as it did in the previous \ac{ICA} and \ac{IVA} experiments.

For Part b), the performance of \ac{PRE} is mostly unchanged by the less strict requirement in the gradient norm stopping condition.
Overall, MISA-GP performed well at mild-to-high SNR levels ($\mathrm{SNRdB} > 0.46$).
Again, the inclusion of greedy permutations enabled a successful performance for MISA-GP.

\subsection{Optimization parameters for Hybrid Data Experiments}
\subsubsection{Single-Subject Data and Multimodal IVA of sMRI, fMRI, and FA}
For the data reduction with PRE, estimating $\A$ from $\hat{\W}^{-}$, we utilized MATLAB's Optimization Toolbox function fmincon.m.
The bounds on the values taken by each element of $\W$ were set to -100 and 100, their typical value was set to TypicalX = 0.1, the maximal number of objective function evaluations was set to 50000, the maximum number of iterations was set to 10000, the initial barrier parameter was set to 0.1, and the number of past gradients utilized for L-BFGS was set to 5. TolX and TolFun were set as described in Section~\ref{sec:numopt} above, with $b = 4$, or as $10^{-8}$ (whichever was smallest).
This optimization was performed on each dataset separately.

For MISA, we utilized the same parameters, except: TolFun = TolX = $10^{-9}$, the initial barrier parameter was set to 1, and the number of past gradients utilized for L-BFGS was set to 10.

\vspace{8pt}
\subsubsection{Multimodal MISA of fMRI, and EEG}
For the initialization of $\W_0$ we used simple RE, estimating $\A$ from $\hat{\W}^{\top}$ and utilizing MATLAB's Optimization Toolbox function fmincon.m.
The bounds on the values taken by each element of $\W$ were set to -100 and 100, their typical value was set to TypicalX = 0.1, the maximal number of objective function evaluations was set to 50000, the maximum number of iterations was set to 10000, the initial barrier parameter was set to 0.1, and the number of past gradients utilized for L-BFGS was set to 5. TolX and TolFun were set as described in Section~\ref{sec:numopt} above, with $b = 4$, or as $10^{-8}$ (whichever was smallest).
This optimization was performed on each dataset separately.

For MISA-GP, we utilized the same parameters, except: TolFun = TolX = $10^{-9}$, the initial barrier parameter was set to 1, and the number of past gradients utilized for L-BFGS was set to 10.

\newpage
\section{Automatic Differentiation}\label{MISA2}

The \ac{AD} is a technique for automatically differentiating programs or algorithms without need for analytical derivations nor finite differencing.
It comprises two major approaches: forward and reverse differentiation.
In the forward mode, functions are compositions of a finite set of basic operations.
The derivatives of each basic operation are known and, thus, can be combined via chain rule to yield the final gradient.
In the reverse mode, only the function is computed in a forward fashion, not the derivatives.
After that, a reverse traversal of the composite operations propagates the derivatives backward from the output by supplementing the output with adjoint partial derivative terms in a compounding fashion.
In either case, the chain rule is the core engine for differentiation of function compositions.
Here, we utilize the reverse mode as it yields the gradient with respect to all variables (the elements of $\W$) in a single reverse sweep.

Our use of \ac{AD} in unconventional, however.
Rather than applying AD to a program or algorithm that evaluates equation (9) %
for a given $\W$ at every iteration of the numerical optimization, we apply it \emph{offline} in order to obtain an analytical form of the gradient with respect to $\W$.
This allows us to identify simplifications that lead to a more efficient implementation of the gradient.
This is also less susceptible to numerical instability.

\subsection{Basic Reverse Mode Operations}\label{MISA3}
In order to facilitate the understanding of how \ac{AD} works in reverse mode, a few basic rules and examples are presented.
Firstly, consider the case of $J(\W) = \x^\top\W\x$.
The goal is to obtain $\frac{\partial J}{\partial \W}$. Thus, forward evaluation of $J$ follows standard operation precedence, leading to the sequence: $z_1 = \W\x$, $J = \x^\top z_1$.
Accordingly, the reverse sweep starts with $\lhu{J} = \frac{\partial J}{\partial J} = 1$, which is supplemented with the adjoint partial derivative $\frac{\partial J}{\partial z_1} = \x$, i.e., $\lhu{z_1} = \x\lhu{J}$. Following, $\lhu{z_1}$ is supplemented with $\frac{\partial z_1}{\partial \W} = \x^\top$, i.e., $\lhu{\W} = \lhu{z_1}\x^\top = \x\x^\top$.
This becomes very intuitive once we realize $\frac{\partial J}{\partial \W} = \frac{\partial J}{\partial z_1} \left( \frac{\partial J}{\partial J} \right) \frac{\partial z_1}{\partial \W}$, i.e., the adjoint partial derivatives appear around $\lhu{J}$ in a compounding fashion. Whether an adjoint term appears to the left or right depends on a few basic rules.
These rules are summarized below:
\begin{IEEEenumerate}
	\item The basic \emph{product rule} for the matrix product $z_1 = z_2 z_3$ is twofold: if $z_2$ is a function of the variable of differentiation ($\W$) then $\lhu{z_2} = \lhu{z_1} z_3^\top$, and if  $z_3$ is a function of $\W$ then $\lhu{z_3} = z_2^\top \lhu{z_1}$.
	When both terms are function of $\W$ both rules apply, causing the reverse sweep to branch and leading to more than one answer.
	In that case, the final answer is the sum of the answers (leaves) from each branch.
	\item The Hadamard (element-wise) product $z_1 = z_2 \circ z_3$ leads to another product rule, specifically, $\lhu{z_2} = \lhu{z_1} \circ z_3$ and $\lhu{z_3} = z_2 \circ \lhu{z_1}$.
	\item Other Hadamard operations simply apply standard derivation rules element-wise.
	For $z_1 = f\left(z_2\right)$, it yields $\lhu{z_2} = \frac{\partial f\left(z_2\right)}{\partial z_2} \circ \lhu{z_1}$.
	\item Matrix rearrangement operations such as the transpose, $\mathrm{vec}_{mn}\left(\cdot\right)$, or $\mathrm{diag}\left(\cdot\right)$ have a form like $z_1 = f\left(z_2\right)$, leading to the rule $\lhu{z_2} = f^{-1}(\lhu{z_1})$, where $f^{-1}(\cdot)$ is the transpose, $\mathrm{reshape}_{mn}(\cdot)$, or $\mathrm{diag}^{-1}(\cdot)$, respectively, ignoring elements set to zero by $f(\cdot)$. Thus, $\mathrm{vec}_{mn}\left(\cdot\right)$ stacks columns of a matrix and $\mathrm{reshape}_{mn}(\cdot)$ unstacks them back into a $m \times n$ matrix, while $\mathrm{diag}\left(\cdot\right)$ creates a diagonal matrix from a vector and $\mathrm{diag}^{-1}(\cdot)$ returns the diagonal elements of a matrix into a vector.
	\item Simple addition of the form $z_1 = z_2 + z_3$ also yields a branch per term as $\lhu{z_2} = \lhu{z_1}$ and $\lhu{z_3} = \lhu{z_1}$.
	\item Matrix inversion of the form $z_1 = z_2^{-1}$ yields the rule $\lhu{z_2} = -z_1^\top \lhu{z_1} z_1^\top$.
	\item The determinant of a matrix $z_1 = \det \left(z_2\right)$ leads to the rule $\lhu{z_2} = \lhu{z_1} z_1 z_2^{-\top}$.
	\item The trace of a matrix $z_1 = \mathrm{tr}(z_2)$ yields the rule $\lhu{z_2} = \lhu{z_1}\eye$.
\end{IEEEenumerate}

\newpage
\subsection{Derivation of the Gradient for MISA with Scale Control, Equation (9)}\label{MISA4}
Let
\begin{IEEEeqnarray}{Lcl}
	\check{I}(\y) & \> = \> &
	- \sum_{m=1}^M J_{D_m} + \frac{1}{2}\sum_{k=1}^K J_{C_k}
	- f \left(K,\beta_k,\lambda_k,\eta_k,d_k,\nu_k\right) \nonumber
	\\
	&&
	- \sum_{k=1}^K \frac{\eta_k-1}{N}\sum_{n=1}^{N} J_{F_{kn}}
	+ \sum_{k=1}^K \frac{\lambda_k}{N}\sum_{n=1}^{N} J_{E_{kn}} \mathrm{,}%
	\nonumber
\end{IEEEeqnarray}
\noindent where
\begin{IEEEeqnarray}{Rll}
	J_{D_m}  & = \> &
	\ln\abs{\det{\Lambdabf_m}} = \sum_{i = 1}^{C_m} \ln\abs{\sigma_{mi}} \mathrm{,} \nonumber
	\\
	&&
	\mathrm{with} \> \Lambdabf_m = \U_m^\top \W_m \V_m
	\>\> \mathrm{and} \>\> \W_m \overset{SVD}{=} \U_m \Lambdabf_m \V_m^\top \nonumber
	\\ 
	J_{C_k} & = \> &
	\ln\det\left(\bm{\gamma}_k^{} \Sigmabf_k^\y \bm{\gamma}_k^\top\right) \mathrm{,} \nonumber
	\\
	&&
	\mathrm{with} \> \bm{\gamma}_k^{} = {\left(\left(\eye_{d_k} \circ \Sigmabf_k^\y\right)^{-1}\right)}^{\circ \frac{1}{2}}
	\nonumber
	\\
	&&\mathrm{and} \> \Sigmabf_k^\y = \frac{1}{N-1}\PP_k\W\X\X^\top\W^\top\PP_k^\top \nonumber
	\\
	J_{F_{kn}} & = \> &
	\ln\left( \y_{kn}^\top {\left[ \bm{\gamma}_k^{}\Sigmabf_k^\y \bm{\gamma}_k^\top \right]}^{-1} \y_{kn}\right) \nonumber
	\\
	J_{E_{kn}} & = \> &
	{\left( \y_{kn}^\top {\left[ \bm{\gamma}_k^{}\Sigmabf_k^\y \bm{\gamma}_k^\top \right]}^{-1} \y_{kn}\right)}^{\beta_k} \nonumber
	\\ 
	f \left(K,\beta_k,\lambda_k,\eta_k,d_k,\nu_k\right) & = \> &
	\sum_{k=1}^K \left[
	\ln\beta_k + \nu_k \ln \lambda_k + \ln\Gamma\left( \frac{d_k}{2} \right) - \frac{d_k}{2} \ln \pi - \ln\Gamma\left( \nu_k \right)
	\right] \mathrm{.} \nonumber
\end{IEEEeqnarray}

Then, the gradient of $J$ with respect to $\W$ is
\begin{IEEEeqnarray}{Lcl}
	\nabla\check{I}(\W)_{m[k]} & \> = \> &
	- \sum_{m=1}^M \nabla_\W J_{D_m} + \frac{1}{2}\sum_{k=1}^K \nabla_\W J_{C_k} \nonumber \\
	&&
	- \sum_{k=1}^K \frac{\eta_k-1}{N}\sum_{n=1}^{N} \nabla_\W J_{F_{kn}}
	+ \sum_{k=1}^K \frac{\lambda_k}{N}\sum_{n=1}^{N} \nabla_\W J_{E_{kn}} \mathrm{.}%
	\nonumber
\end{IEEEeqnarray}
In the following, $\nabla_\W J_{D_m}$, $\nabla_\W J_{C_k}$, $\nabla_\W J_{F_{kn}}$, and $\nabla_\W J_{E_{kn}}$ terms are derived separately.

\vfill
\subsection{Derivation of $\nabla_\W J_{D_m}$:}\label{gradJD}
Here we show that
\begin{IEEEeqnarray}{Lcl}
	\nabla_{\W} J_{D_m} &\>=\>&
	\frac{\partial}{\partial \W}\ln\abs{\det{\Lambdabf_m}} = \mathbf{U}_m^{} \Lambdabf_m^{-1} \mathbf{V}_m^\top = (\W_m^{-})^\top%
	\\%
	&& \mathrm{with} \> \Lambdabf_m^{} = \mathbf{U}_m^\top \W_m^{} \mathbf{V}_m^{}%
	\nonumber%
\end{IEEEeqnarray}
\begin{footnotesize}
	\begin{minipage}[t]{0.5\textwidth}
		\begin{IEEEeqnarray}{*Rcl}
			J_{D_m} &\>=\>& \ln \abs{\det \Lambdabf_m}%
			\nonumber \\%
			&=& \ln z_1 \mathrm{,} ~~z_1 = \abs{\det (\mathbf{U}_m^\top \W_m^{} \mathbf{V}_m^{})}%
			\nonumber \\%
			\lhu{z_1} &=& \frac{\partial \ln z_1}{\partial z_1} \lhu{J_{D_m}} \mathrm{,} ~~\lhu{J_{D_m}}=1%
			\nonumber \\%
			&=& \frac{1}{z_1}%
			\nonumber \\%
			z_1 &=& \abs{z_2} \mathrm{,} ~~z_2 = \det (\mathbf{U}_m^\top \W_m^{} \mathbf{V}_m^{})%
			\nonumber \\%
			\lhu{z_2} &=& \frac{\partial \abs{z_2}}{\partial z_2} \lhu{z_1}%
			\nonumber \\%
			&=& \mathrm{sign}(z_2) \frac{1}{z_1} = \frac{\mathrm{sign}(z_2)}{\abs{z_2}} = \frac{1}{z_2}%
			\nonumber \\%
			z_2 &=& \det z_3 \mathrm{,} ~~z_3 = \mathbf{U}_m^\top \W_m^{} \mathbf{V}_m^{} = \Lambdabf_m^{}%
			\nonumber%
		\end{IEEEeqnarray}
	\end{minipage}%
	\begin{minipage}[t]{0.5\textwidth}
		\begin{IEEEeqnarray}{*Rcl}
			\lhu{z_3} &\>=\>& \lhu{z_2} z_2 z_3^{-\top}%
			\nonumber \\%
			&=& \frac{1}{z_2} z_2 \left(\mathbf{V}_m^\top \W_m^{} \mathbf{U}_m^{}\right)^{-1}%
			\nonumber \\%
			&=& \Lambdabf_m^{-1}%
			\nonumber \\%
			z_3 &=& \mathbf{U}_m^\top z_4 \mathrm{,} ~~z_4 = \W_m^{} \mathbf{V}_m^{}%
			\nonumber \\%
			\lhu{z_4} &=& \mathbf{U}_m \lhu{z_3}%
			\nonumber \\%
			&=& \mathbf{U}_m^{} \Lambdabf_m^{-1}%
			\nonumber \\%
			z_4 &=& \W_m^{} \mathbf{V}_m^{}%
			\nonumber \\%
			\lhu{\W_m} &=& \lhu{z_4} \mathbf{V}_m^{\top}%
			\nonumber \\%
			&=& \mathbf{U}_m^{} \Lambdabf_m^{-1} \mathbf{V}_m^{\top}%
			\nonumber \\%
			&=& (\W_m^{-})^\top%
			\nonumber%
		\end{IEEEeqnarray}
	\end{minipage}
\end{footnotesize}

\newpage
\subsection{Derivation of $\nabla_\W J_{C_k}$:}\label{gradJC}
Here we show that
\begin{IEEEeqnarray}{Rcl}
	\nabla_{\W} J_{C_k} &\>=\>&
	\frac{\partial}{\partial \W}\ln\det\left(\bm{\gamma}_k^{} \Sigmabf_k^\y \bm{\gamma}_k^\top\right) = 2 \PP_k^\top \left[ {\Z_{\Sigmabf}}^{-1} - \left(\eye_{d_k} \circ \Z_{\Sigmabf}\right)^{-1} \right] \PP_k\W\X\X^\top%
	\\%
	&&
	\mathrm{with} \> \bm{\gamma}_k^{} = {\left(\left(\eye_{d_k} \circ \Sigmabf_k^\y\right)^{-1}\right)}^{\circ \frac{1}{2}}%
	\nonumber \\%
	&&
	\mathrm{and} \> \Sigmabf_k^\y = \frac{1}{N-1}\PP_k\W\X\X^\top\W^\top\PP_k^\top%
	\nonumber%
\end{IEEEeqnarray}
\begin{scriptsize}
	\begin{IEEEeqnarray}{*RclRclRcl}
		J_{C_k} &\>=\>&
		\ln\det\left(\bm{\gamma}_k^{} \Sigmabf_k^\y \bm{\gamma}_k^\top\right) \mathrm{,}%
		\nonumber \\%
		&=& \ln z_1 \mathrm{,} ~~z_1 = \det\left(\bm{\gamma}_k^{} \Sigmabf_k^\y \bm{\gamma}_k^\top\right)%
		\nonumber \\%
		\lhu{z_1} &=& \frac{\partial \ln z_1}{\partial z_1} \lhu{J_{C_k}} \mathrm{,} ~~\lhu{J_{C_k}}=1%
		\nonumber \\%
		&=& \frac{1}{z_1}%
		\nonumber \\%
		z_1 &=& \det z_2 \mathrm{,} ~~z_2 = \bm{\gamma}_k^{} \Sigmabf_k^\y \bm{\gamma}_k^\top%
		\nonumber \\%
		\lhu{z_2} &=& \lhu{z_1} z_1 z_2^{-\top}%
		\nonumber \\%
		&=& \bm{\gamma}_k^{-1} {\Sigmabf_k^\y}^{-1} \bm{\gamma}_k^{-1}%
		\nonumber \\%
		z_2 &=& \bm{\gamma}_k z_3 \mathrm{,} ~~z_3 = \Sigmabf_k^\y \bm{\gamma}_k^\top%
		\nonumber \\%
		\lhu{z_3} &=& \bm{\gamma}_k^\top \lhu{z_2}
		& &&
		& \lhu{{\bm\gamma}_k} &\>=\>& \lhu{z_2} z_3^\top%
		\nonumber \\%
		&=& \bm{\gamma}_k^\top \bm{\gamma}_k^{-1} {\Sigmabf_k^\y}^{-1} \bm{\gamma}_k^{-1}
		& &&
		& &=& \bm{\gamma}_k^{-1} {\Sigmabf_k^\y}^{-1} \bm{\gamma}_k^{-1} \bm{\gamma}_k^{} \Sigmabf_k^\y%
		\nonumber \\%
		&=& {\Sigmabf_k^\y}^{-1} \bm{\gamma}_k^{-1}
		& &&
		& &=& \bm{\gamma}_k^{-1}%
		\nonumber \\%
		z_3 &=& \Sigmabf_k^\y z_4 \mathrm{,} ~~z_4 = \bm{\gamma}_k^\top
		& &&
		& \bm{\gamma}_k &=& z_9^{\frac{1}{2}} \mathrm{,} ~~z_9 = \left(\eye_{d_k} \circ \Sigmabf_k^\y\right)^{-1} = \bm{\gamma}_k^2%
		\nonumber \\%
		\lhu{\Sigmabf_k^\y} &=& \lhu{z_3} z_4^\top
		& \lhu{z_4} &\>=\>& \Sigmabf_k^\y \lhu{z_3}
		& \lhu{z_9} &=& \frac{\partial z_9^{\frac{1}{2}}}{\partial z_9} \lhu{{\bm\gamma}_k}%
		\nonumber \\%
		&=& {\Sigmabf_k^\y}^{-1} \bm{\gamma}_k^{-1} \bm{\gamma}_k^{}
		& &=& \Sigmabf_k^\y {\Sigmabf_k^\y}^{-1} \bm{\gamma}_k^{-1}
		& &=& \frac{1}{2} z_9^{-\frac{1}{2}} \bm{\gamma}_k^{-1}%
		\nonumber \\%
		&=& {\Sigmabf_k^\y}^{-1}
		& &=& \bm{\gamma}_k^{-1}
		& &=& \frac{1}{2} \bm{\gamma}_k^{-2}%
		\nonumber \\%
		\Sigmabf_k^\y &=& \frac{1}{N-1} \Z_{\Sigmabf} \mathrm{,} ~~\Z_{\Sigmabf} = \PP_k\W\X\X^\top\W^\top\PP_k^\top
		& ~~~~z_4 &=& {\bm{\gamma}}_k
		& z_9^{} &=& z_{10}^{-1} \mathrm{,} ~~z_{10} = \eye_{d_k} \circ \Sigmabf_k^\y%
		\nonumber \\%
		\lhu{\Z_{\Sigmabf}} &=& \frac{1}{N-1} \lhu{\Sigmabf_k^\y}
		& \lhu{{\bm\gamma}_k} &=& \lhu{z_4}^\top
		& \lhu{z_{10}} &=& -z_9^\top\lhu{z_9}z_9^\top%
		\nonumber \\%
		\Z_{\Sigmabf} &=& \PP_k z_5 \mathrm{,} ~~z_5 = \W\X\X^\top\W^\top\PP_k^\top
		& &=& \bm{\gamma}_k^{-1}
		& &=& -\frac{1}{2} \bm{\gamma}_k^2 \bm{\gamma}_k^{-2} \bm{\gamma}_k^2%
		\nonumber \\%
		\lhu{z_5} &=& \PP_k^\top \lhu{\Z_{\Sigmabf}}
		& &\vdots&
		& &=& -\frac{1}{2} \bm{\gamma}_k^2%
		\nonumber \\%
		&=& \frac{1}{N-1} \PP_k^\top \lhu{\Sigmabf_k^\y}
		& \lhu{\W} &=& \frac{2}{N-1} \PP_k^\top \lhu{\Sigmabf_k^\y} \PP_k\W\X\X^\top
		& ~~~~z_{10} &=& \eye_{d_k} \circ \Sigmabf_k^\y%
		\nonumber \\%
		z_5 &=& \W z_6 \mathrm{,} ~~z_6 = \X\X^\top\W^\top\PP_k^\top
		& &&
		& \lhu{\Sigmabf_k^\y} &=& \eye_{d_k} \circ \lhu{z_{10}}%
		\nonumber \\%
		\lhu{z_6} &=& \W^\top \lhu{z_5}
		& \lhu{\W} &\>=\>& \lhu{z_5} z_6^\top
		& &=& \eye_{d_k} \circ \left(-\frac{1}{2} \bm{\gamma}_k^2\right)%
		\nonumber \\%
		&=& \frac{1}{N-1} \W^\top\PP_k^\top \lhu{\Sigmabf_k^\y}
		& &=& \frac{1}{N-1} \PP_k^\top \lhu{\Sigmabf_k^\y} \PP_k\W\X\X^\top
		& &=& -\frac{1}{2} \left(\eye_{d_k} \circ \Sigmabf_k^\y\right)^{-1}%
		\nonumber \\%
		z_6 &=& \X\X^\top z_7 \mathrm{,} ~~z_7 = \W^\top\PP_k^\top
		& &&
		& &\vdots&%
		\nonumber \\%
		\lhu{z_7} &=& \X\X^\top \lhu{z_6}
		& &&
		& \lhu{\W} &=& \frac{2}{N-1} \PP_k^\top \lhu{\Sigmabf_k^\y} \PP_k\W\X\X^\top%
		\nonumber \\%
		&=& \frac{1}{N-1} \X\X^\top\W^\top\PP_k^\top \lhu{\Sigmabf_k^\y}
		& &&
		& &&%
		\nonumber \\%
		z_7 &=& z_8 \PP_k^\top \mathrm{,} ~~z_8 = \W^\top
		& &&
		& &&%
		\nonumber \\%
		\lhu{z_8} &=& \lhu{z_7} \PP_k
		& &&
		& &&%
		\nonumber \\%
		&=& \frac{1}{N-1} \X\X^\top\W^\top\PP_k^\top \lhu{\Sigmabf_k^\y} \PP_k
		& &&
		& &&%
		\nonumber \\%
		z_8 &=& \W^\top
		& &&
		& &&%
		\nonumber \\%
		\lhu{\W} &=& \lhu{z_8}^\top
		& &&
		& &&%
		\nonumber \\%
		&=& \frac{1}{N-1} \PP_k^\top \left[\lhu{\Sigmabf_k^\y}\right]^\top \PP_k\W\X\X^\top
		& &&
		& &&%
		\nonumber%
	\end{IEEEeqnarray}
\end{scriptsize}

Thus, collecting all $\lhu{\W}$ together we have the final form:
\begin{IEEEeqnarray}{Rcl}
	\lhu{\W} &\>=\>& \frac{2}{N-1} \PP_k^\top \left[ {\Sigmabf_k^\y}^{-1} - \left(\eye_{d_k} \circ \Sigmabf_k^\y\right)^{-1} \right] \PP_k\W\X\X^\top
	\nonumber \\%
	&=& 2 \PP_k^\top \left[ {\Z_{\Sigmabf}}^{-1} - \left(\eye_{d_k} \circ \Z_{\Sigmabf}\right)^{-1} \right] \PP_k\W\X\X^\top
	\nonumber%
\end{IEEEeqnarray}

\vfill
\subsection{Derivation of $\nabla_\W J_{F_{kn}}$:}\label{gradJF}
Here we show that
\begin{IEEEeqnarray}{Rcl}
	\nabla_{\W} J_{F_{kn}} &\>=\>&
	\frac{\partial}{\partial \W} \ln\left( \y_{kn}^\top {\left[ \bm{\gamma}_k^{}\Sigmabf_k^\y \bm{\gamma}_k^\top \right]}^{-1} \y_{kn}\right)
	\nonumber \\%
	&=& 2 \PP_k^\top {\left[ \bm{\gamma}_k^{}\Sigmabf_k^\y \bm{\gamma}_k^\top \right]}^{-1} \y_{kn}^{} z_{kn}^{-1} \x_n^\top + \frac{2}{N-1} \PP_k^\top \left[ T_1 + T_2 \right] \PP_k\W\X\X^\top%
	\\%
	&&
	\mathrm{with} \> T_1 = -{\Sigmabf_k^\y}^{-1}\bm{\gamma}_k^{-1} \y_{kn}^{} z_{kn}^{-1} \y_{kn}^\top \bm{\gamma}_k^{-1}{\Sigmabf_k^\y}^{-1}
	\nonumber \\
	&&
	T_2 = \eye_{d_k} \circ \left( {\Sigmabf_k^\y}^{-1}\bm{\gamma}_k^{-1} \y_{kn}^{} z_{kn}^{-1} \y_{kn}^\top \bm{\gamma}_k \right)
	\nonumber \\
	&&
	\> \y_{kn} = \PP_k \W \x_{n}
	\nonumber \\
	&&
	\bm{\gamma}_k^{} = {\left(\left(\eye_{d_k} \circ \Sigmabf_k^\y\right)^{-1}\right)}^{\circ \frac{1}{2}}%
	\nonumber \\%
	&&
	\mathrm{and} \> \Sigmabf_k^\y = \frac{1}{N-1}\PP_k\W\X\X^\top\W^\top\PP_k^\top%
	\nonumber%
\end{IEEEeqnarray}
\begin{scriptsize}
	\begin{IEEEeqnarray}{*RclRclRcl}
		J_{F_{kn}} &\>=\>& \ln z_{kn} \mathrm{,} ~~z_{kn} = \y_{kn}^\top {\left[ \bm{\gamma}_k^{}\Sigmabf_k^\y \bm{\gamma}_k^\top \right]}^{-1} \y_{kn}
		& &&
		& &&%
		\nonumber \\%
		\lhu{z_{kn}} &=& z_{kn}^{-1}
		& &&
		& &&%
		\nonumber \\%
		z_{kn} &=& z_1 z_2 \mathrm{,} ~~z_1 = \y_{kn}^\top \mathrm{,} ~~z_2 = \mathbf{M}^{-1}\y_{kn}
		& &&
		& &&%
		\nonumber \\%
		\lhu{z_1} &=& \lhu{z_{kn}} z_2^\top
		& ~~~~\lhu{z_2} &\>=\>& z_1^\top \lhu{z_{kn}}
		& &&%
		\nonumber \\%
		&=& z_{kn}^{-1} \y_{kn}^\top \mathbf{M}^{-1}
		& &=& \y_{kn} z_{kn}^{-1}
		& &&%
		\nonumber \\%
		z_1 &=& \y_{kn}^\top
		& z_2 &=& z_4 \y_{kn} \mathrm{,} ~~z_4 = \mathbf{M}^{-1}
		& &&%
		\nonumber \\%
		\lhu{\y_{kn}} &=& \lhu{z_1}^\top
		& \lhu{z_4} &=& \lhu{z_2} \y_{kn}^\top
		& ~~~~\lhu{\y_{kn}} &\>=\>& z_4^\top \lhu{z_2}%
		\nonumber \\%
		&=& \mathbf{M}^{-1} \y_{kn}^{} z_{kn}^{-1}
		& &=& \y_{kn}^{} z_{kn}^{-1} \y_{kn}^\top
		& &=& \mathbf{M}^{-1} \y_{kn} z_{kn}^{-1}%
		\nonumber \\%
		\y_{kn} &=& \PP_k z_3 \mathrm{,} ~~z_3 = \W\x_n
		& z_4 &=& \mathbf{M}^{-1}
		& &\vdots&%
		\nonumber \\%
		\lhu{z_3} &=& \PP_k^\top \lhu{\y_{kn}}
		& \lhu{\mathbf{M}} &=& -z_4^\top \lhu{z_4} z_4^\top
		& \lhu{\W} &=& \PP_k^\top \mathbf{M}^{-1} \y_{kn}^{} z_{kn}^{-1} \x_n^\top %
		\nonumber \\%
		&=& \PP_k^\top \mathbf{M}^{-1} \y_{kn}^{} z_{kn}^{-1}
		& &=& -\mathbf{M}^{-1} \y_{kn}^{} z_{kn}^{-1} \y_{kn}^\top \mathbf{M}^{-1}
		& &&%
		\nonumber \\%
		z_3 &=& \W \x_n
		& &\vdots& \text{using derivation in~\ref{gradJC}}
		& &&%
		\nonumber \\%
		\lhu{\W} &=& \lhu{z_3}\x_n^\top
		& \lhu{\W} &=& \frac{2}{N-1} \PP_k^\top 
		& &&%
		\nonumber \\%
		&=& \PP_k^\top \mathbf{M}^{-1} \y_{kn}^{} z_{kn}^{-1} \x_n^\top
		& && \left[ -\bm{\gamma}_k^\top \mathbf{M}^{-1} \bm{\gamma}_k^{} + \eye_{d_k} \circ \left( -\frac{1}{2} \bm{\gamma}_k^\top \mathbf{M}^{-1} \bm{\gamma}_k^{} \Sigmabf_k^{\y} \bm{\gamma}_k^{2} \right) \right]
		& &&%
		\nonumber \\%
		&=& \PP_k^\top {\left[ \bm{\gamma}_k^{}\Sigmabf_k^\y \bm{\gamma}_k^\top \right]}^{-1} \y_{kn}^{} z_{kn}^{-1} \x_n^\top
		& && \PP_k\W\X\X^\top
		& &&%
		\nonumber%
	\end{IEEEeqnarray}
\end{scriptsize}

Thus, collecting all $\lhu{\W}$ together we have the final form:
\begin{IEEEeqnarray}{Rcl}
	\lhu{\W} &\>=\>& 2 \PP_k^\top {\left[ \bm{\gamma}_k^{}\Sigmabf_k^\y \bm{\gamma}_k^\top \right]}^{-1} \y_{kn}^{} z_{kn}^{-1} \x_n^\top
	\nonumber \\%
	&+& \frac{2}{N-1} \PP_k^\top \left[ -{\Sigmabf_k^\y}^{-1}\bm{\gamma}_k^{-1} \y_{kn}^{} z_{kn}^{-1} \y_{kn}^\top \bm{\gamma}_k^{-1}{\Sigmabf_k^\y}^{-1} + \eye_{d_k} \circ \left( {\Sigmabf_k^\y}^{-1}\bm{\gamma}_k^{-1} \y_{kn}^{} z_{kn}^{-1} \y_{kn}^\top \bm{\gamma}_k \right)  \right] \PP_k\W\X\X^\top
	\nonumber%
\end{IEEEeqnarray}

\newpage
\subsection{Derivation of $\nabla_\W J_{F_{kn}}$:}\label{gradJE}
Here we show that
\begin{IEEEeqnarray}{Rcl}
	\nabla_{\W} J_{F_{kn}} &\>=\>&
	\frac{\partial}{\partial \W} \left( \y_{kn}^\top {\left[ \bm{\gamma}_k^{}\Sigmabf_k^\y \bm{\gamma}_k^\top \right]}^{-1} \y_{kn}\right)^{\beta_k}
	\nonumber \\%
	&=& 2\beta_k \PP_k^\top {\left[ \bm{\gamma}_k^{}\Sigmabf_k^\y \bm{\gamma}_k^\top \right]}^{-1} \y_{kn}^{} z_{kn}^{\beta_k-1} \x_n^\top + \frac{2\beta_k}{N-1} \PP_k^\top \left[ T_1 + T_2 \right] \PP_k\W\X\X^\top%
	\\%
	&&
	\mathrm{with} \> T_1 = -{\Sigmabf_k^\y}^{-1}\bm{\gamma}_k^{-1} \y_{kn}^{} z_{kn}^{\beta_k-1} \y_{kn}^\top \bm{\gamma}_k^{-1}{\Sigmabf_k^\y}^{-1}
	\nonumber \\
	&&
	T_2 = \eye_{d_k} \circ \left( {\Sigmabf_k^\y}^{-1}\bm{\gamma}_k^{-1} \y_{kn}^{} z_{kn}^{\beta_k-1} \y_{kn}^\top \bm{\gamma}_k \right)
	\nonumber \\
	&&
	\> \y_{kn} = \PP_k \W \x_{n}
	\nonumber \\
	&&
	\bm{\gamma}_k^{} = {\left(\left(\eye_{d_k} \circ \Sigmabf_k^\y\right)^{-1}\right)}^{\circ \frac{1}{2}}%
	\nonumber \\%
	&&
	\mathrm{and} \> \Sigmabf_k^\y = \frac{1}{N-1}\PP_k\W\X\X^\top\W^\top\PP_k^\top%
	\nonumber%
\end{IEEEeqnarray}
\begin{scriptsize}
	\begin{IEEEeqnarray}{*RclRclRcl}
		J_{E_{kn}} &\>=\>& z_{kn}^{\beta_k} \mathrm{,} ~~z_{kn} = \y_{kn}^\top {\left[ \bm{\gamma}_k^{}\Sigmabf_k^\y \bm{\gamma}_k^\top \right]}^{-1} \y_{kn}
		& &&
		& &&%
		\nonumber \\%
		\lhu{z_{kn}} &=& \beta_k z_{kn}^{\beta_k-1}
		& &&
		& &&%
		\nonumber \\%
		&\vdots& \text{using derivation in~\ref{gradJF}}
		& &&
		& &&%
		\nonumber \\%
		\lhu{\W} &\>=\>& 2\beta_k \PP_k^\top {\left[ \bm{\gamma}_k^{}\Sigmabf_k^\y \bm{\gamma}_k^\top \right]}^{-1} \y_{kn}^{} z_{kn}^{\beta_k-1} \x_n^\top
		\nonumber \\%
		&+& \frac{2\beta_k}{N-1} \PP_k^\top \left[ -{\Sigmabf_k^\y}^{-1}\bm{\gamma}_k^{-1} \y_{kn}^{} z_{kn}^{\beta_k-1} \y_{kn}^\top \bm{\gamma}_k^{-1}{\Sigmabf_k^\y}^{-1} + \eye_{d_k} \circ \left( {\Sigmabf_k^\y}^{-1}\bm{\gamma}_k^{-1} \y_{kn}^{} z_{kn}^{\beta_k-1} \y_{kn}^\top \bm{\gamma}_k \right)  \right] \PP_k\W\X\X^\top
		\nonumber%
	\end{IEEEeqnarray}
\end{scriptsize}

\ifCLASSOPTIONcaptionsoff
  \newpage
\fi

\bibliographystyle{IEEEtran}
\bibliography{IEEEabrv,rogers_refs}